%% file: main.tex
\documentclass{article} 
\usepackage{etoolbox}
\newtoggle{arxiv}
\toggletrue{arxiv}

\iftoggle{arxiv}{
  \setlength{\textwidth}{6.5in}
  \setlength{\textheight}{9in}
  \setlength{\oddsidemargin}{0in}
  \setlength{\evensidemargin}{0in}
  \setlength{\topmargin}{-0.5in}
  \newlength{\defbaselineskip}
  \setlength{\defbaselineskip}{\baselineskip}
  \setlength{\marginparwidth}{0.8in}
}{
\usepackage[compact]{titlesec}
\titlespacing{\section}{0pt}{*1}{*0}
\titlespacing{\subsection}{0pt}{*1.5}{*0}

\usepackage[subtle, mathdisplays=tight, charwidths=normal, leading=normal]{savetrees}

\addtolength\textfloatsep{-0.5em}
\addtolength\intextsep{-0.2em}

\def\setstretch#1{\renewcommand{\baselinestretch}{#1}}
\setstretch{0.985}
\addtolength{\parskip}{-1pt}

}

\usepackage[pagebackref=true,breaklinks,colorlinks,citecolor=blue,citecolor=blue,linkcolor=blue,urlcolor=blue,hypertexnames=false]{hyperref}
\usepackage{url}

\usepackage{graphicx}
\usepackage{amsmath}
\usepackage{amssymb}
\usepackage{booktabs}
\usepackage{makecell}
\usepackage{multirow}
\usepackage{overpic}
\usepackage{comment}
\usepackage{float}
\usepackage[export]{adjustbox}
\usepackage{enumitem}
\usepackage{array}
\usepackage[svgnames]{xcolor}
\usepackage{colortbl}%
\usepackage{caption}
\usepackage{wrapfig}
\usepackage{subcaption}
\usepackage{tabularx}
\usepackage{natbib}

\definecolor{lightpink}{rgb}{1, 0.8, 0.8}
\definecolor{lightcyan}{rgb}{0.85, 1, 1}
\newcommand\sotaa{\textcolor{red}}
\newcommand\sotab{\textcolor{blue}}

\newcommand{\myrowcolour}{\rowcolor[gray]{0.925}}
\newcommand{\myrowcolourpink}{\rowcolor{pink}}
\newcommand\revise{\textcolor{black}}

\def\vs{\emph{vs.\ }}

\def\ie{\emph{i.e.,\ }}

\title{Hierarchical Information Flow for Generalized Efficient Image Restoration}

\author{
\textbf{Yawei Li}$^{1}$\quad
\textbf{Bin Ren}$^{2,3}$\quad
\textbf{Jingyun Liang}$^{1}$\quad
\textbf{Rakesh Ranjan}$^{4}$\quad
\textbf{Mengyuan Liu}$^{5}$\\
\textbf{Nicu Sebe}$^{3}$\quad
\textbf{Ming-Hsuan Yang}$^{6}$\quad 
\textbf{Luca Benini}$^{1}$ \\
$^1$ETH Z\"urich,
$^2$University of Pisa,
$^3$University of Trento,
$^4$Meta Reality Labs,\\
$^5$Peking University,
$^6$University of California, Merced
}

\date{}

\begin{document}

\maketitle

\input{sections_revise/0-abstract}
\input{sections_revise/1-introduction}
\input{sections_revise/2-relatedworks}
\input{sections_revise/3-motivation}

\input{sections_revise/4-method}
\input{sections_revise/5-model-scaling-up}

\input{sections_revise/6-experiments}
\input{sections_revise/7-conclusion}

\bibliography{main}
\bibliographystyle{iclr2025_conference}

\clearpage
\input{sections_revise/X-supplementary}

\end{document}

%% file: sections_revise/0-abstract.tex
\begin{abstract}
While vision transformers show promise in numerous image restoration (IR) tasks, the challenge remains in efficiently generalizing and scaling up a model for multiple IR tasks.
%
%
To strike a balance between efficiency and model capacity for a generalized transformer-based IR method, we propose a \textbf{h}ierarchical \textbf{i}nformation flow mechanism for \textbf{i}mage \textbf{r}estoration, dubbed \textbf{Hi-IR}, which progressively \revise{propagates information} among pixels in a bottom-up manner. 
Hi-IR constructs a hierarchical information tree representing the degraded image across three levels. 
Each level encapsulates different types of information, with higher levels encompassing broader objects and concepts and lower levels focusing on local details. 
%
Moreover, the hierarchical tree architecture removes long-range self-attention, improves the computational efficiency and memory utilization, thus preparing it for effective model scaling.
Based on that, we explore model scaling to improve our method's capabilities, which is expected to positively impact IR in large-scale training settings. 
Extensive experimental results show that Hi-IR achieves state-of-the-art performance in seven common image restoration tasks, 
affirming its effectiveness and generalizability.
\end{abstract}


%% file: sections_revise/1-introduction.tex
\section{Introduction}
\label{sec:introduction}

Image restoration (IR) aims to improve image quality by recovering high-quality visuals from observations degraded by noise, blur, and downsampling. 
To address this series of inherently ill-posed problems, numerous methods have been developed primarily for a single degradation, including convolutional neural networks (CNNs)~\citep{dong2014learning,kim2016accurate,lim2017enhanced}, vision transformers (ViTs)~\citep{chen2021pre,liang2021swinir,li2023efficient}, and state space models (Mamba)~\citep{gu2023mamba,guo2024mambair}.  
However, the intricate and varied nature of degradation presents formidable challenges to the prevailing IR methodologies. 
In particular, several coupled problems remain for general IR:
\begin{itemize}[leftmargin=*]
    \item First, \textit{there is a lack of a generalized computational mechanism for efficient IR.} 
    A general IR framework needs to deal with images with varying characteristics, such as different types and intensities of degradation, as well as varying resolutions. 
    Techniques designed for specific IR tasks might not apply to other problems. \revise{Simply combining computational mechanisms designed for different IR tasks does not necessarily result in an efficient solution.} \revise{Thus, it is a challenge to design a mechanism that is both efficient and capable of generalizing well to different IR tasks.}
    \item Second, \textit{there is no systematic approach for guiding model scaling.} Current image restoration networks are typically limited to 10-20M parameters. Addressing multiple degradations often requires increasing the model capacity by scaling up the model size. Yet, diminished model performance is observed by simply scaling up the model. Therefore, the challenge of systematically scaling up IR models remains unresolved. 
    \item Third, \textit{it is still unclear how well a single model can generalize across different IR tasks.} Existing approaches tend to focus on either a single task or a subset of IR tasks. The generalizability of a single model across a broader range of IR tasks has to be thoroughly validated.
\end{itemize}

This paper addresses the aforementioned questions in Sec.~\ref{sec:methodology}, Sec.~\ref{sec:model_scaling_up}. and Sec.~\ref{sec:experiments}, respectively. 
We propose a hierarchical information flow principle designed specifically for general IR tasks. This principle establishes relationships between pixels on multiple levels and progressively aggregates information across multiple levels, which is essential for general IR.
Compared with existing approaches such as convolution~\citep{zhang2018residual}, global attention~\citep{chen2021pre}, and window attention~\citep{li2023efficient}, hierarchical information flow balances complexity with the efficiency of comprehending global contexts, ensuring an optimized process for integrating information across various scales and regions.
The underlying design principle opens the door to different realizations. 
Considering the effectiveness and efficiency for image modeling, we propose a new architecture based on a three-level \textbf{h}ierarchical \textbf{i}nformation flow mechanism for \textbf{i}mage \textbf{r}estoration (\ie \textbf{Hi-IR}). 
Hi-IR employs a series of progressive computational stages for efficient information flow. 
The first-level (L1) computational block works within individual patches, fostering local information exchange and generating intermediate node patches. 
Then, a second-level (L2) block works across the intermediate node patches and allows for the effective propagation of information beyond the local scope. 
As a final step, the third-level (L3) information flow block bridges the gaps between the isolated node patches from the first two stages.

\begin{figure}[!t]
\begin{center}
    \includegraphics[width=1.0\linewidth]{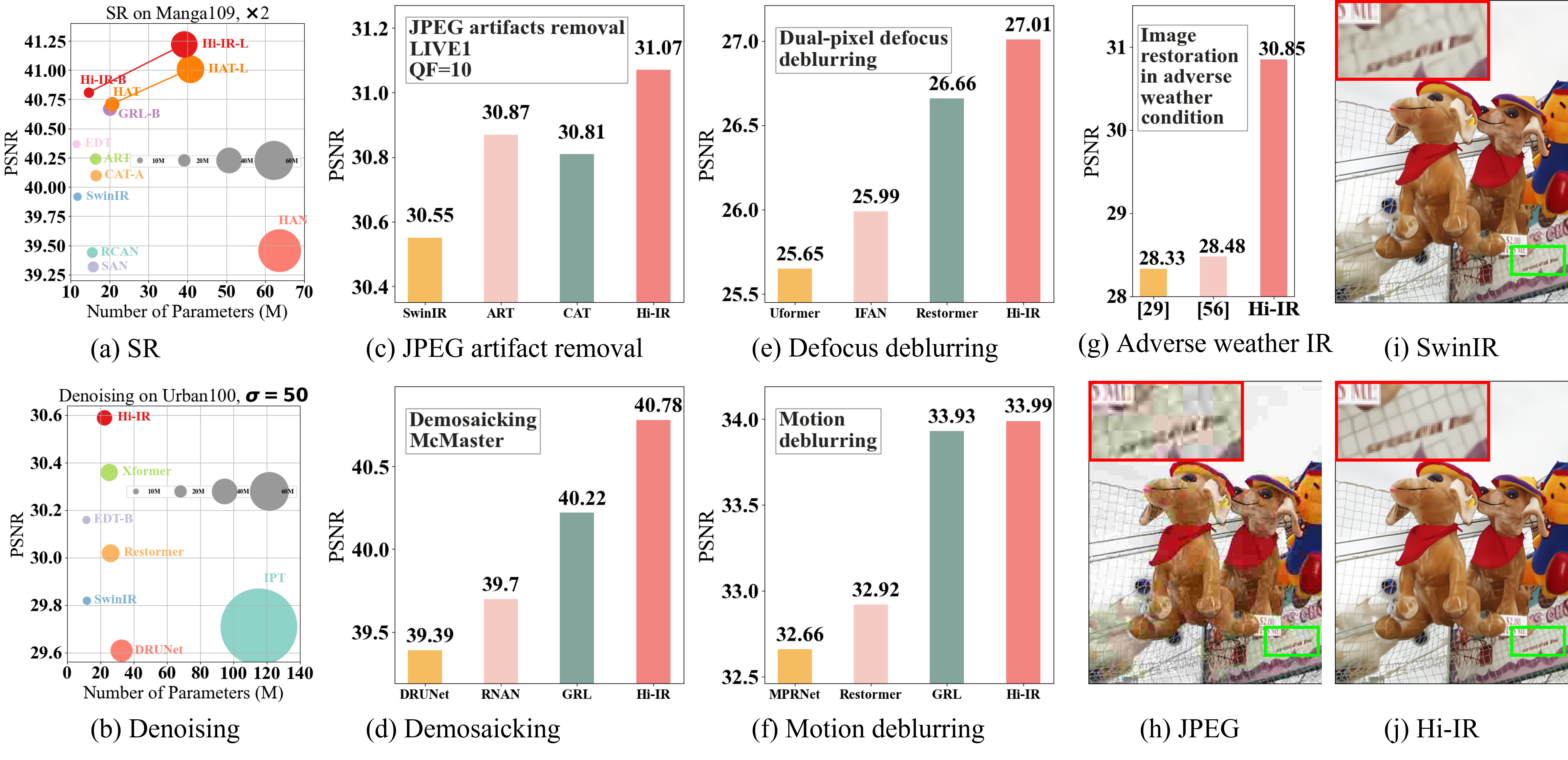}%
    \vspace{-2mm}
    \caption{The proposed Hi-IR is notable for its efficiency and effectiveness (a)-(b), generalizability across seven image restoration tasks (a)-(g), and improvements in the visual quality of restored images (h)-(j).}
    \label{fig:teaser}
\end{center}%
\end{figure}

Motivated by the scaling law~\citep{brown2020language,touvron2023llama,kang2023scaling,saharia2022photorealistic,yu2024scaling}, we scale up the model to enhance the model capacity. We analyze the reason why it is difficult to scale up IR models. 
As a remedy to the notorious problem~\citep{lim2017enhanced,chen2023activating}, this paper proposes three strategies that systematically encompass model training, weight initialization, and model design to enable effective model scaling.

This paper validates the generalizability of the proposed hierarchical information flow mechanism through rigorous experiments on multiple aspects. First, we investigate the performance of the model trained for a specific degradation type and intensity, including downsampling, motion blur, defocus blur, noise, and JPEG compression. Second, we validate that the model can handle a single degradation type with multiple intensities.
Furthermore, we demonstrate that a single model can generalize effectively across multiple tasks, validating its versatility.
Our main contributions are summarized as follows:
\begin{itemize}[leftmargin=*]
    \item[$\bullet$] We introduce a novel hierarchical information flow principle for image restoration, which facilitates progressive global information exchange and mitigates the curse of dimensionality.

    \item[$\bullet$] We propose Hi-IR, a compact image restoration model guided by the design principle, to propagate information for image restoration efficiently. 

    \item[$\bullet$] We examine the challenge of training convergence for model scaling-up in IR and propose mitigation strategies.

    \item[$\bullet$] Extensive experiments demonstrate the generalizability of the proposed hierarchical information flow mechanism. The proposed Hi-IR consistently outperforms state-of-the-art image restoration methods for multiple tasks.
\end{itemize}

%% file: sections_revise/2-relatedworks.tex
\section{Related Work}
\label{sec:related-work}
\bigskip
\noindent{\textbf{Image Restoration}} focuses on recovering high-quality images from their degraded counterparts. 
As a challenging problem, IR has captured substantial interest in academic and industrial circles, leading to practical applications such as denoising, deblurring, super-resolution (SR), and so on.
The landscape of IR has shifted with the evolution of deep learning and the increased availability of computational resources, notably GPUs. Neural network-based pipelines, fueled by advancements in deep learning, have supplanted earlier model-based solutions~\citep{richardson1972bayesian,liang2021swinir,li2023lsdir}. 
Numerous CNN models have been proposed~\citep{anwar2020densely,li2022blueprint,dong2014learning,zhang2017beyond} for different IR tasks.
However, despite their effectiveness, CNNs have been found to struggle in propagating long-range information within degraded input images. 
This challenge is attributed to the limited receptive field of CNNs, which, in turn, constrains the overall performance of CNN-based methods~\citep{chen2022cross,zhang2022accurate,li2023efficient}.

\bigskip
\noindent{\textbf{Vision Transformer-based Models for IR}} have been proposed to address the problem of global information propagation inspired by the success of Transformer architecture in machine translation~\citep{vaswani2017attention} and high-level vision tasks~\citep{dosovitskiy2020image}. 
Specifically, IPT~\citep{chen2021pre} applies ViTs for IR. 
Despite promising results, it is difficult to use full-range self-attention within the ViTs because the computational complexity increases quadratically with the image size. 
As a remedy, numerous methods explore ViTs in an efficient yet effective manner. 
In particular, SwinIR~\citep{liang2021swinir} conducts multi-head self-attention (MSA) window-wise. 
A shift operation is applied to achieve the global interactive operation~\citep{liu2021swin}. 
Uformer~\citep{wang2022uformer} proposes to propagate much more global information with a UNet structure but still with window self-attention. 
Other methods~\citep{zamir2022restormer,chen2022cross,ren2024sharing} re-design the attention operation with much more exquisite efforts, \revise{such as cross-covariance across channel dimensions~\citep{zamir2022restormer}, rectangle-window self-attention~\citep{li2021efficient}, sparse self-attention~\cite{huang2021shuffle}, and graph-attention~\citep{ren2024sharing}, spatial shuffle~\citep{huang2021shuffle}, and random spatial shuffle~\cite{xiao2023random}. }
However, these transformer-based solutions cannot balance the ability to generalize to multiple IR tasks and the computational complexity of global modeling.
%
In this paper, we propose a general and efficient IR solution which hierarchically propagates information in a tree-structured manner, simultaneously incorporating inputs from lower and higher semantic levels.

%% file: sections_revise/3-motivation.tex
\section{Methodology}
\label{sec:methodology}

\subsection{Motivation}
This paper aims to propose a general and efficient IR framework.
Before presenting technical details, we discuss the motivation behind the proposed hierarchical information flow mechanism.

In this work, we demonstrate the pivotal role of the information flow in decoding low-level features, which become more pronounced with the introduction of ViTs. 
CNNs employ successive convolutions that inherently facilitate progressive information flow beyond local fields. In contrast, image restoration transformers typically achieve information flow via self-attention across manually partitioned windows, combined with a window-shifting mechanism. 
When the flow of contextual information between different regions or features within an image is restricted, a model's ability to reconstruct high-quality images from low-quality counterparts is significantly hindered. 
This effect can be observed by deliberately isolating the information flow in Swin transformer.
In Tab.~\ref{tab:information_isolation}, the flow of information across windows is prohibited by removing the window-shifting mechanism, which leads to a decrease in PSNR on the validation datasets (specifically, a 0.27 dB drop for DF2K training, and a 0.23 dB drop for LSDIR training). 
The obvious reductions indicate that information isolation degrades the performance of IR techniques, likely because the algorithms are deprived of the contextual clues necessary for accurately reconstructing finer image details. 

\input{tables/ablation_model_design1}

Secondly, we observe that information propagation on fully connected graphs is not always necessary or beneficial for improving the performance of the IR networks~\citep{chen2021pre,zamir2022restormer}. 
%
As ViTs generate distinct graphs for each token, early attempts to facilitate global information dissemination led to the curse of dimensionality, causing quadratic growth in computational complexity with token increase~\citep{wang2020linformer,liu2021swin}.
%
Subsequent attention mechanisms, building graphs based on windows, achieve better IR results. 
However, the benefits of expanding the window size tend to plateau.
Tab.~\ref{tab:plateau} shows the effect of window size versus performance. 
The quality of the reconstructed images improves as the window size grows from 8 to 32, evident from rising PSNR values. 
Yet, with larger windows, the gains decrease, accompanied by a sharp increase in memory footprint and computational demands, resulting in a plateau effect. 
This prompt a reassessment of the information propagation mechanism on large windows. 
The challenge lies in balancing the scope and the complexity of window attention while enhancing global information propagation efficiency.

\begin{figure}[!t]
    \centering
    \includegraphics[width=0.99\linewidth]{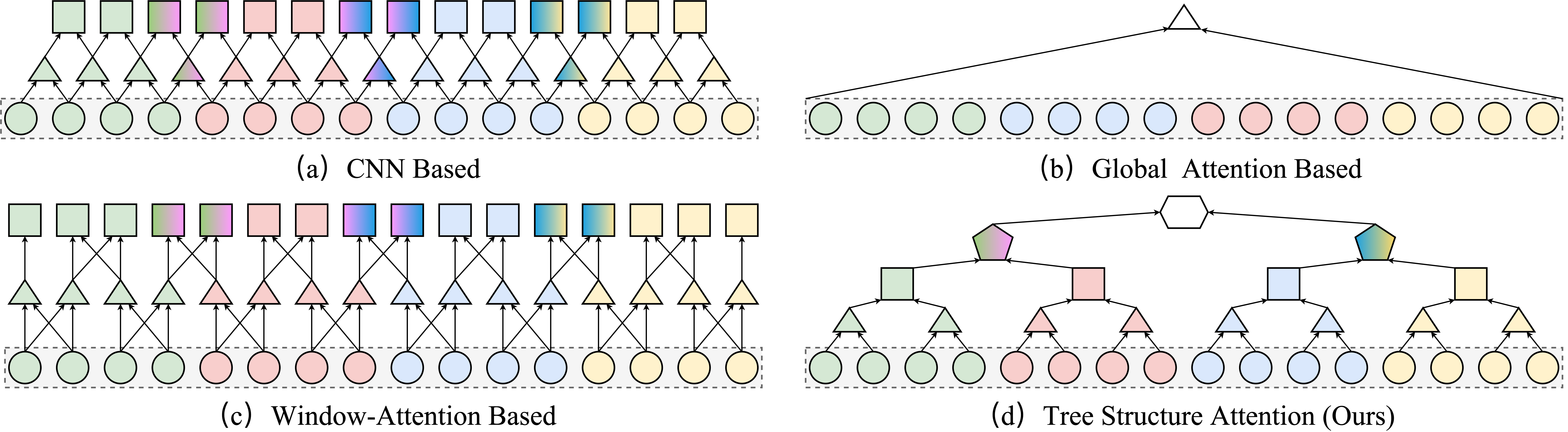}
    \vspace{-2mm}
    \caption{Illustration of information flow principles. \revise{The colors represent local information, with their blending indicating propagation beyond the local region.} (a) The CNN-based. (b) The original ViTs based. (c) Window attention based. (d) The proposed hierarchical information flow prototype.}
    \label{fig:motivation}
\end{figure}

\bigskip
\noindent\textbf{Effective information flow.}
The above analysis emphasizes the crucial role of effective information flow in modern architectural designs. 
CNN-based methods propagate information slowly within a small region covered by the filter (Fig.~\ref{fig:motivation}(a)).
A large receptive field has to be achieved by the stack of deep layers. 
Global attention based ViT propagates information directly across the whole sequence with a single step. 
However, the computational complexity grows quadratically with the increase of tokens (Fig.~\ref{fig:motivation}(b)). 
To address this problem, window attention in Fig.~\ref{fig:motivation}(c) propagates information across two levels but still has a limited receptive field even with shift operation.

To facilitate fast and efficient information flow across the image, we propose a hierarchical information flow principle shown in Fig.~\ref{fig:motivation}(d). 
In this model, information flows progressively from the local scope, aggregated in several intermediate levels, and disseminated across the whole sequence. 
This new design principle is more efficient in that it enables a global understanding of the input sequence with several operations. 
\revise{Moreover, the actual implementation of the tree structure such as the depth of the tree can be configured to ensure computational efficiency.}
One realization in this work is a three-level information flow model shown below. The space and time complexity of the three information flow mechanisms is given in Appx.~\ref{sec:space_time_complexity}. The proposed hierarchical information flow mechanism is more efficient in propogating information to the global range under similar space and time complexity of window attention.

%% file: tables/ablation_model_design1.tex
\begin{table}[!tb]
\parbox{0.45\linewidth}{\centering
    \vspace{-4mm}
    \caption{Removing shifted windows leads to degraded SR performance. PSNR is reported on Urban100 dataset for $4\times$ SR.}
    \label{tab:information_isolation}
\setlength{\extrarowheight}{0.7pt}
\setlength{\tabcolsep}{2.5pt}
\scalebox{0.75}{
    \begin{tabular}{c|cc}
        \toprule[0.1em]
        \multirow{2}{*}{Training Dataset}               & \multicolumn{2}{c}{Window Shift} \\ \cline{2-3}
    
        & Yes & No \\ \midrule[0.1em]
        
        DF2K~\citep{agustsson2017ntire} &27.45  &27.18 (\textcolor{red}{-0.27}) \\
        
        LSDIR~\citep{li2023lsdir}       &27.87	&27.64 (\textcolor{red}{-0.23}) \\
        \bottomrule[0.1em]
    \end{tabular}
    }
}
\hspace{4pt}
\parbox{0.53\linewidth}{
\centering
    \vspace{-4mm}
    \caption{Plateau effect of enlarged window size \revise{reported on Urban100 for $4\times$ SR}. Window size larger than 32 is not investigated due to the OOM issue. }
    \label{tab:plateau}
\setlength{\extrarowheight}{0.7pt}
\setlength{\tabcolsep}{2.5pt}
\scalebox{0.75}{
    \begin{tabular}{ccccc}
    \toprule[0.1em]
    Window size & PSNR & PSNR gain & GPU Mem. & Computation \\ \midrule[0.1em]
    8	&27.42	&0.00	&14.63GB	& $\times 1$\\
    16	&27.80	&+0.38	&17.22GB	& $\times 4$\\
    32	&28.03	&+0.22	&27.80GB	& \textcolor{red}{$\times 16$}\\ 
    \bottomrule[0.1em]
    \end{tabular}
    }
}
\end{table}

%% file: sections_revise/4-method.tex

\subsection{Hierarchical Tree-Structured Information Flow}
\label{subsec:tipm}

\begin{figure}[!t]
    \centering
    \includegraphics[width=0.99\linewidth]{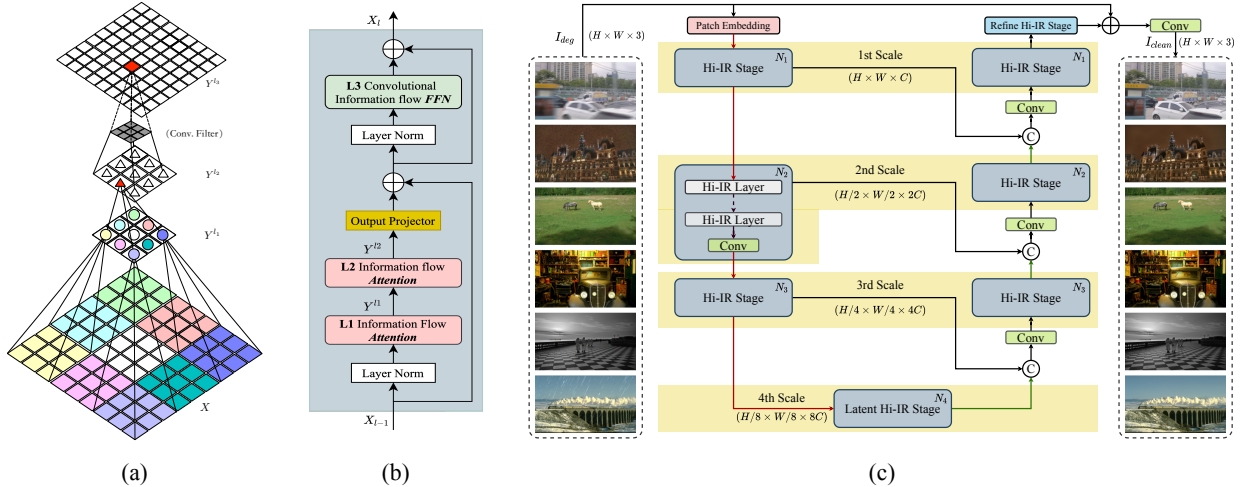}
    \caption{Illustrations of: 
    (a) The hierarchical information flow.
    (b) The proposed hierarchical information flow transformer layer.
    (c) The overall framework of the proposed Hi-IR.}
    \label{fig:architecture}
\end{figure}

As shown in Fig.~\ref{fig:architecture}(a) - (c), the hierarchical tree-structured information flow mechanism consists of three levels and aims to effectively model both the local and the global information for a given feature $X \in \mathbb{R}^{H \times W \times C}$ efficiently. \revise{We denote the information within $X$ as $l_0$ level meta-information.}

\bigskip
\noindent\textbf{L1 information flow attention} is achieved by \revise{applying MSA to the input feature $X$ within a $p\times p$ patch. To facilitate the MSA, the input feature is first partitioned into local patches, leading to $X'\in \mathbb{R}^{\frac{HW}{p^2}\times p^2 \times C}$. Then feature $X'$ is linearly projected into query ($Q^{l_1}$), key ($K^{l_1}$), and value ($V^{l_1}$). Self-attention within the local patches is denoted as $Y_{i}^{l_1} = \operatorname{SoftMax}({Q_{i}^{l_1}(K_{i}^{l_1})^{\top}}/{\sqrt{d}})V_{i}^{l_1}$, where $i$ index the windows, and $d$ represents the head dimension.}
%
This process is shown in Fig.~\ref{fig:architecture}(a). 
\revise{Each node within the $Y^{l_1}$ grid represents all the $l_0$ level meta-information derived from its corresponding original window, marked by the same color.}

\bigskip
\noindent\textbf{L2 information flow attention} is achieved upon the previous $l_1$ level information $Y^{l_1}$. 
Despite the expanded scope of information within each grid of $Y^{l_1}$, comprehensive cross-window information propagation remains a challenge. 
%
%
\revise{As indicated conceptually in Fig.~\ref{fig:motivation}(d), 2D $s \times s$ non-overlapping local patches $p \times p$ in L1 information flow should be grouped together to form a broader $P \times P$ region for L2 information flow. Different from the previous operations~\citep{xiao2023random,huang2021shuffle}, we do not expand to the whole image in this phase due to two considerations: 1) \textbf{The computational complexity of attention in the global image can be quite high}; 2) \textbf{Not all global image information is relevant to the reconstruction of a specific pixel}.}
\revise{To facilitate MSA, the dispersed pixels need to be grouped together via a permutation operation.}
\revise{The seemingly complex operation is simplified by first reshaping the input tensor to $\hat{Y}^{l_1} \in \mathbb{R}^{\frac{H}{P} \times s \times p \times \frac{W}{P} \times s \times p \times C}$, followed by a permutation to form $(Y')^{l_1} \in \mathbb{R}^{(\frac{H}{P} \times \frac{W}{P} \times p^2) \times s^2 \times C}$.}
The simple permutation operation facilitates the distribution of $l_1$ information nodes across a higher level region, ensuring each window contains a comprehensive, cross-window patch-wise $l_2$ information set without hurting the overall information flow.

To better integrate the permuted information $(Y^{\prime})^{l_1}$, we further project $(Y^{\prime})^{l_1}$ to $Q^{l_1}$, $K^{l_1}$, and $V^{l_1}$. And the second MSA ($L_2$ Information flow attention in Fig.~\ref{fig:architecture}(b)) among patches is applied via
$Y_{i}^{l_2} = \operatorname{SoftMax}(Q_{i}^{l_1}(K_{i}^{l_1})^{\top}/{\sqrt{d}})V_{i}^{l_1}$. 
As a result, the larger patch-wise global information (colorful nodes in $Y^{l_1}$) now is well propagated to each triangle node (Fig.~\ref{fig:architecture}) in $Y^{l_2}$.

\bigskip
\noindent\textbf{L3 convolutional information flow FFN} is implemented via a $3 \times 3$ convolution operation between two $1 \times 1$ convolution operations, forming the convolutional feed-forward network in this paper and outputs the third level information $Y^{l_3}$. 
As a result, this design not only aggregates all the channel-wise information more efficiently but also enriches the inductive modeling ability~\citep{chu2022conditional,xu2021vitae} for the proposed mechanism.

\subsection{Hi-IR Layer}
\label{subsec:treeir_layer}
The Hi-IR layer, serving as the fundamental component for both architectures, is constructed based on the innovative tree-structured information flow mechanism (TIFM) introduced above, and the detailed structure is depicted in Fig.~\ref{fig:architecture}(b). 
For each Hi-IR layer, the input feature $X_{l-1}$ first passes through a layer normalization and two consecutive information propagation attentions.
After adding the shortcut, the output $X^{'}_{l}$ is fed into the convolutional feed-forward networks with another shortcut connection and outputs $X_{l}$. 
We formulate this process as follows:
\begin{equation}
    \begin{aligned}
    X^{\prime}{ }_l & =\operatorname{TIFM_{Att}}\left(\mathrm{LN}\left(X_{l-1}\right)\right)+x_{l-1}, \\
    X_l & =\operatorname{TIFM_{Conv}}\left(\operatorname{LN}\left(X^{\prime}{ }_l\right)\right)+X^{\prime}{ }_l 
    \end{aligned}
\end{equation}
where $\operatorname{TIFM_{Att}}$ consists of both the L1 and L2 information flow attention,  $\operatorname{TIFM_{Conv}}$ denotes the L3 convolutional information flow FFN.

\subsection{Overall architecture}
To comprehensively validate the effectiveness of the proposed method, similar to prior methods~\citep{chen2022simple,li2023efficient,ren2024sharing}, we choose two commonly used basic architectures including the U-shape hierarchical architecture shown in Fig.~\ref{fig:architecture}(c) and \revise{the columnar architecture shown in Fig.~\ref{fig:architecture_columnar} of Appx.~\ref{subsec:supp:architecture_details}}.
The columnar architecture is used for image SR while the U-shape architecture is used for other IR tasks.
Specifically, given degraded low-quality image $I_{low} \in \mathbb{R}^{H \times W \times 1/3}$ (1 for the grayscale image and 3 for the color image ), it was first sent to the convolutional feature extractor and outputs the shallow feature $F_{in} \in \mathbb{R}^{H \times W \times C}$ for the following Hi-IR stages/layers. $H$, $W$, and $C$ denote the height, the width, and the channels of $F_{in}$. 
For the U-shape architecture, $F_{in}$ undergoes representation learning within the U-shape structure. In contrast, for the columnar architecture, $F_{in}$ traverses through $N$ consecutive Hi-IR stages.
Both architectures ultimately generate a restored high-quality image $I_{high}$ through their respective image reconstructions. 


%% file: sections_revise/5-model-scaling-up.tex
\section{Model Scaling-Up}
\label{sec:model_scaling_up}

\input{tables/ablation_large_model}

Existing IR models are limited to a model size of 10-20M parameters.
In this paper, we develop models of medium and large sizes.
However, scaling up the model size from 15M to 57M leads to an unexpected performance drop, as shown in the pink rows of Tab.~\ref{tab:scaling_up_convergence}. In addition, as shown in Fig.~\ref{fig:convergence} of Appx.~\ref{sec:supp:model_scaling_up}, the 57M model also converges slower than the 15M model during training.

\subsection{Initial attempts} 

Existing methods handle this problem with weight initialization and rescaling techniques. For example, \citet{chen2023activating} and \citet{lim2017enhanced} reduce the influence of residual convolutional blocks by scaling those branches with a sufficiently small factor (0.01). \citet{wang2018esrgan} rescale the weight parameters in the residual blocks by a factor of 0.1. \cite{liu2022swin} intialize the weight and bias of LayerNorm as 0. In addition, we also tried the truncated normal distribution to initialize the weight parameters.
However, as shown in Tab.~\ref{tab:scaling_up_initialization_rescaling}, none of the four methods improves the convergence and performance of the scaled models, indicating that they do work for the attention modules of the IR transformers.
%
%

\subsection{The proposed model scaling-up solution} 
The initial investigation indicates that the problem can be attributed to the training strategy, the initialization of the weight, and the model design.
Thus, three methods are proposed to mitigate the model scaling problem. 
\textit{First}, we warm up the training for 50k iterations at the beginning. 
As shown in Tab.~\ref{tab:scaling_up_convergence}, this mitigates the problem of degraded performance of scaled up models, but does not solve it completely. 
\textit{Secondly}, we additionally replace heavyweight $3\times3$ convolution (\texttt{conv1} in Tab.~\ref{tab:scaling_up_convergence})
with lightweight operations besides warming up the training.
Two alternatives are considered including a linear layer (\texttt{linear} in Tab.~\ref{tab:scaling_up_convergence}) and a bottleneck block with 3 lightweight convolutions ($1\times1$ conv+$3\times3$ conv+$1\times1$ conv, \texttt{conv3} in Tab.~\ref{tab:scaling_up_convergence}). 
The number of channels of the middle $3\times3$ conv in the bottleneck blocks is reduced by a factor of 4. 
Tab.~\ref{tab:scaling_up_convergence} shows that removing the large $3\times3$ convolutions leads to a much better convergence point for the large models. 
Considering that the bottleneck block leads to better PSNR than linear layers in most cases, it is adopted in all the other experiments.
\textit{Thirdly}, we also investigate the influence of the self-attention mechanism on the convergence of scale-up models. Specifically, two attention mechanisms are compared including dot product attention~\citep{liu2021swin} and cosine similarity attention~\citep{liu2022swin}.
As shown in Tab.~\ref{tab:comparison_attention_type}, dot product self-attention performs better than cosine similarity self-attention. Thus, dot product self-attention is used throughout this paper unless otherwise stated. \revise{The rationale behind why the proposed three strategies are effective for model scaling-up is detailed in Appx.~\ref{sec:supp:model_scaling_up}.}


\subsection{Why does replacing heavyweight $3\times3$ convolution work?}
We hypothesize that replacing dense $3\times3$ convolutions with linear layers and bottleneck blocks works because of the initialization and backpropagation of the network. 

In the Xavier and Kaiming weight initialization method, the magnitude of the weights is inversely related to \texttt{fan\_in}/\texttt{fan\_out} of a layer which is the multiplication of the number of input and output channels and kernel size, namely,
\begin{align}
    f_{in} &= c_{in} \times k ^2, \\
    f_{out} &= c_{out} \times k ^2,
\end{align}
where $f_{in}$ and $f_{out}$ denotes \texttt{fan\_in} and \texttt{fan\_out}, $c_{in}$ and $c_{out}$ denotes input and output channels, and $k$ is kernel size. Thus, when a dense $3\times 3$ convolution is used, $f_{in}$ and $f_{out}$ can be large, which leads to small initialized weight parameters. This in turn leads to small gradients during the backpropagation. When the network gets deeper, the vanishing gradients could lead to slow convergence. When dense $3\times3$ convolution is replaced by linear layers, the kernel size is reduced to 1. When the bottleneck module is used, the number of input and output channels of the middle $3\times3$ convolution in the bottleneck block is also reduced. Thus, both of the two measures decreases the \texttt{fan\_in} and \texttt{fan\_out} values, leading to larger initialized weight parameters.

\begin{figure}[!htb]
    \centering
    \includegraphics[width=0.75\linewidth]{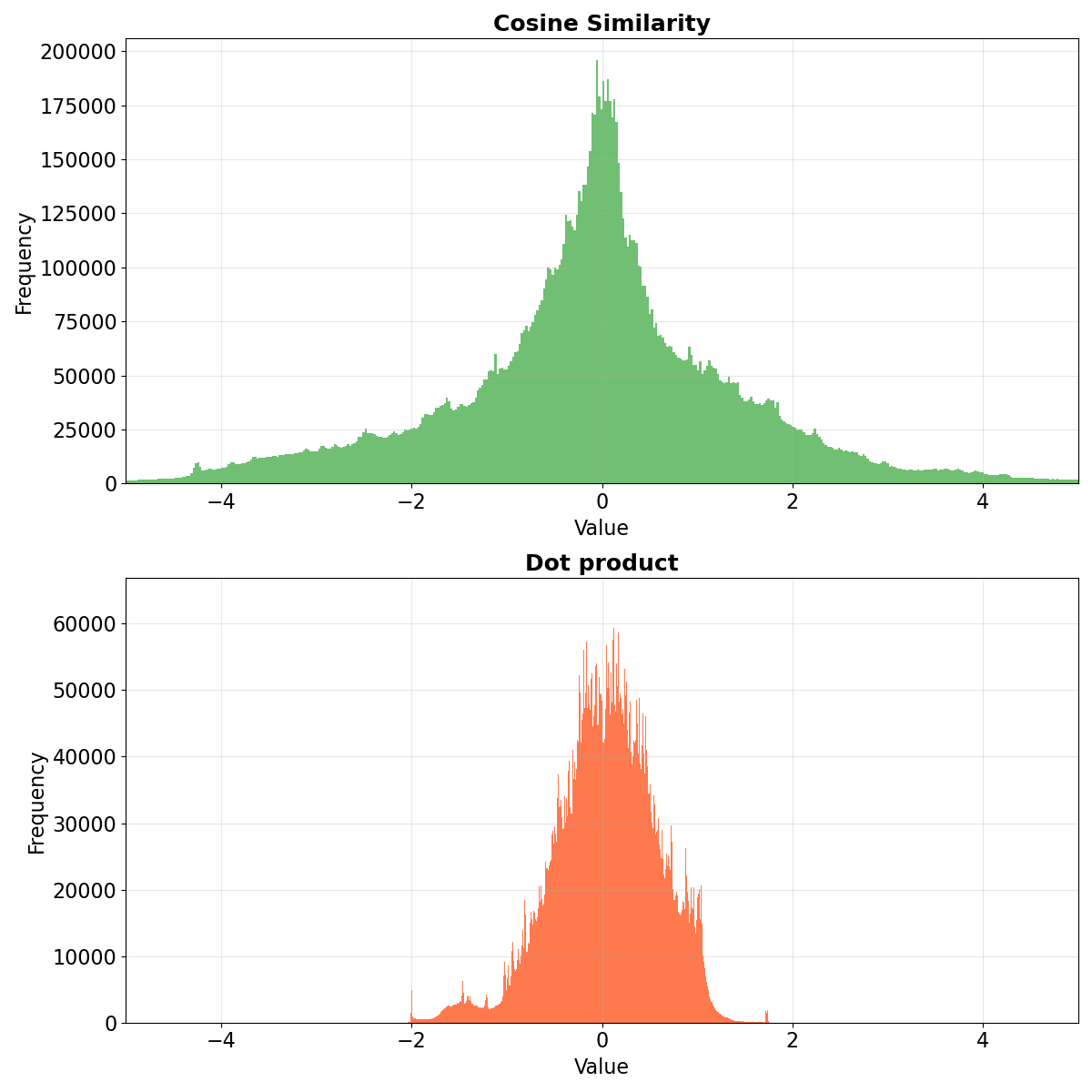}
    \caption{\revise{Comparsion of gradients between dot product and cosine similarity.}}
    \label{fig:gradient_comparison}
\end{figure}

\subsection{\revise{Why does warmup work?}}

\revise{Warmup is effective for training large models primarily because it mitigates issues related to unstable gradients and helps the optimizer gradually adapt to the model's large parameter space \citep{kalra2024warmup,goyal2017accurate}. In the early stages of training, the model's parameters are initialized randomly. A high learning rate at this stage can cause large updates, leading to unstable or divergent training due to exploding or vanishing gradients. Warmup starts with a small learning rate and gradually increases it, allowing the optimizer to find a stable path in the loss landscape before applying larger updates. Warmup enables the model to adapt gradually, avoiding overshooting minima and ensuring smoother convergence.}

\subsection{\revise{Why does dot product work better than cosine similarity?}}

\revise{As shown in Tab.~\ref{tab:comparison_attention_type}, dot product attention works better than cosine similarity attention. We analyze the gradient of dot product and cosine similary as follows. Suppose $\mathbf{q}$ denotes the query and $\mathbf{k}$ denotes the keys. Then dot product and cosine similarity between $\mathbf{q}$ and $\mathbf{k}$ are denoted as $\text{dot\_prod}(\mathbf{q}, \mathbf{k})$ and $\text{cos\_sim}(\mathbf{q}, \mathbf{k})$. 
The gradient of dot product with respect to $\mathbf{q}$ is}
\begin{equation}
        \revise{\frac{\partial}{\partial \mathbf{q}} \text{dot\_prod}(\mathbf{q}, \mathbf{k}) = \mathbf{k}.}
\end{equation}
\revise{The gradient of cosine similarity with respect to $\mathbf{q}$ is}
\begin{equation}
        \revise{\frac{\partial}{\partial \mathbf{q}} \text{cos\_sim}(\mathbf{q}, \mathbf{k}) 
         = \frac{\mathbf{k}}{\|\mathbf{q}\| \|\mathbf{k}\|} - \frac{(\mathbf{q} \cdot \mathbf{k}) \mathbf{q}}{\|\mathbf{q}\|^3 \|\mathbf{k}\|} 
         = \frac{1}{\|\mathbf{q}\|} \left(\mathbf{\hat{k}} - \text{cos\_sim}(\mathbf{q}, \mathbf{k}) \mathbf{\hat{q}}\right),}
\end{equation}
\revise{where $\mathbf{\hat{q}}$ and $\mathbf{\hat{k}}$ are normalized $\mathbf{q}$ and $\mathbf{k}$.}
\revise{The gradients with respect to $\mathbf{k}$ have the similar form. The gradient of cosine similarity involves more terms compared to the gradient of the dot product. This increased complexity in the gradient of cosine similarity makes it more prone to producing large or even unstable gradient values. We conducted a numerical analysis of the gradient values for the two attention methods, with the results presented in Fig.~\ref{fig:gradient_comparison}. As shown in the figure, the gradient of cosine similarity is indeed more prone to producing large values. This issue becomes more pronounced as the model scales up.}


%% file: tables/ablation_large_model.tex

\begin{table}[!h]
\parbox{0.52\linewidth}{\centering
\caption{Model scaling-up exploration with SR.}
\label{tab:scaling_up_convergence}
\setlength{\extrarowheight}{0.7pt}
\setlength{\tabcolsep}{2pt}
\scalebox{0.65}{
    \begin{tabular}{c|c|cc|ccccc}
    \toprule[0.1em]
    \multirow{2}{*}{\textbf{Scale}}  &  \multirow{2}{*}{\makecell{\textbf{Model} \\ \textbf{Size}}} & \multirow{2}{*}{\makecell{\textbf{Warm} \\ \textbf{up}}} & \multirow{2}{*}{\makecell{\textbf{Conv} \\ \textbf{Type}}}  &  \multicolumn{5}{c}{\textbf{PSNR}} 
    \\\cline{5-9}
    &   & & & \textbf{Set5}& \textbf{Set14} & \textbf{BSD100} & \textbf{Urban100} & \textbf{Manga109}
    \\ \hline
    \myrowcolourpink $2\times$	
     &15.69	&No	&\texttt{conv1}	&38.52	&34.47	&32.56	&34.17	&39.77	\\
    \myrowcolourpink $2\times$	
    &57.60	&No	&\texttt{conv1}	&38.33	&34.17	&32.46	&33.60	&39.37	\\ 
    $2\times$	
    &57.60	&Yes	&\texttt{conv1}	&38.41	&34.33	&32.50	&33.80	&39.51	\\
    $2\times$	
    &54.23	&Yes	&\texttt{linear}	&\sotab{38.56}	&\sotaa{34.59}	&\sotaa{32.58}	&\sotab{34.32}	&\sotab{39.87}	\\
    $2\times$	
    &55.73	&Yes	&\texttt{conv3}	&\sotaa{38.65}	&\sotab{34.48}	&\sotaa{32.58}	&\sotaa{34.33}	&\sotaa{40.12}	\\ \midrule
    \myrowcolourpink $3\times$	
    &15.87	&No	&\texttt{conv1}	&35.06	&30.91	&29.48	&30.02	&34.41	\\
    \myrowcolourpink $3\times$	
    &57.78	&No	&\texttt{conv1}	&34.70	&30.62	&29.33	&29.11	&33.96	\\ 
    $3\times$	
    &57.78	&Yes	&\texttt{conv1}	&34.91	&30.77	&29.39	&29.53	&34.12	\\
    $3\times$	
    &54.41	&Yes	&\texttt{linear}	&\sotab{35.13}	&\sotaa{31.04}	&\sotaa{29.52}	&\sotab{30.20}	&\sotab{34.54}	\\
    $3\times$	
    &55.91	&Yes	&\texttt{conv3}	&\sotaa{35.14}	&\sotab{31.03}	&\sotab{29.51}	&\sotaa{30.22}	&\sotaa{34.76}	\\ 
    \bottomrule[0.1em]
    \end{tabular}
    }
}
\hspace{4pt}
\parbox{0.46\linewidth}{
\centering
\caption{Investigated weight intialization and rescaling method for model scaling-up.}
\label{tab:scaling_up_initialization_rescaling}
\setlength{\extrarowheight}{0.7pt}
\setlength{\tabcolsep}{0.7pt}
\scalebox{0.65}{
\begin{tabular}{p{2cm}|p{5cm}|p{1.2cm}p{1.2cm}}
    \toprule[0.1em]
     \multirow{2}{*}{\textbf{Method}} & \multirow{2}{*}{\textbf{Description}} & \multicolumn{2}{c}{\textbf{PSNR on Set5}} \\ \cline{3-4}
     & & $2\times$ & $3\times$ \\ \midrule[0.1em]
     Zero LayerNorm & Initialize the weight and bias of LayerNorm as 0~\citep{liu2022swin}. & 38.35 & 34.81\\ 
     \myrowcolour Residual rescale & Rescale the residual blocks by a factor of 0.01~\citep{lim2017enhanced,chen2023activating}. &38.31 & 34.79 \\ 
     Weight rescale & Rescale the weight parameters in residual blocks by a factor of 0.1~\citep{wang2018esrgan}. & 38.36 & 34.84  \\ 
     \myrowcolour trunc\_normal\_ & Truncated normal distribution & 38.33 & 34.71 \\
     \bottomrule[0.1em]
\end{tabular}
}
}
\vspace{-3mm}
\end{table}


\begin{wrapfigure}{r}{0.45\linewidth}
    \centering
    \vspace{-4mm}
    \captionof{table}{\small{Dot production attention \vs cosine similarity attention for model scaling. PSNR reported for SR.}}
    \vspace{-2mm}
    \label{tab:comparison_attention_type}    \setlength{\tabcolsep}{2.5pt}
    \scalebox{0.68}{
    \begin{tabular}{c|c|ccccc}
    \toprule[0.1em]
    Scale	&Attn. type	&Set5	&Set14	&BSD100	&Urban100	&Manga109 \\
    \midrule
    $2\times$	&cosine sim	&38.43	&34.65	&32.56	&34.13	&39.69\\	
    \myrowcolour $2\times$	&dot prod	&38.56	&34.79	&32.63	&34.49	&39.89\\
    $4\times$	&cosine sim	&33.08	&29.15	&27.96	&27.90	&31.40\\
    \myrowcolour $4\times$	&dot prod	&33.14	&29.09	&27.98	&27.96	&31.44\\
    \bottomrule[0.1em]
    \end{tabular}
    }
    \vspace{-3mm}
\end{wrapfigure} 

%% file: sections_revise/6-experiments.tex
\section{Experiments}
\label{sec:experiments}

In this section, the results of the ablation study are first reported. 
Then we validate the effectiveness and generalizability of Hi-IR on \textbf{7} IR tasks, \ie image SR, image Dn, JPEG image compression artifact removal (CAR), single-image motion deblurring, defocus deblurring and image demosaicking, and IR in adverse weather conditions (AWC). 
More details about the training protocols and the training/test datasets are shown in Appx~\ref{sec:supp:training}.
The best and the second-best quantitative results are reported in \sotaa{red} and \sotab{blue}. 
Note that \textcolor{magenta}{\textdagger} denotes a single model that is trained to handle multiple degradation levels (\ie noise levels, and quality factors) for validating the generalizability of Hi-IR.

\subsection{Ablation Studies}
\begin{figure}[!htb]
    \centering
    \includegraphics[width=0.99\linewidth]{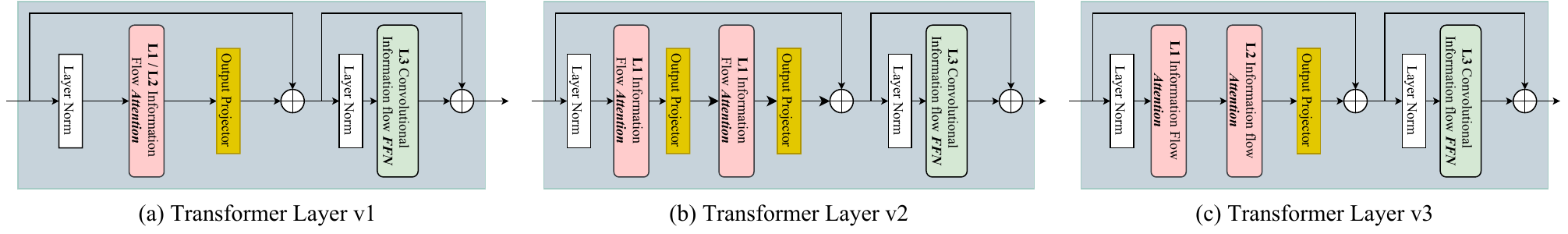}
    \vspace{-2mm}
    \caption{Comparison of three types of transformer layers designed in this paper.}
    \label{fig:attention_version}
    \vspace{-3mm}
\end{figure}

\input{tables/ablation_model_design2}
Extensive ablation experiments explore the following key aspects:

\noindent\textbf{Effect of L1 and L2 information flow.} 
\revise{One design choice for the L1/L2 information flow attentions is to decide whether to interleave them across Transformer layers or to implement them in the same layer. To validate this choice, we develop three versions, including v1 where L1 and L2 attentions alternate in consecutive layers, v2 and v3 where L1 and L2 attentions are used in the same layer (Fig.~\ref{fig:attention_version}). Compared with v1, v2 showed reduced performance despite increased model complexity. To address this issue, we introduce v3, where the projection layer between L1 and L2 is removed and the dimension of $\mathbf{Q}$ and $\mathbf{K}$ in L1/L2 attention is reduced by half to save computational complexities. The v3 L1/L2 information flows can be conceptually unified into a single flow with an expanded receptive field.}
Our ablation study reveals that v3 yielded the best performance, as evidenced by the results in Tab.~\ref{tab:ablation_sr_model_design}. Consequently, v3 was adopted for all subsequent experiments.

\noindent\textbf{\revise{Effect of the depth of the tree structure.}} 
\revise{Ablation study was conducted to evaluate the effect of the tree structure's depth. In Tab.~\ref{tab:ablation_sr_model_design}, the depth of the tree in the v1 model is 3. Removing the L3 information flow reduces the depth to 2, resulting in degraded image SR performance, even on the small Set5 dataset. Additionally, a v4 model was designed by adding an information flow attention beyond L2 to v3 model, creating a depth-4 tree structure. As shown in Tab.~\ref{tab:ablation_sr_model_design}, this increased complexity improves SR results. Thus, well-designed deeper tree structures lead to improved model performance but with increased model complexity.}
%

\noindent\textbf{Efficiency Analysis.} We report the efficiency comparison results on two IR tasks.
For the columnar architecture-based SR, our Hi-IR achieves the best PSNR with much lower parameters (28.6\% reduction) and FLOPs (31.1\% reduction), and runtime (9.95\% reduction) compared to HAT~\citep{chen2023activating}. Similar observation can also be achieved on the denoising task.

\input{tables/sr_table_full}
\input{figs/visual_sr_main}
\input{tables/dn_table}

\input{tables/jpeg_table_grayscale}

\input{tables/jpeg_table_color}

\subsection{Evaluation of Hi-IR on Various IR tasks}
\noindent\textbf{Image SR.} For the classical image SR, we compared our Hi-IR with state-of-the-art SR models. The quantitative results are shown in Tab.~\ref{tab:sr_results_full}. 
Aside from the 2nd-best results across all scales on Set5 and the 2nd-best results for the $2\times$ scale on Set14, the proposed Hi-IR archives the best PSNR and SSIM on all other test sets across all scales. 
In particular, significant improvements in terms of the PSNR on Urban100 (\revise{\ie 0.13 dB for $2\times$ SR of the base model and 0.12 dB for the $4\times$ SR of the large model}) and Manga109 (\ie 0.21 dB for $2\times$ SR) compared to HAT~\citep{chen2023activating}, but with fewer trainable parameters. 
The visual results shown in Fig.~\ref{fig:visual_sr_urban100} also validate the
effectiveness of the proposed Hi-IR in restoring more details and structural content. 
More results are in Tab.~\ref{tab:sr_results_full} of Appx.~\ref{sec:supp:more_experimental_results}, 
Fig.~\ref{fig:supp_visual_sr_b100_part1} to Fig.~\ref{fig:supp_visual_sr_manga109_part1} of Appx.~\ref{sec:supp:more_visual_results}.

\noindent\textbf{Image Denoising.} We provide both the color and the grayscale image denoising results in Tab.~\ref{tab:denoising}.
Our approach demonstrates superior performance on diverse datasets, including Kodak24, McMaster, and Urban100 for color image denoising, as well as Set12 and Urban100 for grayscale image denoising. These comparative analyses serve to reinforce the efficacy of the proposed Hi-IR, suggesting that it may exhibit a higher degree of generalization. Additionally, a closer examination of more visual results is available in Fig.~\ref{fig:supp_visual_dn_urban100} of Appx.~\ref{sec:supp:more_visual_results}, further substantiates the capabilities of Hi-IR. These results illustrate its proficiency in effectively eliminating heavy noise corruption while preserving high-frequency image details. The outcome is sharper edges and more natural textures, with no discernible issues of over-smoothness or over-sharpness.

\noindent\textbf{Image JPEG CAR.} 
For JPEG CAR, the experiments are conducted for color and grayscale images with four quality factors (\ie 10, 20, 30, and 40) under two experimental settings (\ie \textcolor{magenta}{\textdagger}, one single model is trained to handle multiple quality factors, and each model for each image quality). The results for color and grayscale images are shown in Tab.~\ref{tab:jpeg_car_color} and Tab.~\ref{tab:jpeg_car_grapyscale}, respectively. We compare Hi-IR with DnCNN3~\citep{zhang2017beyond},
DRUNet~\citep{zhang2021plug},
SwinIR~\citep{liang2021swinir},
ART~\citep{zhang2022accurate},
CAT~\citep{chen2022cross} for grayscale image JPEG CAR and with QGAC~\citep{ehrlich2020quantization}, FBCNN~\citep{jiang2021FBCNN}, DRUNet~\citep{zhang2021plug}, SwinIR~\citep{liang2021swinir}, GRL-S~\citep{li2023efficient} for color image JPEG CAR.
Specifically, the quantitative results shown in Tab.~\ref{tab:jpeg_car_grapyscale} and Tab.~\ref{tab:jpeg_car_color} validate that the proposed Hi-IR outperforms \revise{most of the other comparison methods} under both settings. Visual comparisons are provided in Fig.~\ref{fig:supp_visual_jpeg_color_bsd500} 
of Appx.~\ref{sec:supp:more_visual_results} to further support the effectiveness of the proposed Hi-IR.

\input{tables/motion_deblur_gopro_hide}
\noindent\textbf{Single-Image Motion Deblurring.} The results regarding the single-image motion deblurring are shown in Tab.~\ref{tab:motion_deblurring} and Tab.~\ref{tab:motion_deblurring_realblur}. For the synthetic datasets, compared with previous stat-of-the-art GRL~\citep{li2023efficient}, the proposed Hi-IR achieves the best results on the GoPro dataset and the second-best results on HIDE datasets. 
For the real dataset, our method also achieves the new state-of-the-art performance of 40.40 PSNR on the RealBlur-R dataset and \revise{32.92} PSNR on the RealBlur-J dataset. The visual results are shown in Fig.~\ref{fig:supp_visual_motion_db_part1} and Fig.~\ref{fig:supp_visual_motion_db_part2} of Appx.~\ref{sec:supp:more_visual_results}.

\noindent\textbf{Defocus Deblurring.} We also validate the effectiveness of our Hi-IR for dual-pixel defocus deblurring. The results in Tab.~\ref{tab:defocus_deblurring} show that Hi-IR outperforms the previous methods for all three scenes. 
Compared with Restormer on the combined scenes, our Hi-IR achieves a decent performance boost of 0.35 dB for dual-pixel defocus deblurring.

\input{tables/defocus_deblur_table}

\noindent\textbf{Image Demosaicking.} We compare 
DDR~\citep{wu2016demosaicing},
DeepJoint~\citep{gharbi2016deep},
RLDD~\citep{guo2020residual},
DRUNet~\citep{zhang2021plug},
RNAN~\citep{zhang2019residual}, and
GRL-S~\citep{li2023efficient} with the proposed method for demosaicking in Tab.~\ref{tab:demosaicking}. It shows that the proposed Hi-IR archives the best performance on both the Kodak and MaMaster test datasets. Especially, \revise{0.12 dB} and \revise{0.56} dB absolute improvement compared to the current state-of-the-art GRL.

\begin{figure}[!t]
    \centering
    \captionsetup[sub]{font=scriptsize} 
    \begin{subfigure}[b]{0.24\textwidth}
        \centering
        \includegraphics[width=\textwidth]{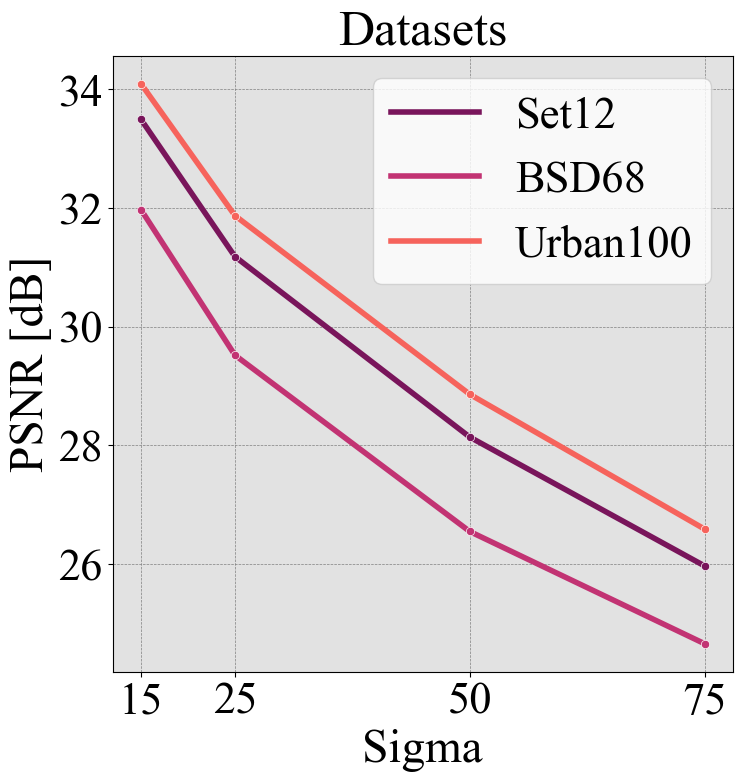}
        \caption{Grayscale image denoising}
        \label{fig:one_model1}
    \end{subfigure}
    \hfill
    \begin{subfigure}[b]{0.24\textwidth}
        \centering
        \includegraphics[width=\textwidth]{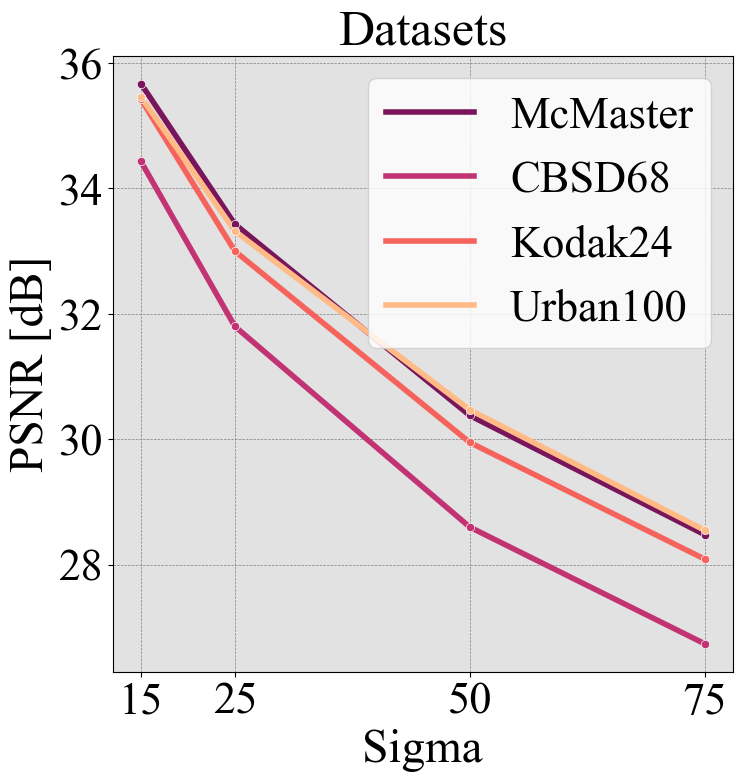}
        \caption{Color image denoising}
        \label{fig:one_model2}
    \end{subfigure}
    \hfill
    \begin{subfigure}[b]{0.24\textwidth}
        \centering
        \includegraphics[width=\textwidth]{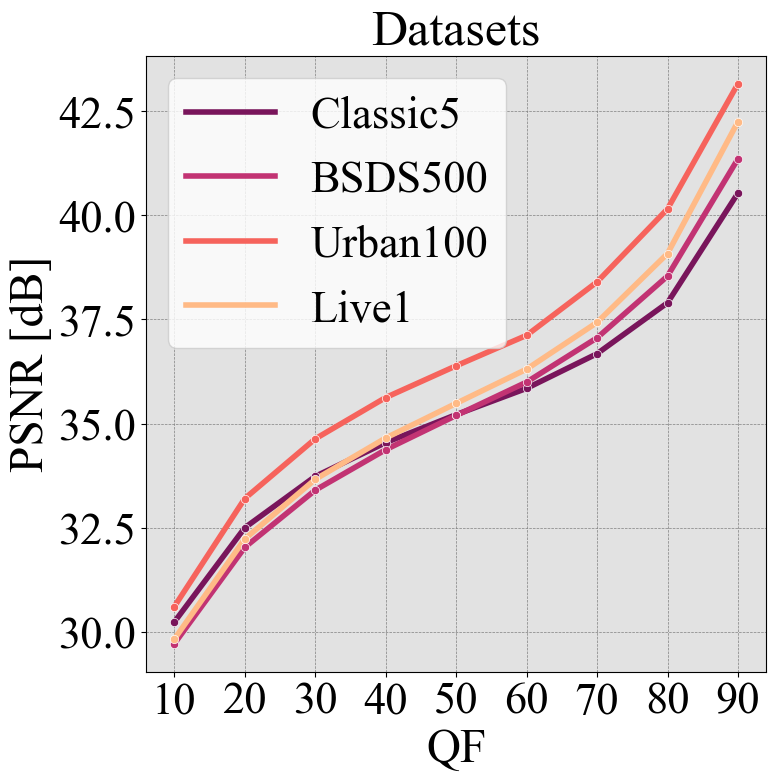}
        \caption{Grayscale image JPEG CAR}
        \label{fig:one_model3}
    \end{subfigure}
    \hfill
    \begin{subfigure}[b]{0.24\textwidth}
        \centering
        \includegraphics[width=\textwidth]{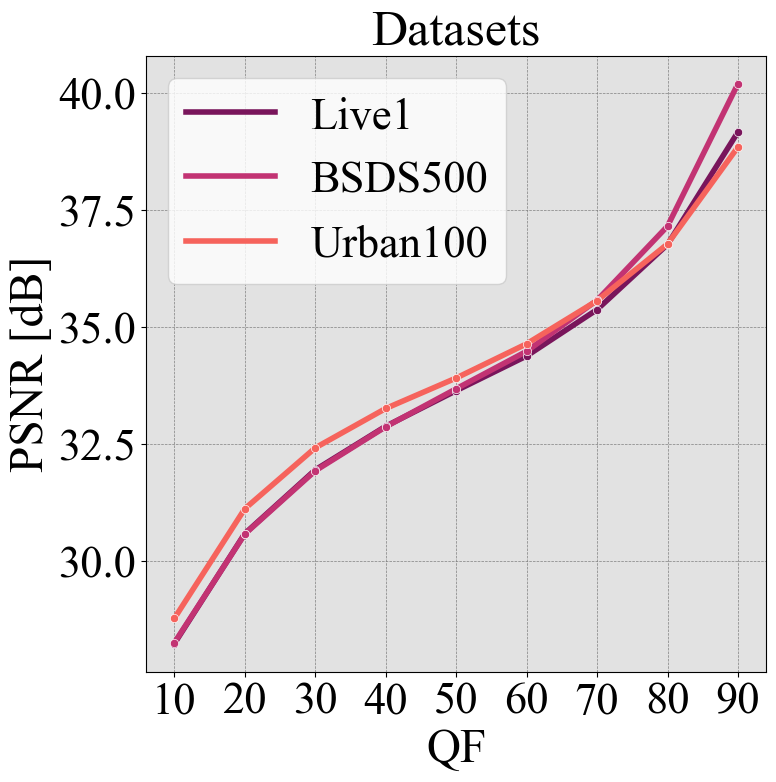}
        \caption{Color image JPEG CAR}
        \label{fig:one_model4}
    \end{subfigure}
    \vspace{-2mm}
    \caption{Training one model for multiple degradation levels.}
    \label{fig:one_model}
    \vspace{-6mm}
\end{figure}

\textbf{One model for multiple degradation levels.}
For image denoising and JPEG CAR, we trained a single model to handle multiple degradation levels. This setup makes it possible to apply one model to deal with images that have been degraded under different conditions, making the model more flexible and generalizable. 
During training, the noise level is randomly sampled from the range $[15, 75]$ while the JPEG compression quality factor is randomly sampled from the range $[10, 90]$. The degraded images are generated online. During the test phase, the degradation level is fixed to a certain value. 
The experimental results are summarized in Fig.~\ref{fig:one_model}. The numerical results for grayscale JPEG CAR are presented in Tab.~\ref{tab:jpeg_car_grapyscale}. 
These results show that in the one-model-multiple-degradation setting \textcolor{magenta}{\textdagger}, the proposed Hi-IR achieves the best performance. 

\noindent\textbf{IR in AWC.} We validate Hi-IR in adverse weather conditions like rain+fog (Test1~\citep{li2020all}), snow (SnowTest100K-L~\citep{liu2018desnownet}), and raindrops (RainDrop~\citep{qian2018attentive}). We compare Hi-IR with All-in-One~\citep{li2020all}
TransWeather~\citep{valanarasu2022transweather}, and SemanIR~\citep{ren2024sharing}.
The PSNR score is reported in Tab.~\ref{tab:weather} for each method. Our method achieves the best performance on Test1 (\ie \revise{4.6\%} improvement) and SnowTest100k-L (\ie \revise{0.09} dB improvement), while the second-best PSNR on RainDrop compared to all other methods. The visual comparison presented in Fig.~\ref{fig:supp_weather_fig} of Appx.~\ref{sec:supp:more_visual_results} also shows that our method can restore better structural context and cleaner details.

\input{tables/allweather}

%% file: tables/ablation_model_design2.tex
\begin{table}[!htb]
\parbox{0.36\linewidth}{\centering
\caption{Ablation study on model design with SR (reported on Set5).}
\label{tab:ablation_sr_model_design}
\setlength{\extrarowheight}{0.7pt}
\setlength{\tabcolsep}{0.7pt}
\scalebox{0.75}{

    \begin{tabular}{c|c|cc|cc}
    \toprule[0.1em]
    
    \multirow{3}{*}{\textbf{Scale}}	& \multirow{3}{*}{\textbf{\makecell{L1/L2 \\ Version}}}	& \multicolumn{4}{c}{\textbf{L3 Version}} \\ \cline{3-6}
    & & \multicolumn{2}{c|}{\textbf{Model size [M]}} & \multicolumn{2}{c}{\textbf{PSNR}} \\ \cline{3-6}
    & & \textbf{with} L3 & \textbf{w/o} L3 & \textbf{with} L3 & \textbf{w/o} L3 \\ \midrule[0.1em]
    $2\times$	&v1	&14.35	&11.87	&38.34	&38.31	\\
    $2\times$	&v2	&19.22	&16.74	&38.30	&38.22	\\
    $2\times$	&v3	&15.69	&13.21	&38.37	&38.35	\\ 
    $2\times$	&\revise{v4}	&\revise{17.19}	& -	&\revise{38.41}	&-	\\ \midrule
    $4\times$	&v1	&14.50	&12.02	&32.89	&32.85	\\
    $4\times$	&v2	&19.37	&16.89	&32.88	&32.77	\\
    $4\times$	&v3	&15.84	&13.36	&32.92	&32.87	\\
    $4\times$	&\revise{v4}	&\revise{17.35}	&-	&\revise{32.95}	&- \\
    
    \bottomrule[0.1em]
    \end{tabular}
    
    }
}
\hspace{2pt}
\parbox{0.62\linewidth}{
\centering
    \caption{Model efficiency \vs accuracy for SR and Dn. PSNR is reported on Urban100 dataset.}
    \label{table:params_flops_runtime}
\setlength{\extrarowheight}{0.7pt}
\setlength{\tabcolsep}{0.6pt}
\scalebox{0.75}{

    \begin{tabular}{c|l|c|cccc}
        \toprule[0.1em]
        \multirow{2}{*}{\textbf{Task}}	&\multirow{2}{*}{\textbf{Network}}	&\multirow{2}{*}{\textbf{Arch.}}	&\textbf{Params}	&\textbf{FLOPs}	&\textbf{Runtime}	&\textbf{PSNR}	\\
        & & & \textbf{[M]} & \textbf{[G]} & \textbf{[ms]} & \textbf{[dB]} \\  \hline
        \multirow{4}{*}{$4\times$ SR}	&\revise{SwinIR}~\citep{liang2021swinir}	&Columnar	&11.90	&215.32	&152.24	&27.45	\\
        	&\revise{CAT}~\citep{chen2022cross}	&Columnar	&16.60	&387.86	&357.97	&27.89	\\
        	&\revise{HAT}~\citep{chen2023activating}	&Columnar	&20.77	&416.90	&368.61	&28.37	\\ 
        	&Hi-IR (Ours)	&Columnar	&14.83	&287.20	&331.92	&28.44	\\ \midrule
        \multirow{4}{*}{Dn 50}	&\revise{SwinIR}~\citep{liang2021swinir}	&Columnar	&11.50	&804.66	&1772.84	&27.98	\\
        	&\makecell[l]{\revise{Restormer} \\~~\citep{zamir2022restormer}}	&U-shape	&26.13	&154.88	&210.44	&28.29	\\
        	&\revise{GRL}~\citep{li2023efficient}	&Columnar	&19.81	&1361.77	&3944.17	&28.59	\\
        	&Hi-IR (Ours)	&U-shape	&22.33	&153.66	&399.05	&28.91	\\ \bottomrule[0.1em]
    \end{tabular}

    }
}
\vspace{-3mm}
\end{table}

%% file: tables/sr_table_full.tex
\begin{table}[!ht]
    \scriptsize
    \setlength{\extrarowheight}{0.7pt}
    \setlength{\tabcolsep}{2pt}
    \setlength{\abovecaptionskip}{0cm}
    \begin{center}
    \caption{\textbf{\textit{Classical image SR}} results. Top-2 results are highlighted in  \textcolor{red}{red} and  \textcolor{blue}{blue}.}%
    \label{tab:sr_results_full}
    \scalebox{0.9}{
        \begin{tabular}{l|c|r|cc|cc|cc|cc|cc}
        \toprule[0.1em]
        \multirow{2}{*}{\textbf{Method}} & \multirow{2}{*}{\textbf{Scale}} & {\textbf{Params}} &  \multicolumn{2}{c|}{\textbf{Set5}} &  \multicolumn{2}{c|}{\textbf{Set14}} &  \multicolumn{2}{c|}{\textbf{BSD100}} &  \multicolumn{2}{c|}{\textbf{Urban100}} &  \multicolumn{2}{c}{\textbf{Manga109}} 
        \\
        \cline{4-13}
        &  & \multicolumn{1}{c|}{\textbf{[M]}} & PSNR$\uparrow$ & SSIM$\uparrow$ & PSNR$\uparrow$ & SSIM$\uparrow$ & PSNR$\uparrow$ & SSIM$\uparrow$ & PSNR$\uparrow$ & SSIM$\uparrow$ & PSNR$\uparrow$ & SSIM$\uparrow$
        \\
        \midrule[0.1em]
        \myrowcolour EDSR~\citep{lim2017enhanced} & $2\times$ & 40.73 & 38.11 & 0.9602 & 33.92 & 0.9195 & 32.32 & 0.9013 & 32.93 & 0.9351 & 39.10 & 0.9773\\
        SRFBN~\citep{li2019feedback} & $2\times$ & 2.14 & 38.11 & 0.9609 & 33.82 & 0.9196 & 32.29 & 0.9010 & 32.62 & 0.9328 & 39.08 & 0.9779\\
        \myrowcolour RCAN~\citep{zhang2018rcan}&	$2\times$&	15.44&	38.27&	0.9614&	34.12&	0.9216&	32.41&	0.9027&	33.34&	0.9384&	39.44&	0.9786\\
        SAN~\citep{dai2019SAN}&	$2\times$&	15.71&	38.31&	0.9620&	34.07&	0.9213&	32.42&	0.9028&	33.10&	0.9370&	39.32&	0.9792\\

        \myrowcolour HAN~\citep{niu2020HAN}&	$2\times$&	63.61&	38.27&	0.9614&	34.16&	0.9217&	32.41&	0.9027&	33.35&	0.9385&	39.46&	0.9785\\
        
            
        NLSA~\citep{mei2021NLSA}&	$2\times$&	42.63&	38.34&	0.9618&	34.08&	0.9231&	32.43&	0.9027&	33.42&	0.9394&	39.59&	0.9789\\
        \myrowcolour IPT~\citep{chen2021pre}&	$2\times$&	115.48&	38.37&	-&	34.43&	-&	32.48&	-&	33.76&	-&	-&	-\\ \hline
        
        SwinIR~\citep{liang2021swinir}&	$2\times$&	11.75&	38.42&	0.9623&	34.46&	0.9250&	32.53&	0.9041&	33.81&	0.9427&	39.92&	0.9797\\  
        \myrowcolour CAT-A~\citep{chen2022cross}    & $2\times$ &16.46 & 38.51 & 0.9626 & 34.78 & 0.9265 & 32.59 & 0.9047 & 34.26 & 0.9440 & 40.10 & 0.9805 \\
        ART~\citep{zhang2022accurate}	&$2\times$ &16.40	&38.56	&0.9629	&34.59	&0.9267	&32.58	&0.9048	&34.3	&0.9452	&40.24	&0.9808	\\		
        \myrowcolour EDT~\citep{li2021efficient}&	$2\times$&	11.48&	 {38.63}&	 {0.9632}&	 {34.80}&	0.9273&	 {32.62}&	0.9052&	34.27&	0.9456&	 {40.37}&	 {0.9811}\\ 
        GRL-B~\citep{li2023efficient} &	$2\times$&	20.05&	 {38.67}&	 {0.9647}&	 {35.08}&	 \sotab{0.9303}&	 {32.67}&	 \sotab{0.9087}&	 {35.06}&	 \sotaa{0.9505}&	 {40.67}&	 {0.9818}\\ 
        \myrowcolour HAT~\citep{chen2023activating}	&$2\times$	&20.62	&38.73	&0.9637	&35.13	&0.9282	&32.69	&0.9060	&34.81	&\sotab{0.9489}	&40.71	&0.9819	\\
        Hi-IR-B (Ours)	&$2\times$& 14.68	&38.71	&\sotab{0.9657}	&35.16	&0.9299	&32.73	&\sotab{0.9087}	&34.94	&0.9484	&40.81	&0.9830			\\
        \myrowcolour HAT-L~\citep{chen2023activating}	&$2\times$	&40.70	&\sotaa{38.91}	&0.9646	&\sotaa{35.29}	&0.9293	&\sotab{32.74}	&0.9066	&\sotab{35.09}	&\sotaa{0.9505}	&\sotab{41.01}	&\sotab{0.9831}	\\
        Hi-IR-L (Ours)	&$2\times$& 39.07	&\sotab{38.87}	&\sotaa{0.9663}	&\sotab{35.27}	&\sotaa{0.9311}	&\sotaa{32.77}	&\sotaa{0.9092}	&\sotaa{35.16}	&\sotaa{0.9505}	&\sotaa{41.22}	&\sotaa{0.9846}			\\
        \midrule[0.1em]
        
        \myrowcolour EDSR~\citep{lim2017enhanced} & $3\times$ & 43.68 & 34.65 & 0.9280 & 30.52 & 0.8462 & 29.25 & 0.8093 & 28.80 & 0.8653 & 34.17 & 0.9476\\
        SRFBN~\citep{li2019feedback} & $3\times$ &2.83 & 34.70 & 0.9292 & 30.51 & 0.8461 & 29.24 & 0.8084 & 28.73 & 0.8641 & 34.18 & 0.9481\\    
        \myrowcolour RCAN~\citep{zhang2018rcan}&	$3\times$&	15.63&	34.74&	0.9299&	30.65&	0.8482&	29.32&	0.8111&	29.09&	0.8702&	34.44&	0.9499\\
        SAN~\citep{dai2019SAN}&	$3\times$&	15.90&	34.75&	0.9300&	30.59&	0.8476&	29.33&	0.8112&	28.93&	0.8671&	34.30&	0.9494\\

        \myrowcolour HAN~\citep{niu2020HAN}&	$3\times$&	64.35&	34.75&	0.9299&	30.67&	0.8483&	29.32&	0.8110&	29.10&	0.8705&	34.48&	0.9500\\
        
            
        NLSA~\citep{mei2021NLSA}&	$3\times$&	45.58&	34.85&	0.9306&	30.70&	0.8485&	29.34&	0.8117&	29.25&	0.8726&	34.57&	0.9508\\
        \myrowcolour IPT~\citep{chen2021pre}&	$3\times$&	115.67&	34.81&	-&	30.85&	-&	29.38&	-&	29.49&	-&	-&	-\\ \hline
        
        SwinIR~\citep{liang2021swinir}&	$3\times$&	11.94&	34.97&	0.9318&	30.93&	0.8534&	29.46&	0.8145&	29.75&	0.8826&	35.12&	0.9537\\  
        \myrowcolour CAT-A~\citep{chen2022cross}      & $3\times$ &16.64 & 35.06 & 0.9326 & 31.04 & 0.8538 & 29.52 & 0.8160 & 30.12 & 0.8862 & 35.38 & 0.9546 \\
        ART~\citep{zhang2022accurate}		&$3\times$ &16.58	&35.07	&0.9325	&31.02	&0.8541	&29.51	&0.8159	&30.1	&0.8871	&35.39	&0.9548	\\		
        \myrowcolour EDT~\citep{li2021efficient}&	$3\times$&	11.66&	35.13&	0.9328&	31.09&	0.8553&	29.53&	0.8165&	30.07&	0.8863&	35.47&	0.9550\\ 
        GRL-B~\citep{li2023efficient} &	$3\times$&	20.24&	35.12&	0.9353&	31.27&	\sotab{0.8611}&	29.56&	0.8235&	30.92&	\sotab{0.8990}&	35.76&	0.9566\\ 
        \myrowcolour HAT~\citep{chen2023activating}	&$3\times$	&20.81	&35.16	&0.9335	&31.33	&0.8576	&29.59	&0.8177	&30.7	&0.8949	&35.84	&0.9567	\\
        Hi-IR-B (Ours)	&$3\times$& 14.87	&35.11	&\sotab{0.9372}	&31.37	&0.8598	&29.60	&\sotab{0.8240}	&30.79	&0.8977	&35.92	&\sotab{0.9583}			\\
        \myrowcolour HAT-L~\citep{chen2023activating}	&$3\times$	&40.88	&\sotaa{35.28}	&0.9345	&\sotab{31.47}	&0.8584	&\sotab{29.63}	&0.8191	&\sotab{30.92}	&0.8981	&\sotab{36.02}	&0.9576	\\
        Hi-IR-L (Ours)	&$3\times$& 39.26	&\sotab{35.20}	&\sotaa{0.9380}	&\sotaa{31.55}	&\sotaa{0.8616}	&\sotaa{29.67}	&\sotaa{0.8256}	&\sotaa{31.07}	&\sotaa{0.9020}	&\sotaa{36.12}	&\sotaa{0.9588}			\\ \midrule[0.1em]
        
        \myrowcolour EDSR~\citep{lim2017enhanced} & $4\times$ & 43.09 & 32.46 & 0.8968 & 28.80 & 0.7876 & 27.71 & 0.7420 & 26.64 & 0.8033 & 31.02 & 0.9148\\
        SRFBN~\citep{li2019feedback} & $4\times$ &3.63 & 32.47 & 0.8983 & 28.81 & 0.7868 & 27.72 & 0.7409 & 26.60 & 0.8015 & 31.15 & 0.9160\\

        \myrowcolour RCAN~\citep{zhang2018rcan}&	$4\times$&	15.59&	32.63&	0.9002&	28.87&	0.7889&	27.77&	0.7436&	26.82&	0.8087&	31.22&	0.9173\\
        SAN~\citep{dai2019SAN}&	$4\times$&	15.86&	32.64&	0.9003&	28.92&	0.7888&	27.78&	0.7436&	26.79&	0.8068&	31.18&	0.9169\\

        \myrowcolour HAN~\citep{niu2020HAN}&	$4\times$&	64.20&	32.64&	0.9002&	28.90&	0.7890&	27.80&	0.7442&	26.85&	0.8094&	31.42&	0.9177\\
        
            
        NLSA~\citep{mei2021NLSA}&	$4\times$&	44.99&	32.59&	0.9000&	28.87&	0.7891&	27.78&	0.7444&	26.96&	0.8109&	31.27&	0.9184\\
        \myrowcolour IPT~\citep{chen2021pre}&	$4\times$&	115.63&	32.64&	-&	29.01&	-&	27.82&	-&	27.26&	-&	-&	-\\ \hline
        
        SwinIR~\citep{liang2021swinir}&	$4\times$&	11.90&	32.92&	0.9044&	29.09&	0.7950&	27.92&	0.7489&	27.45&	0.8254&	32.03&	0.9260\\  
        \myrowcolour CAT-A~\citep{chen2022cross}      & $4\times$ &16.60 & {33.08} & 0.9052 & 29.18 & 0.7960 & 27.99 & 0.7510 & 27.89 & 0.8339 & 32.39 & 0.9285 \\
        ART~\citep{zhang2022accurate}	&$4\times$	& 16.55 &33.04	&0.9051	&29.16	&0.7958	&27.97	&0.7510	&27.77	&0.8321	&32.31	&0.9283	\\		
        \myrowcolour EDT~\citep{li2021efficient}&	$4\times$&	11.63&	 {33.06}&	0.9055&	 {29.23}&	0.7971&	 {27.99}&	0.7510&	27.75&	0.8317&	 {32.39}&	 {0.9283}\\ 
        GRL-B~\citep{li2023efficient} &	$4\times$&	20.20&	 {33.10}&	 {0.9094}&	 {29.37}&	 {0.8058}&	 {28.01}&	 \sotab{0.7611}&	 {28.53}&	 \sotab{0.8504}&	 {32.77}&	 {0.9325}\\ 
        \myrowcolour HAT~\citep{chen2023activating}	&$4\times$	&20.77	&33.18	&0.9073	&29.38	&0.8001	&28.05	&0.7534	&28.37	&0.8447	&32.87	&0.9319	\\
        Hi-IR-B (Ours)	&$4\times$& 14.83	&33.14	&\sotab{0.9095}	&29.40	&\sotab{0.8029}	&28.08	&\sotab{0.7611}	&28.44	&0.8448	&32.90	&0.9323			\\
        \myrowcolour HAT-L~\citep{chen2023activating}	&$4\times$	&40.85	&\sotaa{33.30}	&0.9083	&\sotab{29.47}	&0.8015	&\sotab{28.09}	&0.7551	&\sotab{28.60}	&0.8498	&\sotab{33.09}	&\sotab{0.9335}	\\
        Hi-IR-L (Ours)	&$4\times$& 39.22	&\sotab{33.22}	&\sotaa{0.9103}	&\sotaa{29.49}	&\sotaa{0.8041}	&\sotaa{28.13}	&\sotaa{0.7622}	&\sotaa{28.72}	&\sotaa{0.8514}	&\sotaa{33.13}	&\sotaa{0.9366}				\\\bottomrule[0.1em]
        \end{tabular}}
\end{center}
\end{table}

%% file: figs/visual_sr_main.tex
\begin{figure*}[!tb]
\begin{center}
\scriptsize
\begin{tabular}[b]{c@{ } c@{ } c@{ } c@{ }} 

\hspace{-1mm}  
    \includegraphics[width=0.24\linewidth,valign=t]{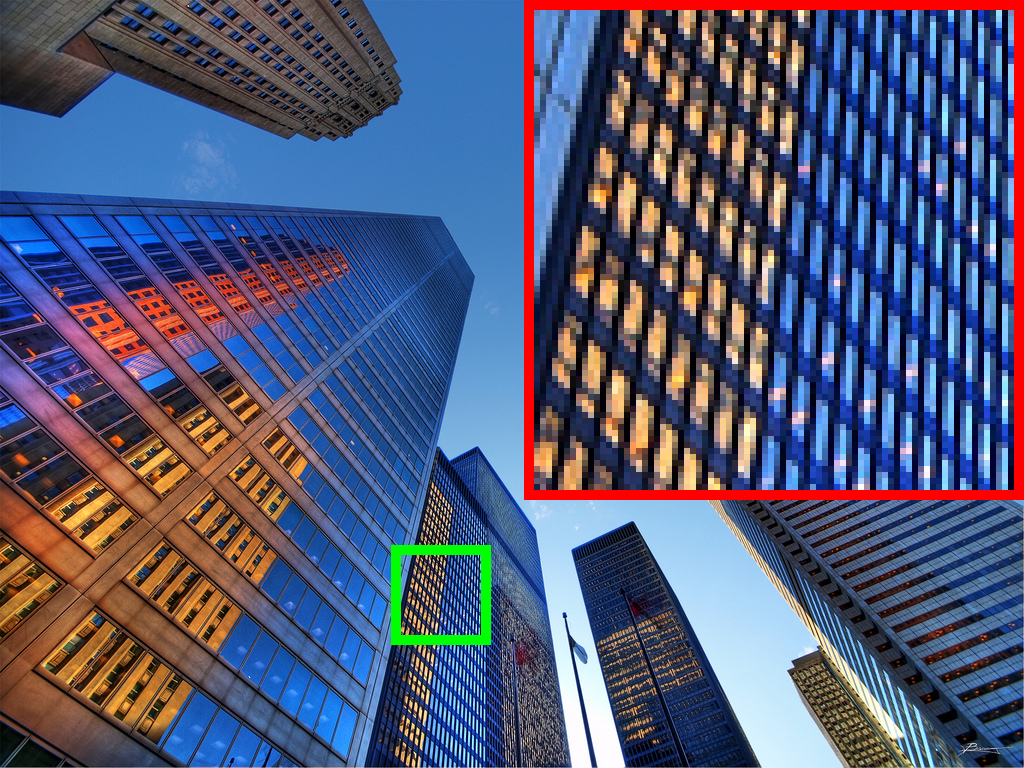} &
    \includegraphics[width=0.24\linewidth,valign=t]{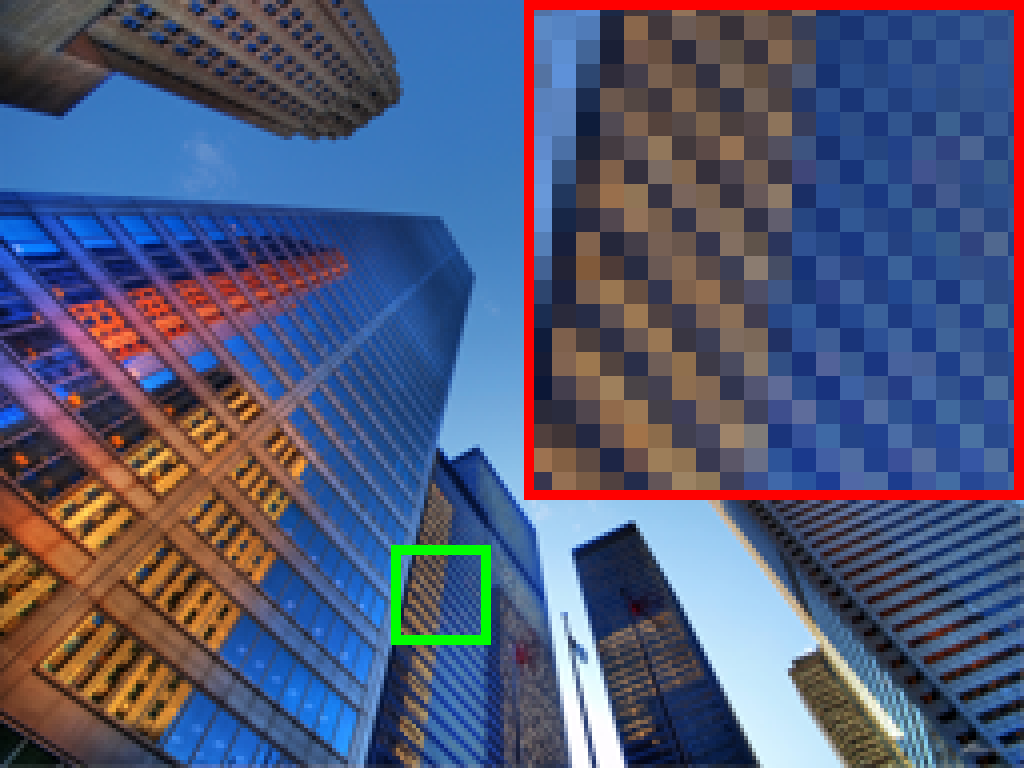} &
    \includegraphics[width=0.24\linewidth,valign=t]{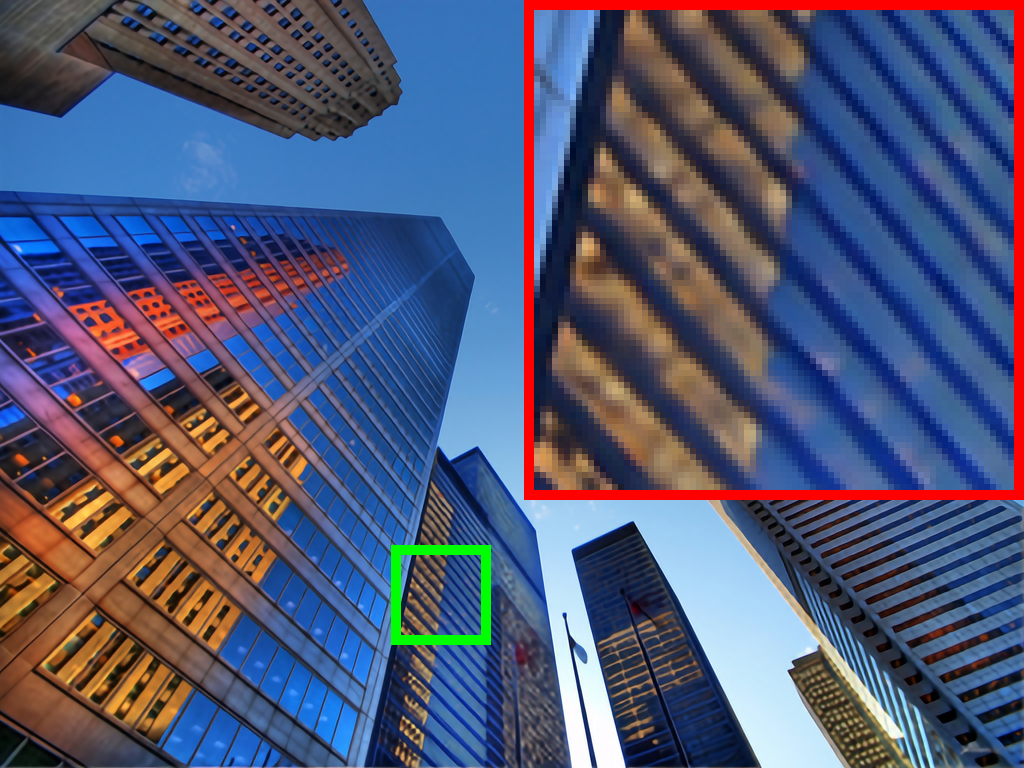} & 
    \includegraphics[width=0.24\linewidth,valign=t]{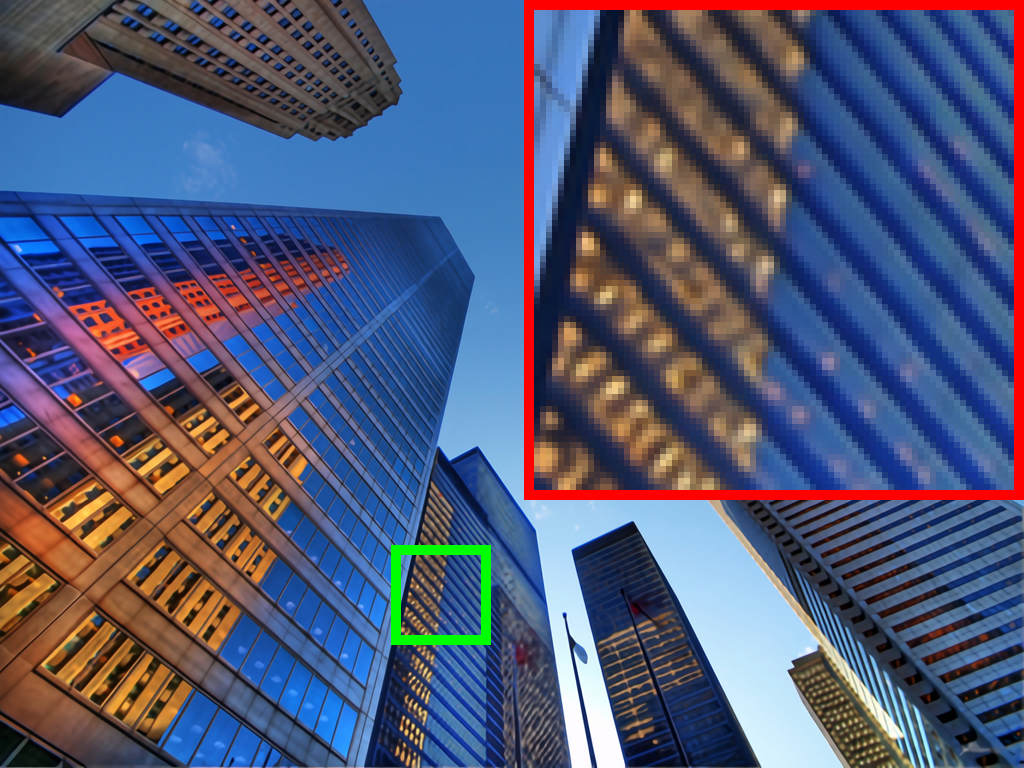} \\
    GT & LR & IPT~\citep{chen2021pre} & SwinIR~\citep{liang2021swinir}  \\ 
\hspace{-1mm}  
    \includegraphics[width=0.24\linewidth,valign=t]{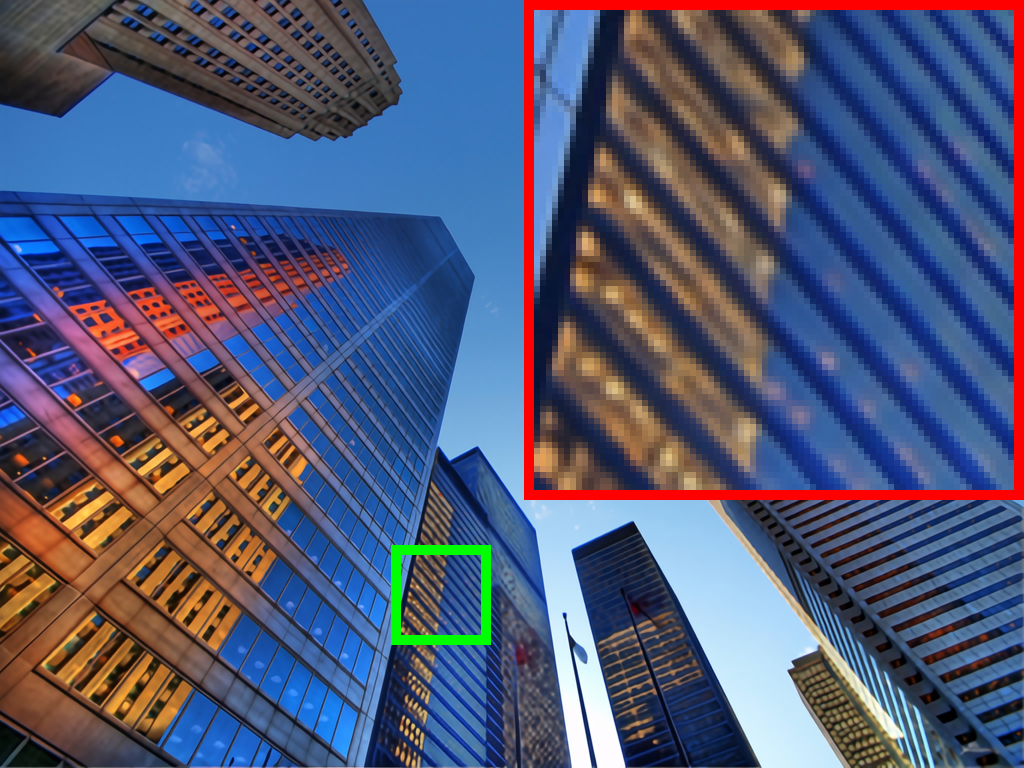} &
    \includegraphics[width=0.24\linewidth,valign=t]{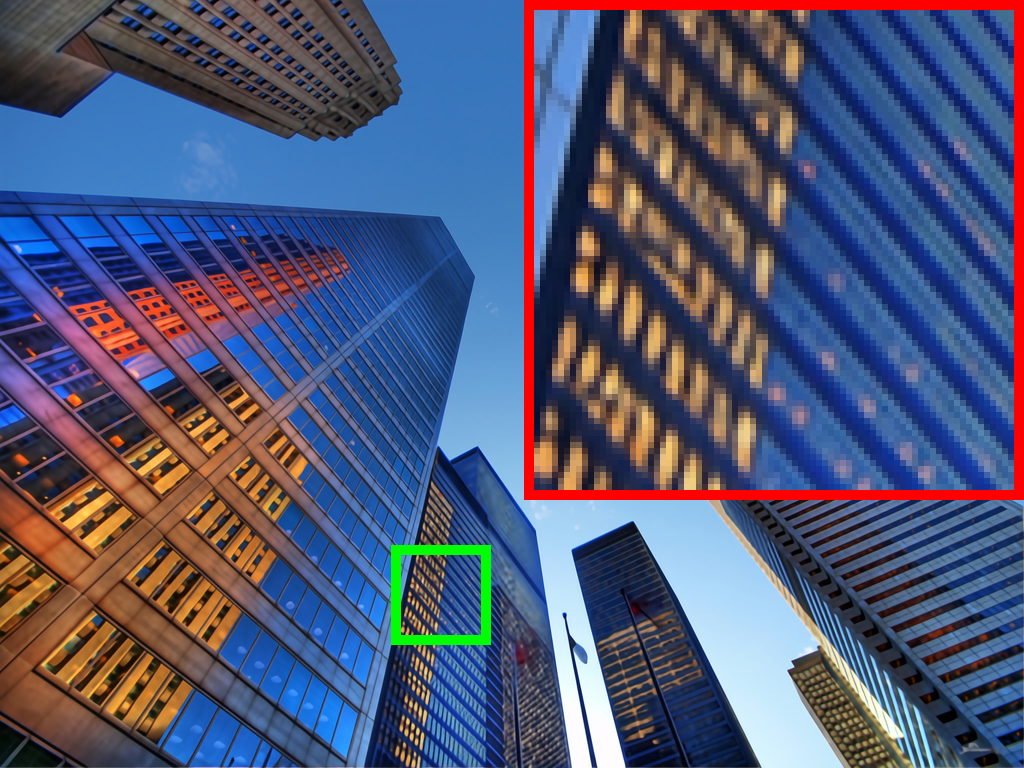} &
    \includegraphics[width=0.24\linewidth,valign=t]{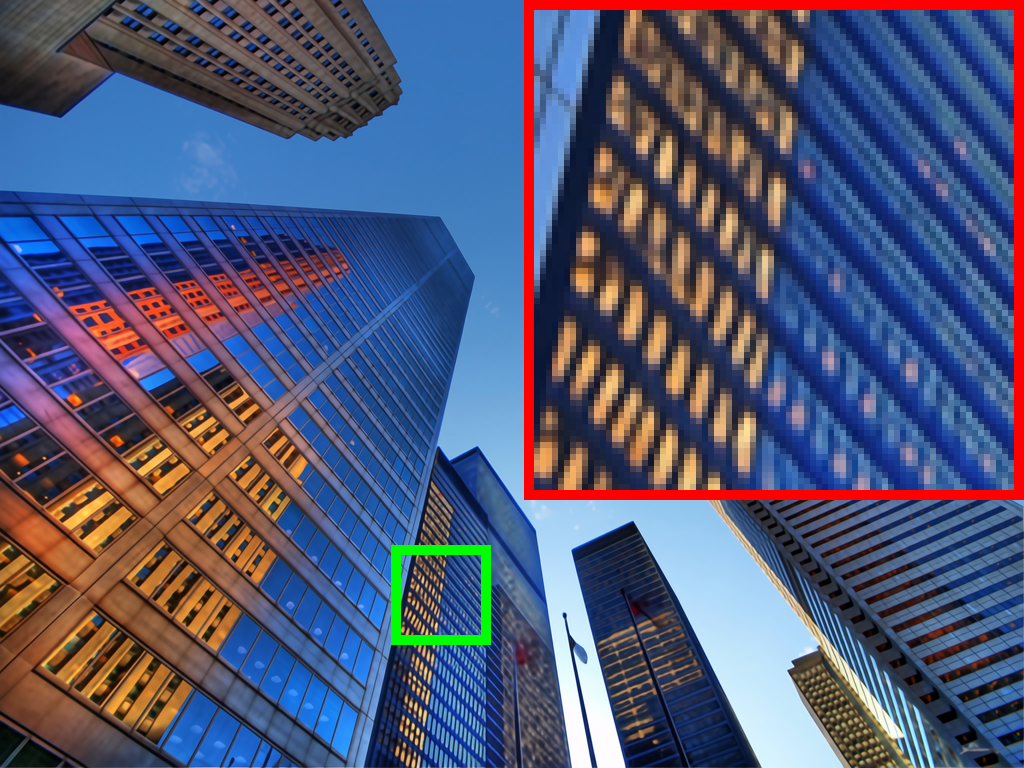} &
    \includegraphics[width=0.24\linewidth,valign=t]{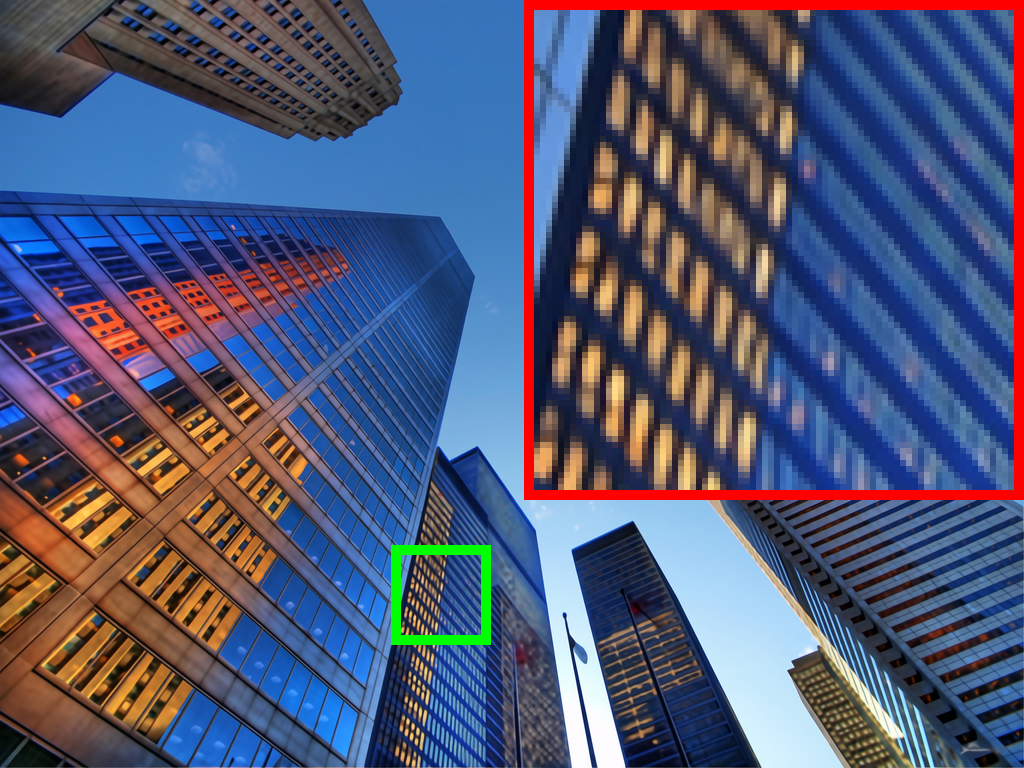} \\ 
    EDT~\citep{li2021efficient} & GRL~\citep{li2023efficient} & HAT~\citep{chen2023activating} & Hi-IR (Ours) \\ \\    

\hspace{-1mm}  
    \includegraphics[width=0.24\linewidth,valign=t]{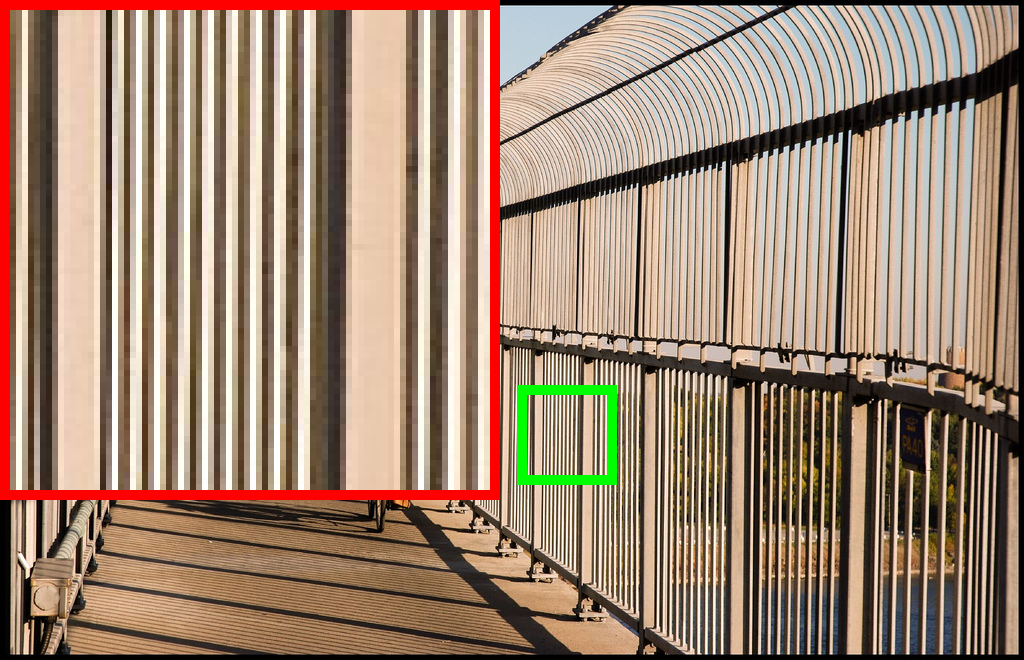} &
    \includegraphics[width=0.24\linewidth,valign=t]{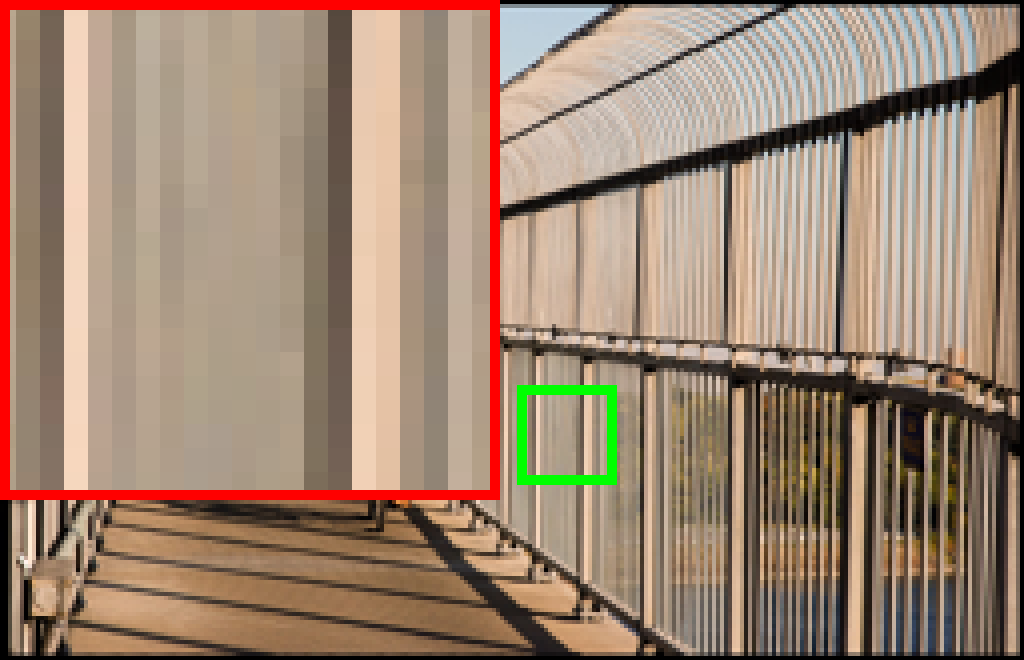} &
    \includegraphics[width=0.24\linewidth,valign=t]{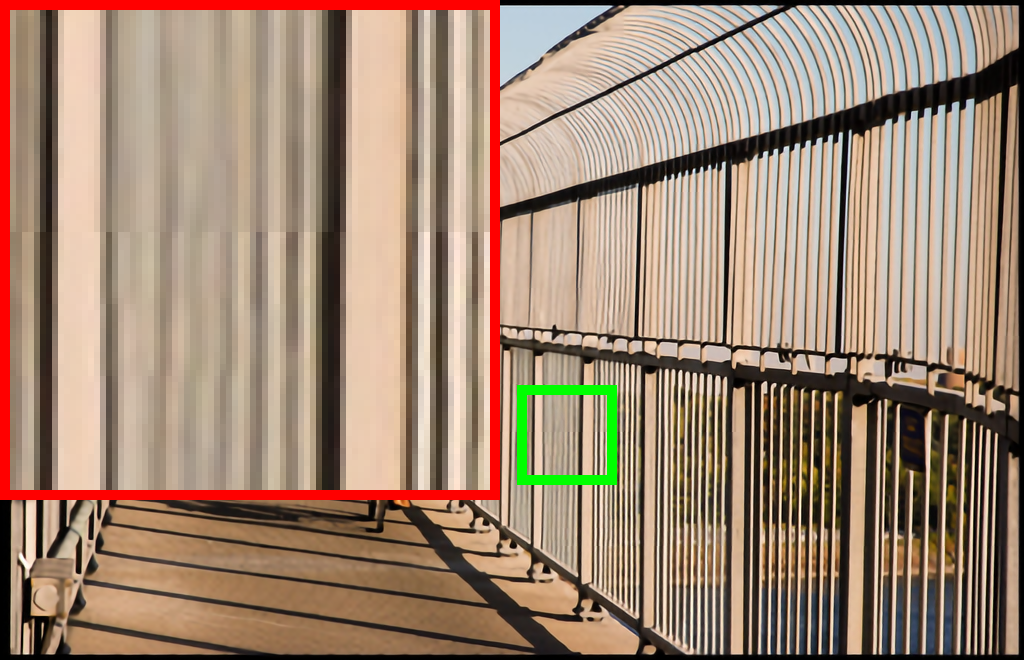} & 
    \includegraphics[width=0.24\linewidth,valign=t]{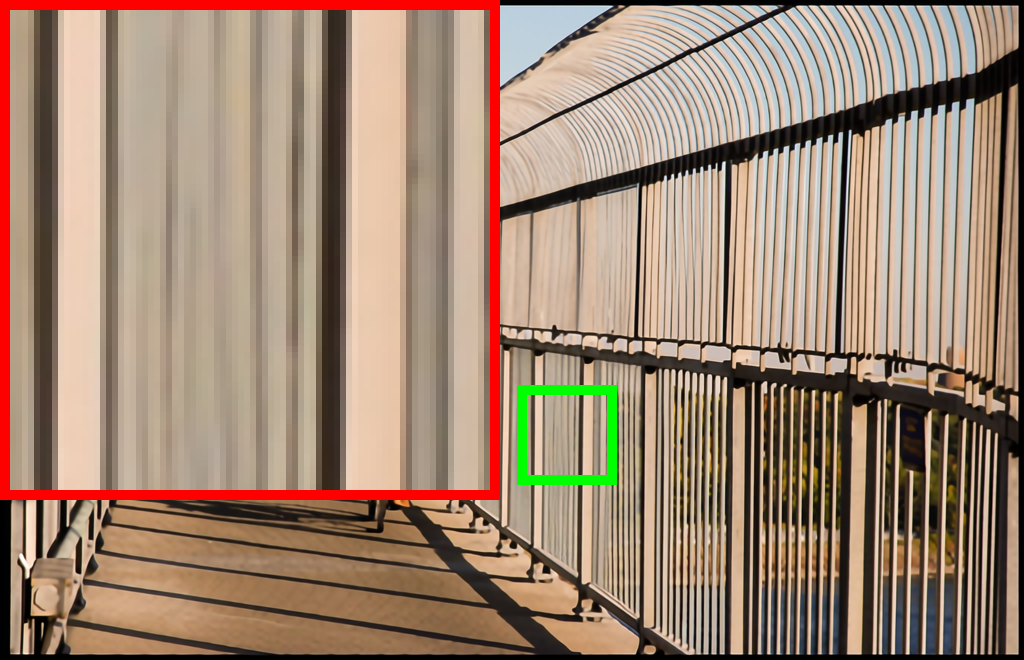} \\ 
    GT & LR & IPT~\citep{chen2021pre} & SwinIR~\citep{liang2021swinir}  \\ 
\hspace{-1mm}  
    \includegraphics[width=0.24\linewidth,valign=t]{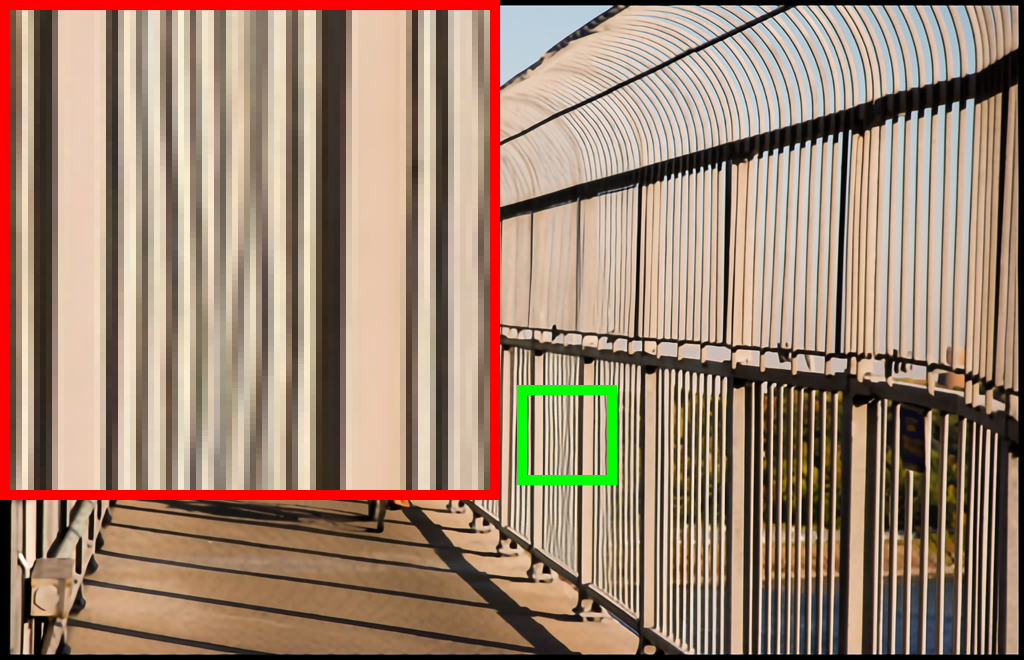} &
    \includegraphics[width=0.24\linewidth,valign=t]{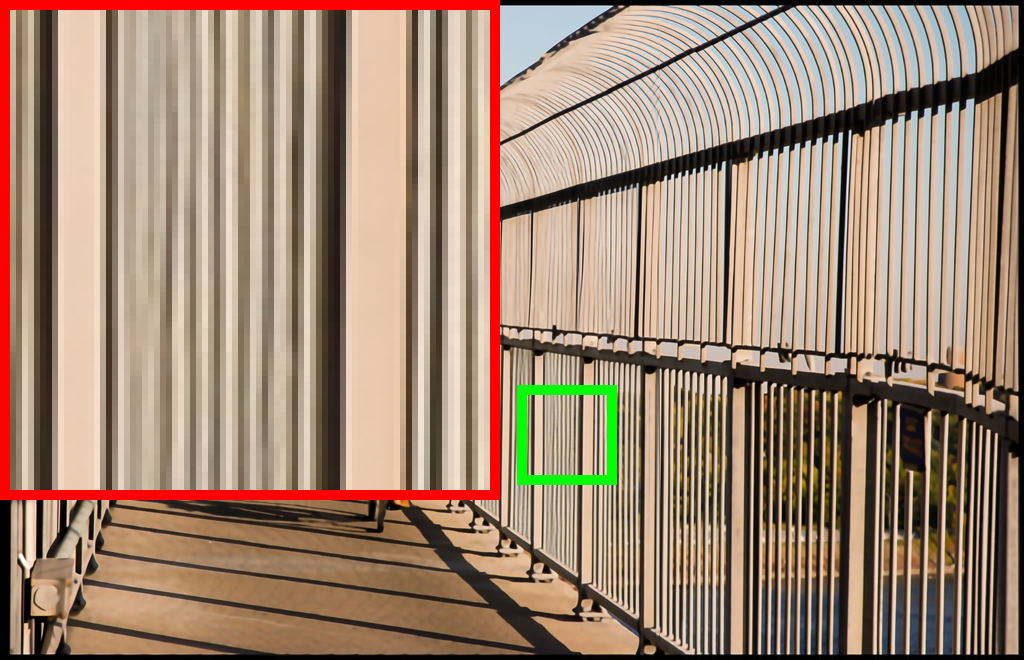} &
    \includegraphics[width=0.24\linewidth,valign=t]{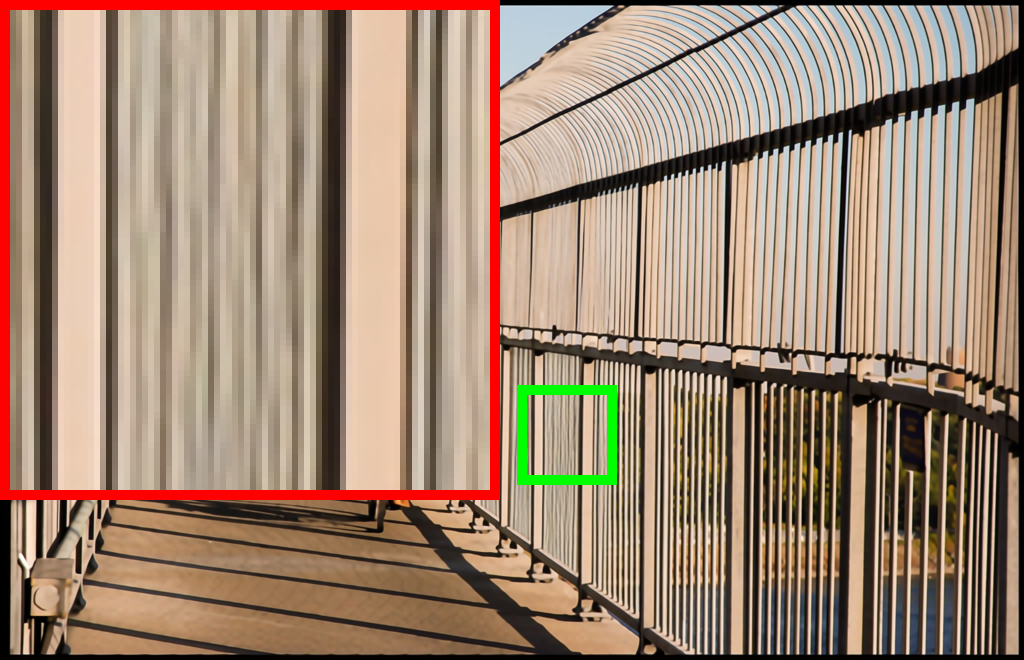} &
    \includegraphics[width=0.24\linewidth,valign=t]{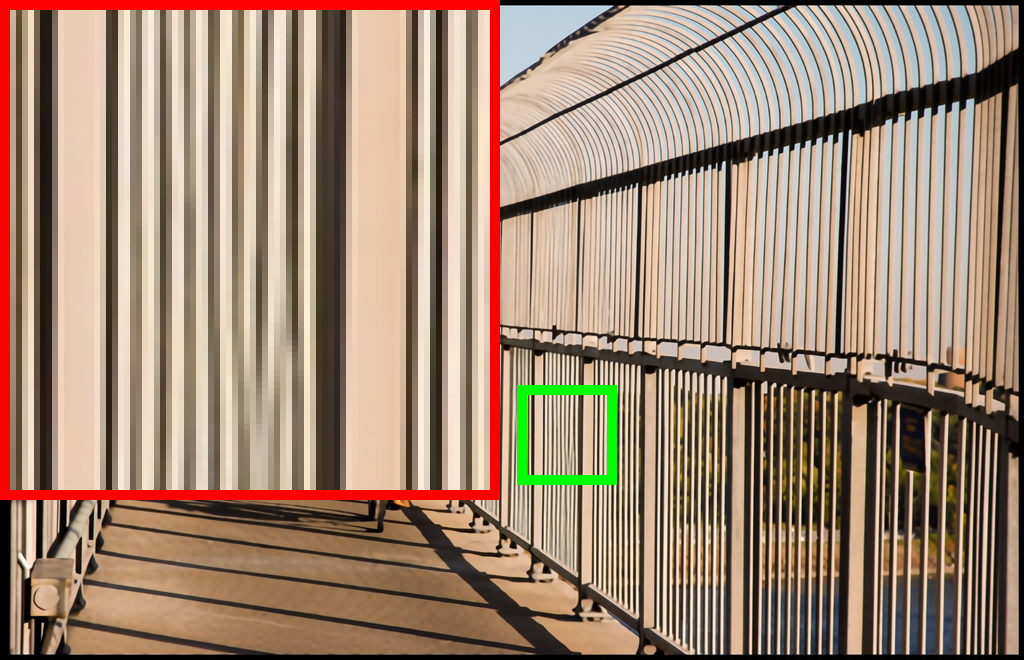} \\ 
    EDT~\citep{li2021efficient} & GRL~\citep{li2023efficient} & HAT~\citep{chen2023activating} & Hi-IR (Ours) \\ \\    
    
\end{tabular}
\end{center}
\vspace{-4mm}
\caption{Visual results for classical image $\times 4$ SR on Urban100 dataset.}
\label{fig:visual_sr_urban100}
\end{figure*}

%% file: tables/dn_table.tex
\begin{table*}[!htb]
\centering
\caption{\textit{\textbf{Color and grayscale image denoising}} results. 
}
\label{tab:denoising}
\setlength{\extrarowheight}{0.7pt}
\setlength{\tabcolsep}{1.2pt}
\scalebox{0.65}{
    \begin{tabular}{l | c |  c c c | c c c | c c c || c c c | c c c  }
    \toprule[0.1em]
    \multirow{3}{*}{\textbf{Method}} & \multicolumn{1}{c|}{\multirow{3}{*}{ \makecell{\textbf{Params} \\ \textbf{[M]}} }}
     & \multicolumn{9}{c||}{\textbf{Color}} & \multicolumn{6}{c}{\textbf{Grayscale}} \\ \cline{3-17}
    & 
    & \multicolumn{3}{c|}{\textbf{Kodak24}} & \multicolumn{3}{c|}{\textbf{McMaster}} & \multicolumn{3}{c||}{\textbf{Urban100}}  & \multicolumn{3}{c|}{\textbf{Set12}}  & \multicolumn{3}{c}{\textbf{Urban100}} \\
            &  & $\sigma$$=$$15$ & $\sigma$$=$$25$ & $\sigma$$=$$50$ & $\sigma$$=$$15$ & $\sigma$$=$$25$ & $\sigma$$=$$50$ & $\sigma$$=$$15$ & $\sigma$$=$$25$ & $\sigma$$=$$50$ & $\sigma$$=$$15$ & $\sigma$$=$$25$ & $\sigma$$=$$50$ 
            & $\sigma$$=$$15$ & $\sigma$$=$$25$ & $\sigma$$=$$50$ \\ \midrule
    \myrowcolour DnCNN~\citep{kiku2016beyond}	&0.56	
    &34.60	&32.14	&28.95	&33.45	&31.52	&28.62	&32.98	&30.81	&27.59				&32.86	&30.44	&27.18				&32.64	&29.95	&26.26	\\		
    RNAN~\citep{zhang2019residual}	&8.96	
    &-	&-	&29.58	&-	&-	&29.72	&-	&-	&29.08				&-	&-	&27.70				&-	&-	&27.65	\\		
    \myrowcolour IPT~\citep{chen2021pre}	&115.33	
    &-	&-	&29.64	&-	&-	&29.98	&-	&-	&29.71				&-	&-	&-				&-	&-	&-	\\		
    EDT-B~\citep{li2021efficient}	&11.48	
    &35.37	&32.94	&29.87	&35.61	&33.34	&30.25	&35.22	&33.07	&30.16				&-	&-	&-				&-	&-	&-	\\		
    \myrowcolour DRUNet~\citep{zhang2021plug}	&32.64	
    &35.31	&32.89	&29.86	&35.40	&33.14	&30.08	&34.81	&32.60	&29.61				&33.25	&30.94	&27.90				&33.44	&31.11	&27.96	\\		
    SwinIR~\citep{liang2021swinir}	&11.75	
    &35.34	&32.89	&29.79	&35.61	&33.20	&30.22	&35.13	&32.90	&29.82				&33.36	&31.01	&27.91				&33.70	&31.30	&27.98	\\		
    \myrowcolour Restormer~\citep{zamir2022restormer}	&26.13	
    &\sotaa{35.47}	&\sotaa{33.04}	&\sotaa{30.01}	&35.61	&\sotab{33.34}	&30.30	&35.13	&32.96	&30.02				&33.42	&31.08	&28.00				&33.79	&31.46	&28.29	\\		
    Xformer~\citep{zhang2023xformer}	& 25.23	
    &{35.39}	&{32.99}	&{29.94}	&\sotab{35.68}	&\sotaa{33.44}	&\sotab{30.38}	&\sotab{35.29}	&\sotab{33.21}	&\sotab{30.36}	&\sotab{33.46}	&\sotab{31.16}	&\sotab{28.10}	&\sotab{33.98}	&\sotab{31.78}	&\sotab{28.71}	\\		
    \myrowcolour Hi-IR (Ours)	& 22.33	
    &\sotab{35.42}	&\sotab{33.01}	&\sotab{29.98}	&\sotaa{35.69}	&\sotaa{33.44}	&\sotaa{30.42}	&\sotaa{35.46}	&\sotaa{33.34}	&\sotaa{30.59}	&\sotaa{33.48}	&\sotaa{31.19}	&\sotaa{28.15}	&\sotaa{34.11}	&\sotaa{31.92}	&\sotaa{28.91}	\\		
    \bottomrule[0.1em]
    \end{tabular}
}
\vspace{-2mm}
\end{table*}

%% file: tables/jpeg_table_grayscale.tex
\begin{table}[t]
\centering
\caption{\textit{\textbf{Grayscale image JPEG compression artifact removal}} results. 
\textcolor{magenta}{\textdagger}A single model is trained to handle multiple noise levels.} 
\label{tab:jpeg_car_grapyscale}
\setlength{\extrarowheight}{0.7pt}
\setlength{\tabcolsep}{2.4pt}
\scalebox{0.65}{
    \begin{tabular}{c | c| c c | c c | c c | c c || c c | c c | c c | c c    }
    \toprule[0.1em]
    \multirow{2}{*}{Set} & \multirow{2}{*}{QF} & \multicolumn{2}{c|}{JPEG}  & \multicolumn{2}{c|}{\makecell{\textcolor{magenta}{\textdagger}DnCNN3}} & \multicolumn{2}{c|}{\makecell{\textcolor{magenta}{\textdagger}DRUNet}} & \multicolumn{2}{c||}{\textcolor{magenta}{\textdagger}Hi-IR (Ours)} 
    & \multicolumn{2}{c|}{\makecell{SwinIR}} & \multicolumn{2}{c|}{\makecell{ART}} & \multicolumn{2}{c|}{CAT}  & \multicolumn{2}{c}{Hi-IR (Ours)}  \\ \cline{3-18}
    & & PSNR$\uparrow$ & SSIM$\uparrow$ & PSNR$\uparrow$ & SSIM$\uparrow$ & PSNR$\uparrow$ & SSIM$\uparrow$ & PSNR$\uparrow$ & SSIM$\uparrow$ & PSNR$\uparrow$ & SSIM$\uparrow$ & PSNR$\uparrow$ & SSIM$\uparrow$ & PSNR$\uparrow$ & SSIM$\uparrow$ & PSNR$\uparrow$ & SSIM$\uparrow$ \\
    \midrule[0.1em]
    {\multirow{4}{*}{\rotatebox[origin=c]{90}{\makecell{Classic5}}}}
    	&10	&27.82	&0.7600					&29.40	&0.8030											&\sotab{30.16}	&\sotab{0.8234}		&\sotaa{30.25}	&\sotaa{0.8236}		
     &\sotab{30.27}	&0.8249		&\sotab{30.27}	&\sotab{0.8258}		&30.26	&\sotaa{0.8250}					&{30.38}	&\sotaa{0.8266}		\\											
    	  &20	&30.12	&0.8340					&31.63	&0.8610											&\sotab{32.39}	&\sotab{0.8734}		&\sotaa{32.51}	&\sotaa{0.8737}		
     &32.52	&0.8748		& -	&	-	&\sotab{32.57}	&\sotaa{0.8754}					&\sotaa{32.62}	&\sotab{0.8751}		\\											
    	&30	&31.48	&0.8670					&32.91	&0.8860											&\sotab{33.59}	&\sotab{0.8949}		&\sotaa{33.74}	&\sotaa{0.8954}		
     &33.73	&0.8961		&33.74	&\sotaa{0.8964}		&\sotab{33.77}	&\sotaa{0.8964}					&\sotaa{33.80}	&\sotab{0.8962}		\\											
    	  &40	&32.43	&0.8850					&33.77	&0.9000											&\sotab{34.41}	&\sotab{0.9075}		&\sotaa{34.55}	&\sotaa{0.9078}		
     &34.52	&0.9082		&34.55	&\sotab{0.9086}		&{34.58}	&\sotaa{0.9087}					&\sotaa{34.61}	&{0.9082}		\\											
     \hline									
    {\multirow{4}{*}{\rotatebox[origin=c]{90}{\makecell{LIVE1}}}}	
    	&10	&27.77	&0.7730					&29.19	&0.8120											&\sotab{29.79}	&\sotab{0.8278}		&\sotaa{29.84}	&\sotaa{0.8328}		
     &29.86	&0.8287		&\sotab{29.89}	&\sotab{0.8300}		&\sotab{29.89}	&0.8295					&\sotaa{29.94}	&\sotaa{0.8359}		\\											
    	  &20	&30.07	&0.8510					&31.59	&0.8800											&\sotab{32.17}	&\sotab{0.8899}		&\sotaa{32.24}	&\sotaa{0.8926}		
     &32.25	&0.8909		&-	&	-	&\sotab{32.30}	&\sotab{0.8913}					&\sotaa{32.31}	&\sotaa{0.8938}		\\											
    	&30	&31.41	&0.8850					&32.98	&0.9090											&\sotab{33.59}	&\sotab{0.9166}		&\sotaa{33.67}	&\sotaa{0.9192}		
     &33.69	&0.9174		&\sotab{33.71}	&\sotab{0.9178}		&\sotaa{33.73}	&0.9177					&\sotaa{33.73}	&\sotaa{0.9223}		\\											
    	  &40	&32.35	&0.9040					&33.96	&0.9250											&\sotab{34.58}	&\sotab{0.9312}		&\sotaa{34.66}	&\sotaa{0.9347}		
     &34.67	&0.9317		&34.70	&\sotab{0.9322}		&\sotaa{34.72}	&0.9320					&\sotab{34.71}	&\sotaa{0.9347}		\\											
    \hline											
    {\multirow{4}{*}{\rotatebox[origin=c]{90}{\makecell{Urban100}}}}	
    	&10	&26.33	&0.7816					&28.54	&0.8484											&\sotab{30.31}	&\sotab{0.8745}		&\sotaa{30.62}	&\sotaa{0.8808}		
     &30.55	&0.8835		&\sotab{30.87}	&\sotab{0.8894}		&30.81	&0.8866					&\sotaa{31.07}	&\sotaa{0.8950}		\\											
    	  &20	&28.57	&0.8545					&31.01	&0.9050											&\sotab{32.81}	&\sotab{0.9241}		&\sotaa{33.21}	&\sotaa{0.9256}		
     &33.12	&0.9190		&	-&-		&33.38	&\sotaa{0.9269}					&\sotaa{33.51}	&\sotab{0.9250}		\\											
    	&30	&30.00	&0.9013					&32.47	&0.9312											&\sotab{34.23}	&\sotab{0.9414}		&\sotaa{34.64}	&\sotaa{0.9478}		
     &34.58	&0.9417		&\sotab{34.81}	&0.9442		&\sotab{34.81}	&\sotab{0.9449}					&\sotaa{34.86}	&\sotaa{0.9459}		\\											
    	  &40	&31.06	&0.9215					&33.49	&0.9412											&\sotab{35.20}	&\sotab{0.9547}		&\sotaa{35.63}	&\sotaa{0.9566}		
     &35.50	&0.9515		&\sotab{35.73}	&\sotab{0.9553}		&\sotab{35.73}	&0.9511					&\sotaa{35.77}	&\sotaa{0.9561}		\\											
     \bottomrule[0.1em]
    \end{tabular}
}
\vspace{-3mm}
\end{table}

%% file: tables/jpeg_table_color.tex
\begin{table}[t]
    \centering
    \caption{\textit{\textbf{Color image JPEG compression artifact removal} results.}}
    \label{tab:jpeg_car_color}
    \setlength{\extrarowheight}{0.7pt}
    \setlength{\tabcolsep}{2.4pt}
    \scalebox{0.66}{
    \begin{tabular}{c | c | c c | c c | c c | c c | c c || c c | c c | c c}
    \toprule[0.1em]
    \multirow{2}{*}{Set} & \multirow{2}{*}{QF} & \multicolumn{2}{c|}{JPEG}  & \multicolumn{2}{c|}{\makecell{\textcolor{magenta}{\textdagger}QGAC}} & \multicolumn{2}{c|}{\makecell{\textcolor{magenta}{\textdagger}FBCNN}} & \multicolumn{2}{c|}{\textcolor{magenta}{\textdagger}DRUNet} & \multicolumn{2}{c||}{\textcolor{magenta}{\textdagger}Hi-IR (Ours)} & \multicolumn{2}{c|}{SwinIR} & \multicolumn{2}{c|}{GRL-S} & \multicolumn{2}{c}{Hi-IR (Ours)} \\ \cline{3-18}
    & & PSNR & SSIM & PSNR & SSIM & PSNR & SSIM & PSNR & SSIM & PSNR & SSIM & PSNR & SSIM & PSNR & SSIM & PSNR & SSIM  \\
    \midrule[0.1em]
    {\multirow{4}{*}{\rotatebox[origin=c]{90}{\makecell{LIVE1}}}}
    &10	&25.69	&0.7430		&27.62	&0.8040		&27.77	&0.8030		&\sotab{27.47}	&\sotab{0.8045}		&\sotaa{28.24}	&\sotaa{0.8149}		&28.06	&0.8129		&\sotab{28.13}	&\sotab{0.8139}		&\sotaa{28.36}	&\sotaa{0.8180}		\\
    &20	&28.06	&0.8260		&29.88	&0.8680		&30.11	&0.8680		&\sotab{30.29}	&\sotab{0.8743}		&\sotaa{30.59}	&\sotaa{0.8786}		&30.44	&0.8768		&\sotab{30.49}	&\sotab{0.8776}		&\sotaa{30.66}	&\sotaa{0.8797}		\\
    &30	&29.37	&0.8610		&31.17	&0.8960		&31.43	&0.8970		&\sotab{31.64}	&\sotab{0.9020}		&\sotaa{31.95}	&\sotaa{0.9055}		&31.81	&0.9040		&\sotab{31.85}	&\sotab{0.9045}		&\sotaa{32.02}	&\sotaa{0.9063}		\\
    &40	&30.28	&0.8820		&32.05	&0.9120		&32.34	&0.9130		&\sotab{32.56}	&\sotab{0.9174}		&\sotaa{32.88}	&\sotaa{0.9205}		&32.75	&0.9193		&\sotab{32.79}	&\sotab{0.9195}		&\sotaa{32.94}	&\sotaa{0.9210}		\\
    \midrule[0.1em]
    {\multirow{4}{*}{\rotatebox[origin=c]{90}{\makecell{BSD500}}}}
    &10	&25.84	&0.7410		&27.74	&\sotab{0.8020}		&\sotab{27.85}	&0.7990		&27.62	&0.8001		&\sotaa{28.26}	&\sotaa{0.8070}		&28.22	&0.8075		&\sotab{28.26}	&\sotab{0.8083}		&\sotaa{28.35}	&\sotaa{0.8092}		\\
    &20	&28.21	&0.8270		&30.01	&0.8690		&30.14	&0.8670		&\sotab{30.39}	&\sotab{0.8711}		&\sotaa{30.58}	&\sotaa{0.8741}		&30.54	&0.8739		&\sotab{30.57}	&\sotab{0.8746}		&\sotaa{30.61}	&\sotaa{0.8740}		\\
    &30	&29.57	&0.8650		&31.330	&0.8980		&31.45	&0.8970		&\sotab{31.73}	&\sotab{0.9003}		&\sotaa{31.93}	&\sotaa{0.9029}		&31.90	&0.9025		&\sotab{31.92}	&\sotab{0.9030}		&\sotaa{31.99}	&\sotaa{0.9035}		\\
    &40	&30.52	&0.8870		&32.25	&0.9150		&32.36	&0.9130		&\sotab{32.66}	&\sotab{0.9168}		&\sotaa{32.87}	&\sotaa{0.9193}		&32.84	&0.9189		&\sotab{32.86}	&\sotab{0.9192}		&\sotaa{32.92}	&\sotaa{0.9195}		\\
    \hline
    {\multirow{4}{*}{\rotatebox[origin=c]{90}{\makecell{Urban100}}}}
    &10	&24.46	&0.7612		&-	&-		&-	&-		&\sotab{27.10}	&\sotab{0.8400}		&\sotaa{28.78}	&\sotaa{0.8666}		&28.18	&0.8586		&\sotab{28.54}	&\sotab{0.8635}		&\sotaa{29.11}	&\sotaa{0.8727}		\\
    &20	&26.63	&0.8310		&-	&-		&-	&-		&\sotab{30.17}	&\sotab{0.8991}		&\sotaa{31.12}	&\sotaa{0.9087}		&30.53	&0.9030		&\sotab{30.93}	&\sotab{0.9067}		&\sotaa{31.36}	&\sotaa{0.9115}		\\
    &30	&27.96	&0.8640		&-	&-		&-	&-		&\sotab{31.49}	&\sotab{0.9189}		&\sotaa{32.42}	&\sotaa{0.9265}		&31.87	&0.9219		&\sotab{32.24}	&\sotab{0.9247}		&\sotaa{32.57}	&\sotaa{0.9279}		\\
    &40	&28.93	&0.8825		&-	&-		&-	&-		&\sotab{32.36}	&\sotab{0.9301}		&\sotaa{33.26}	&\sotaa{0.9363}		&32.75	&0.9329		&\sotab{33.09}	&\sotab{0.9348}		&\sotaa{33.37}	&\sotaa{0.9373}		\\
    \bottomrule[0.1em]
    \end{tabular}}
\end{table}

%% file: tables/motion_deblur_gopro_hide.tex
\begin{table}[!t]
\parbox{0.48\linewidth}{\centering
\caption{\textit{\textbf{Single-image motion deblurring}} on GoPro and HIDE dataset. {GoPro} dataset is used for training. 
}
\label{tab:motion_deblurring}
\setlength{\extrarowheight}{0.7pt}
\setlength{\tabcolsep}{0.7pt}
\scalebox{0.55}{
    \begin{tabular}{l | c | c | c}
    \toprule[0.1em]
     & \textbf{GoPro} & \textbf{HIDE}  & Average \\
     \textbf{Method} & PSNR$\uparrow$ / SSIM$\uparrow$ & PSNR$\uparrow$ / SSIM$\uparrow$ & PSNR$\uparrow$ / SSIM$\uparrow$ \\
    \midrule[0.1em]
    DeblurGAN-v2~\citep{deblurganv2}	&29.55 / 0.934		&26.61 / 0.875		&28.08 / 0.905		\\
    \myrowcolour SRN~\citep{tao2018scale}	&30.26 / 0.934		&28.36 / 0.915		&29.31 / 0.925		\\
    SPAIR~\citep{purohit2021spatially_spair}	&32.06 / 0.953		&30.29 / 0.931		&31.18 / 0.942		\\
    \myrowcolour MIMO-UNet+~\citep{cho2021rethinking_mimo}&32.45 / 0.957		&29.99 / 0.930		&31.22 / 0.944		\\
    MPRNet~\citep{zamir2021multi}	&32.66 / 0.959		&30.96 / 0.939		&31.81 / 0.949		\\
    \myrowcolour MAXIM-3S~\citep{tu2022maxim}	&32.86 / 0.961 & 32.83 / 0.956 & 32.85 / 0.959	\\
    Restormer~\citep{zamir2022restormer}	&32.92 / {0.961}		&31.22 / 0.942	&32.07 / 0.952
    		\\
    \myrowcolour Stripformer~\citep{tsai2022stripformer} & 33.08 / {0.962} & 31.03 / 0.940 &32.06 / 0.951 \\
    ShuffleFormer~\citep{xiao2023random} &	33.38 / \sotab{0.965}	& 31.25 / 0.943 & 31.32 / 0.954 \\
    
    \myrowcolour GRL-B~\citep{li2023efficient}	&\sotab{33.93} / \sotaa{0.968}		&\sotaa{31.65} / \sotaa{0.947}		&\sotab{32.79} / \sotaa{0.958}	\\
    Hi-IR-L (Ours)  &\sotaa{33.99} / \sotaa{0.968}		&\sotab{31.64} / \sotaa{0.947}		&\sotaa{32.82} / \sotaa{0.958}	\\
    \bottomrule[0.1em]
    \end{tabular}}
}
\hspace{4pt}
\parbox{0.48\linewidth}{
\centering
\caption{\textit{\textbf{Single image motion deblurring} on RealBlur dataset.} \textcolor{violet}{\textdagger}: Methods trained on RealBlur.}
\label{tab:motion_deblurring_realblur}
\setlength{\extrarowheight}{0.7pt}
\setlength{\tabcolsep}{0.7pt}
\scalebox{0.55}{
    \begin{tabular}{l | c | c  |c}
    \toprule[0.1em]
     & {\textbf{RealBlur-R}} & {\textbf{RealBlur-J}} & Average\\
     \textbf{Method} & PSNR$\uparrow$ / SSIM$\uparrow$ & PSNR$\uparrow$ / SSIM$\uparrow$ & PSNR$\uparrow$ / SSIM$\uparrow$ \\
    \midrule[0.1em]

    

    						
    						

    \textcolor{violet}{\textdagger}DeblurGAN-v2
    &36.44 / 0.935		&29.69 / 0.870		&33.07 / 0.903	\\
    \myrowcolour \textcolor{violet}{\textdagger}SRN~\citep{tao2018scale}&38.65 / 0.965		&31.38 / 0.909		&35.02 / 0.937	\\
    \textcolor{violet}{\textdagger}MPRNet~\citep{zamir2021multi}	&39.31 / 0.972		&31.76 / 0.922		&35.54 / 0.947	\\
    \myrowcolour \textcolor{violet}{\textdagger}MIMO-UNet+~\citep{cho2021rethinking_mimo}	&- / -		&32.05 / 0.921		& - / -	\\
    \textcolor{violet}{\textdagger}MAXIM-3S~\citep{tu2022maxim}	&39.45 / 0.962		&\sotab{32.84} / \sotaa{0.935}		&36.15 / 0.949	\\
    \myrowcolour \textcolor{violet}{\textdagger}BANet~\citep{tsai2022banet}	&39.55 / 0.971		&32.00 / 0.923		&35.78 / 0.947	\\
    \textcolor{violet}{\textdagger}MSSNet~\citep{kim2022mssnet}	&39.76 / 0.972		&32.10 / 0.928		&35.93 / 0.950	\\
    DeepRFT+~\citep{mao2023intriguing} & 39.84 / 0.972 & 32.19 / 0.931 & 36.02 / 0.952 \\
    \myrowcolour \textcolor{violet}{\textdagger}Stripformer~\citep{tsai2022stripformer} & 39.84 / \sotab{0.974}  & 32.48 / 0.929 &36.16 / 0.952 \\
    \textcolor{violet}{\textdagger}GRL-B~\citep{li2023efficient} & \sotab{40.20} / \sotab{0.974}		&32.82 / 0.932	 &\sotab{36.51} / \sotab{0.953} \\
    \myrowcolour \textcolor{violet}{\textdagger}Hi-IR-L	(Ours) &\sotaa{40.40} / \sotaa{0.976}		&\sotaa{32.92} / \sotab{0.933}		&\sotaa{36.66} / \sotaa{0.954}	\\
    \bottomrule[0.1em]
    \end{tabular}}
}
\vspace{-3mm}
\end{table}

%% file: tables/defocus_deblur_table.tex
\begin{table}[!t]
    \begin{center}
    \caption{\textit{\textbf{Defocus deblurring}} results. 
    \textbf{D:} dual-pixel defocus deblurring.}
    \label{tab:defocus_deblurring}
    \setlength{\tabcolsep}{2.0pt}
    \setlength{\extrarowheight}{0.7pt}
    \scalebox{0.65}{
    \begin{tabular}{l | c | c | c | c | c | c | c | c | c | c | c | c}
    \toprule[0.1em]
    \multirow{2}{*}{Method} & \multicolumn{4}{c|}{\textbf{Indoor Scenes}} & \multicolumn{4}{c|}{\textbf{Outdoor Scenes}} & \multicolumn{4}{c}{\textbf{Combined}} \\ \cline{2-13}
    	&PSNR$\uparrow$ & SSIM$\uparrow$ & MAE$\downarrow$ & LPIPS$\downarrow$			&PSNR$\uparrow$ & SSIM$\uparrow$ & MAE$\downarrow$ & LPIPS$\downarrow$ &PSNR$\uparrow$ & SSIM$\uparrow$ & MAE$\downarrow$ & LPIPS$\downarrow$		\\ \hline
    DPDNet$_D$~\citep{abuolaim2020defocus}	&27.48	&0.849	&0.029	&0.189	&22.90	&0.726	&0.052	&0.255	&25.13	&0.786	&0.041	&0.223	\\
    \myrowcolour RDPD$_D$~\citep{abdullah2021rdpd}	&28.10	&0.843	&0.027	&0.210	&22.82	&0.704	&0.053	&0.298	&25.39	&0.772	&0.040	&0.255	\\
    Uformer$_D$~\citep{wang2022uformer}	&28.23	&0.860	&0.026	&0.199	&23.10	&0.728	&0.051	&0.285	&25.65	&0.795	&0.039	&0.243	\\
    \myrowcolour IFAN$_D$~\citep{Lee_2021_CVPRifan}	&28.66	&0.868	&\textcolor{blue}{0.025}	&0.172	&23.46	&0.743	&0.049	&0.240	&25.99	&0.804	&0.037	&0.207	\\
    Restormer$_D$~\citep{zamir2022restormer}	&\textcolor{blue}{29.48}	&\textcolor{blue}{0.895}	&\textcolor{red}{0.023}	&\textcolor{blue}{0.134}	&\textcolor{blue}{23.97}	&\textcolor{blue}{0.773}	&\textcolor{blue}{0.047}	&\textcolor{blue}{0.175}	&\textcolor{blue}{26.66}	&\textcolor{blue}{0.833}	&\textcolor{blue}{0.035}	&\textcolor{blue}{0.155}	\\
    \myrowcolour Hi-IR$_D$-B (Ours)	&\textcolor{red}{29.70}	&\textcolor{red}{0.902}	&\textcolor{red}{0.023}	&\textcolor{red}{0.116}	&\textcolor{red}{24.46}	&\textcolor{red}{0.798}	&\textcolor{red}{0.045}	&\textcolor{red}{0.154}	&\textcolor{red}{27.01}	&\textcolor{red}{0.848}	&\textcolor{red}{0.034}	&\textcolor{red}{0.135}	\\ \bottomrule[0.1em]
    \end{tabular}}
    \end{center}
    \vspace{-2mm}
\end{table}

%% file: tables/allweather.tex
\begin{table}[!t]
\parbox{0.53\linewidth}{
\centering
\caption{\textit{\textbf{Image demosaicking}} results.}
\label{tab:demosaicking}
\vspace{-2mm}
\setlength{\extrarowheight}{0.7pt}
\setlength{\tabcolsep}{1.5pt}
\scalebox{0.65}{
    \begin{tabular}{l | c c c c  c c c c }
    \toprule
    Datasets	&Matlab	&\makecell{DDR}	&\makecell{DeepJoint}	&\makecell{RLDD}	 &\makecell{DRUNet} &\makecell{RNAN}		&\makecell{GRL-S} &\makecell{ Hi-IR (Ours)}	\\ \hline
    
    Kodak	&35.78	&41.11	&42.00	&42.49	&42.68	&43.16	&\sotab{43.57}	&\sotaa{43.69}	\\
    McMaster	&34.43	&37.12	&39.14	&39.25	&39.39	&39.70	&\sotab{40.22}	&\sotaa{40.78}	\\
    \bottomrule[0.1em]
    \end{tabular}
    }
}
\hspace{4pt}
\parbox{0.43\linewidth}{\centering
\caption{\textit{\textbf{IR in AWC}} results.}
\label{tab:weather}
\vspace{-2mm}
\setlength{\extrarowheight}{0.7pt}
\setlength{\tabcolsep}{1.0pt}
\scalebox{0.65}{
        \begin{tabular}{c|cccc}
        \toprule[0.1em]
        Dataset & All-in-One & TransWeather & SemanIR & Ours \\ \midrule
        \textbf{RainDrop}  
        & \sotaa{31.12} & 28.84 &30.82 & \sotab{30.84} \\  
        
        \textbf{Test1 (rain+fog)} 
        & 24.71 & {27.96} & \sotab{29.57}& \sotaa{30.93} \\
        
        \textbf{SnowTest100k-L} 
        & 28.33 & {28.48} & \sotab{30.76} & \sotaa{30.85} \\   
        \bottomrule[0.1em]
    \end{tabular}
    
    }
}

\vspace{-2mm}
\end{table}

        
        

%% file: sections_revise/7-conclusion.tex
\section{Conclusion}
\label{sec:conclusion}
In this paper, we introduced a hierarchical information flow principle for IR. Leveraging this concept, we devised a new model called Hi-IR, which progressively propagates information within local regions, facilitates information exchange in non-local ranges, and mitigates information isolation in the global context. We investigated how to scale up an IR model.
The effectiveness and generalizability of Hi-IR was validated through comprehensive experiments across various IR tasks.

%% file: sections_revise/X-supplementary.tex
\appendix
\section*{\Large{\textbf{Appendix}}}
\label{sec:Appendix}


\section{Experimental Settings}
\label{sec:supp:training}
\subsection{Architecture Details}
\label{subsec:supp:architecture_details}
We choose two commonly used basic architectures for IR tasks including the U-shape hierarchical architecture and the columnar architecture.
The columnar architecture is used for image SR while the U-shape architecture is used for other IR tasks including image denoising, JPEG CAR, image deblurring, IR in adverse weather conditions, image deblurring, and image demosaicking. 
We included details on the structure of the Hi-IR in Tab.~\ref{table:model_details}. This table outlines the number of Hi-IR stages and the distribution of Hi-IR layers within each stage for a thorough understanding of our model's architecture.

\begin{figure}[!h]
    \centering
    \includegraphics[width=0.91\linewidth]{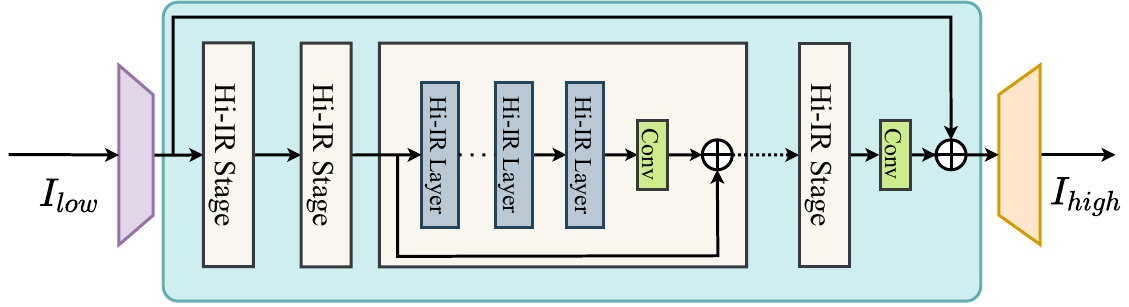}
    \vspace{-2mm}
    \caption{The columnar Hi-IR architecture.}
    \label{fig:architecture_columnar}
    \vspace{0mm}
\end{figure}

\begin{table}[!h]
    \centering
    \small
    \vspace{-3mm}
    \caption{The details of the Hi-IR stages and Hi-IR layers per stage for both architectures.}
    \setlength{\extrarowheight}{1pt}
    \setlength\tabcolsep{3pt} 
    \label{table:model_details}
    \vspace{-3mm}
    \scalebox{0.86}{
    \begin{tabular}{c|ccc|cc}
    \toprule[0.1em]
    & \multicolumn{3}{c|}{U-shaped architecture}         & \multicolumn{2}{c}{Columnar architecture}                    \\ \hline
    & Down Stages & Upstages & Latent Stage & \multicolumn{1}{c}{Hi-IR-Base} & Hi-IR-Large \\ \hline
    Num. of Hi-IR Stages      & 3           & 3        & 1            & \multicolumn{1}{c}{6}           & 8            \\
    Num. of Hi-IR Layer/Stage & 6           & 6        & 6            & \multicolumn{1}{c}{6}           & 8            \\ \bottomrule[0.1em]
    \end{tabular}}
    \vspace{-4mm}
\end{table}

\subsection{Training Details}
The proposed Hi-IR explores \textbf{7} different IR tasks, and the training settings vary slightly for each task. These differences encompass the architecture of the proposed Hi-IR, variations in training phases, choice of the optimizer, employed loss functions, warm-up settings, learning rate schedules, batch sizes, and patch sizes. We have provided a comprehensive overview of these details.

In addition, there are several points about the training details we want to make further explanation. 1) 
For image SR, the network is pre-trained on ImageNet~\citep{deng2009imagenet}.
This is inspired by previous works~\citep{dong2014learning,chen2021pre,li2021efficient,chen2023activating}. 
2) The optimizer used for IR in AWC is Adam~\citep{kingma2014adam}, while AdamW~\citep{loshchilov2018decoupled} is used for the rest IR tasks. 
3) The training losses for IR in AWC are the smooth L1 and the Perception VGG loss~\citep{johnson2016perceptual,simonyan2015very}. For image deblurring, the training loss is the Charbonnier loss. For the rest IR task, the L1 loss is commonly used during the training. 4) For IR in AWC, we adopted similar training settings as Transweather~\citep{valanarasu2022transweather}, the model is trained for a total of 750K iterations.


\subsection{Data and Evaluation}
The training dataset and test datasets for different IR tasks are described in this section.
For IR in AWC, we used a similar training pipeline as Transweather with only one phase. Additionally, for tasks such as image super-resolution (SR), JPEG CAR, image denoising, and demosaicking, how the corresponding low-quality images are generated is also briefly introduced below.

\noindent\textbf{Image SR.}
For image SR, the LR image is synthesized by \texttt{Matlab} bicubic downsampling function before the training. We investigated the upscalingg factors $\times2$, $\times3$, and $\times4$. 

\begin{itemize}[leftmargin=*]
    \item  The training datasets: DIV2K~\citep{agustsson2017ntire} and Flickr2K~\citep{lim2017enhanced}. 

    \item  The test datasets: 
    Set5~\citep{bevilacqua2012low}, Set14~\citep{zeyde2010single}, BSD100~\citep{martin2001database}, Urban100~\citep{huang2015single}, and Manga109~\citep{matsui2017sketch}.
\end{itemize}

\noindent\textbf{Image Denoising.}
For image denoising, we conduct experiments on both color and grayscale image denoising. During training and testing, noisy images are generated by adding independent additive white Gaussian noise (AWGN) to the original images. The noise levels are set to $\sigma = 15, 25, 50$. We train individual networks at different noise levels. The network takes the noisy images as input and tries to predict noise-free images. Additionally, we also tried to train one model for all noise levels.

\begin{itemize}[leftmargin=*]
    \item  The training datasets: DIV2K~\citep{agustsson2017ntire}, Flickr2K~\citep{lim2017enhanced}, WED~\citep{ma2016waterloo}, and BSD400~\citep{martin2001database}. 

    \item  The test datasets for color image: CBSD68~\citep{martin2001database}, Kodak24~\citep{franzen1999kodak}, McMaster~\citep{zhang2011color}, and Urban100~\citep{huang2015single}.
    
    \item  The test datasets for grayscale image: Set12~\citep{zhang2017beyond}, BSD68~\citep{martin2001database}, and Urban100~\citep{huang2015single}.
\end{itemize}

\noindent\textbf{JPEG compression artifact removal.}
For JPEG compression artifact removal, the JPEG image is compressed by the \texttt{cv2} JPEG compression function. The compression function is characterized by the quality factor. We investigated four compression quality factors including 10, 20, 30, and 40. The smaller the quality factor, the more the image is compressed, meaning a lower quality. We also trained one model to deal with different quality factors.

\begin{itemize}[leftmargin=*]
    \item  The training datasets: DIV2K~\citep{agustsson2017ntire}, Flickr2K~\citep{lim2017enhanced}, and WED~\citep{ma2016waterloo}. 

    \item  The test datasets: Classic5~\citep{foi2007Classic5}, LIVE1~\citep{sheikh2005live}, Urban100~\citep{huang2015single}, BSD500~\citep{arbelaez2010contour}. 
\end{itemize}

\noindent\textbf{IR in Adverse Weather Conditions.} 
For IR in adverse weather conditions, the model is trained on a combination of images degraded by a variety of adverse weather conditions. The same training and test dataset is used as in Transweather~\citep{valanarasu2022transweather}. The training data comprises 9,000 images sampled from Snow100K \citep{liu2018desnownet}, 1,069 images from Raindrop \citep{qian2018attentive}, and 9,000 images from Outdoor-Rain \citep{li2019heavy}. Snow100K includes synthetic images degraded by snow, Raindrop consists of real raindrop images, and Outdoor-Rain contains synthetic images degraded by both fog and rain streaks. The proposed method is tested on both synthetic and real-world datasets.

\begin{itemize}[leftmargin=*]
    \item  The test datasets: test1 dataset~\citep{li2020all, li2019heavy}, the RainDrop test dataset~\citep{qian2018attentive}, and the Snow100k-L test. 
\end{itemize}

\noindent\textbf{Image Deblurring.}
For single-image motion deblurring, 

\begin{itemize}[leftmargin=*]
    \item  The training datasets: GoPro~\citep{nah2017deep} dataset. 

    \item  The test datasets: 
    GoPro~\citep{nah2017deep}, HIDE~\citep{shen2019human}, RealBlur-R~\citep{rim2020real}, and RealBlur-J~\citep{rim2020real} datasets.
\end{itemize}

\noindent\textbf{Defocus Deblurring.}
The task contains two modes including single-image defocus deblurring and dual-pixel defocus deblurring. For single-image defocus deblurring, only the blurred central-view image is available. For dual-pixel defocus deblurring, both the blurred left-view and right-view images are available. The dual-pixel images could provide additional information for defocus deblurring and thus could lead to better results. PSNR, SSIM, and mean absolute error (MAE) on the RGB channels are reported. Additionally, the image perceptual quality score LPIPS is also reported.

\begin{itemize}[leftmargin=*]
    \item  The training datasets: DPDD~\citep{abuolaim2020defocus} training dataset. The training subset contains 350 scenes.  

    \item  The test datasets: 
    DPDD~\citep{abuolaim2020defocus} test dataset. The test set contains 37 indoor scenes and 39 outdoor scenes
\end{itemize}

\noindent\textbf{Image Demosaicking.}
For image demosaicking, the mosaic image is generated by applying a Bayer filter on the ground-truth image. Then the network try to restore high-quality image. The mosaic image is first processed by the default \texttt{Matlab} demosaic function and then passed to the network as input. 

\begin{itemize}[leftmargin=*]
    \item  The training datasets: DIV2K~\citep{agustsson2017ntire} and Flickr2K~\citep{lim2017enhanced}. 

    \item  The test datasets: Kodak~\citep{franzen1999kodak}, McMaster~\citep{zhang2011color}. 
\end{itemize}

\section{Model Scaling-up}
\label{sec:supp:model_scaling_up}

\begin{figure}[t]
    \centering
    \includegraphics[width=0.99\linewidth]{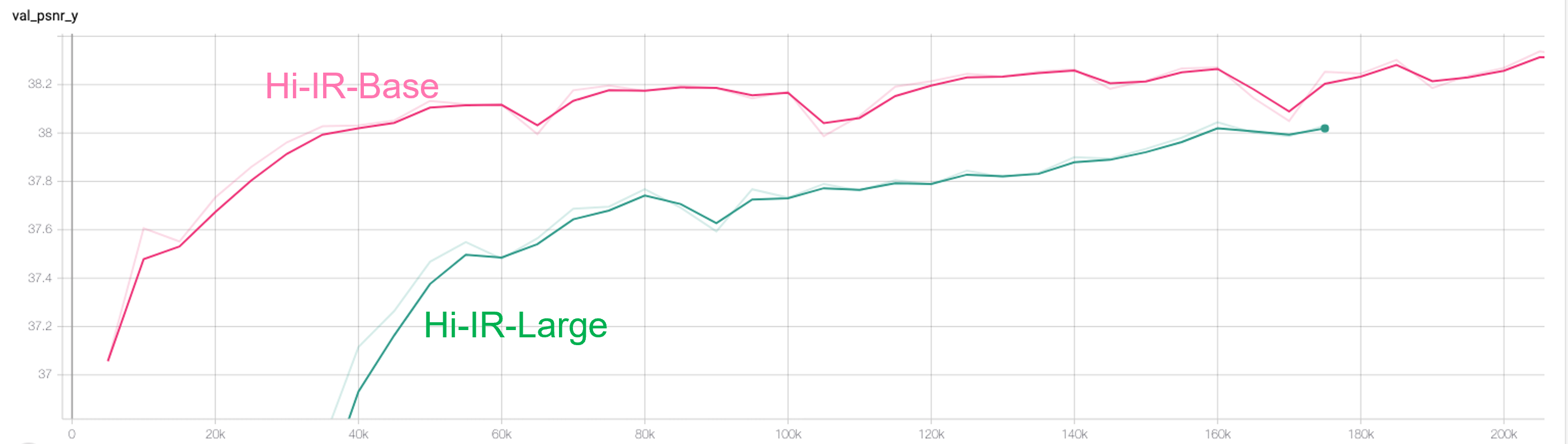}
    \caption{When the SR model is scale-up from Hi-IR-L to Hi-IR-B, the model Hi-IR-L converges slower than Hi-IR-B.}
    \label{fig:convergence}
\end{figure}

\input{tables/ablation_large_model_full}

As mentioned in the main paper, when the initially designed SR model is scaled up from about 10M parameters to about 50M parameters, the performance of the large SR model becomes worse. The effect is shown in Fig.~\ref{fig:convergence}. 
The PSNR curve on the Set5 dataset for the first 200k iterations is shown in this figure. The scale-up model Hi-IR-L converges slower than the smaller model Hi-IR-B.
The same phenomenon could be observed by comparing the first two rows for each upscaling factor in Tab.~\ref{tab:scaling_up_convergence_full}, where scaled-up models converge to worse local minima. A similar problem occurs in previous works~\citep{chen2023activating,lim2017enhanced}.

\section{Space and Time Complexity}
\label{sec:space_time_complexity}

We compare the space and time complexity and the effective receptive field of the proposed method with a couple of other self-attention methods including global attention and window attention. Suppose the input feature has the dimension $B \times C \times H \times W$, the window size of window attention is $p$, the number of attention heads is $h$, larger patch size of the proposed L2 information flow is $P=s \times p$, the expansion ratio of the MLP in transformer layer is $\gamma$. For time complexity, both self-attention and the feed-forward network are considered. For space complexity, we consider the tensors that have to appear in the memory at the same time, which include the input tensor, the query tensor, the key tensor, the value tensor, and the attention map.

The time complexity of the proposed transformer layer is
\begin{equation}
    \mathcal{O}\left((5+2\gamma)BHWC^2 + \frac{3}{2}BHWp^2C+\frac{3}{2}BHWs^2C+9\gamma BHWC\right).
\end{equation}
The last term is very small compared with the former two terms, and can be omitted. Thus, the time complexity is simplified as \begin{equation}
    \mathcal{O}\left((5+2\gamma)BHWC^2 + \frac{3}{2}BHWp^2C+\frac{3}{2}BHWs^2C\right).
\end{equation}

The space complexity of the proposed transformer layer is
\begin{equation}
    \mathcal{O}\left(3BHWC + BHWh\max{(p^2, s^2)}\right).
\end{equation} 
The maximum receptive field of two consecutive transformer layers is $16P$.

In the Tab.~\ref{tab:space_time_complexity}, we list the space and time complexity, and maximum receptive field of global attention, window attention, and the proposed method. As shown in this table, window attention is much more efficient than global attention but with the cost of reduced receptive field. The proposed hierarchicial information flow mechanism is more efficient than window attention in propagating information to the global range. As shown in the third row, to achieve the same receptive field as the proposed method, space and time complexity of window attention is much higher than that of the proposed method. 

\begin{table}[!htb]
\scriptsize
\setlength{\extrarowheight}{0.7pt}
\setlength{\tabcolsep}{1.5pt}
\centering
\scalebox{1.0}{
\begin{tabular}{c|ccc}
    \toprule[0.1em]
\textbf{Attn. method} &\textbf{Time complexity} &\textbf{Space complexity} & \makecell{\textbf{Max receptive field of} \\ \textbf{two transformer layers}} \\ \hline
\myrowcolour Global Attn.                  & $\mathcal{O}\left((4+2\gamma)BHWC^2 + {2}B(HW)^2C\right)$               & $\mathcal{O}\left(4BHWC + B(HW)^2h\right)$                & $H \times W$      \\
Window Attn. ($p \times p$)   & $\mathcal{O}\left((4+2\gamma)BHWC^2 + {2}BHWp^2C\right)$                & $\mathcal{O}\left(4BHWC + BHWhp^2\right)$                 & $2p \times 2p$    \\
\myrowcolour Window Attn. ($8P \times 8P$) & $\mathcal{O}\left((4+2\gamma)BHWC^2 + {128}BHWp^2s^2C\right)$           & $\mathcal{O}\left(4BHWC + 64BHWhp^2s^2\right)$            & $16P \times 16P$  \\
The proposed                  & $\mathcal{O}\left((5+2\gamma)BHWC^2 + \frac{3}{2}BHW(p^2+s^2)C\right)$  & $\mathcal{O}\left(3BHWC + BHWh\max{(p^2, s^2)}\right)$    & $16P \times 16P$  \\
    \bottomrule[0.1em]
    \end{tabular}}
    \caption{Space and time complexity of classical attention mechanisms.}
    \label{tab:space_time_complexity}

\end{table}

\section{More Quantitative Experimental Results}
\label{sec:supp:more_experimental_results}
Due to the limited space in the main manuscript, we only report a part of the experimental result. In this section, we show the full quantitative experimental results for each IR task in the following.



\subsection{Single-image defocus deblurring}
\input{tables/defocus_deblur_table_single_pixel}
In addition to the dual-pixel defocus deblurring results, we also shown single-image defocus deblurring results in Tab.~\ref{tab:defocus_deblurring_single_image}




\subsection{Generalizing one model to more types degradations}

\input{tables/dn_table_one_for_all}

\revise{To validate the generalization capability of the proposed method to different types of degradation, we conducted the following experiments. First, we used the same model for both denoising and JPEG compression artifact removal tasks. Notably, a single model was trained to handle varying levels of degradation. The experimental results for denoising are shown in Tab.~\ref{tab:denoising_one_for_all} while the results for JPEG compression artifact removal are shown in Tab.~\ref{tab:jpeg_car_color} and Tab.~\ref{tab:jpeg_car_grapyscale}. Second, we performed experiments on image restoration under adverse weather conditions, including rain, fog, and snow. The results are shown in Tab.~\ref{tab:weather}. These three sets of experiments collectively highlight that the proposed hierarchical information flow mechanism enables training a single model that generalizes effectively to various types and levels of degradation.}

\section{\revise{Comparison with ShuffleFormer and Shuffle Transformer}}
\revise{We compare with Random shuffle transformer (ShuffleFormer)~\citep{xiao2023random} and Shuffle transformer~\citep{huang2021shuffle}. Both methods use spatial shuffle operations to facilitate non-local information exchange, with one being random and the other deterministic.}

\revise{Random Shuffle Transformer (ShuffleFormer)~\citep{xiao2023random} applies random shuffling on the spatial dimension, which increases the probability of global information existing within a local window. While this operation extends the receptive field globally in a single step, it compromises the relevance of pixels within the window. In contrast, the hierarchical information flow proposed in this paper progressively propagates information from local to global while preserving the relevance of attended pixels. A comparison with ShuffleFormer on image deblurring is presented in Tab.~\ref{tab:motion_deblurring}. Hi-IR outperforms ShuffleFormer by a significant margin while using 55.5\% fewer parameters. This demonstrates the effectiveness of the hierarchical information flow method introduced in this work.}

\revise{Shuffle Transformer~\citep{huang2021shuffle} employs a spatial shuffle operation to aggregate information from distant pixels or tokens. However, it differs from the proposed Hi-IR in several key aspects. First, Shuffle Transformer does not enable progressive information propagation within a hierarchical tree structure. Second, its shuffle operation is based on a fixed grid size of $g = 8$. The distance between pixels in the shuffled window is $H/g$ and $W/g$ along the two axes, which directly depends on the image size. For large images (e.g., 1024 pixels), this design forces distant pixels to attend to one another, often introducing irrelevant information. Consequently, this operation is unsuitable for image restoration tasks, where image sizes can become extremely large. In contrast, the L2 information flow attention proposed in this paper limits the maximum patch size, thereby constraining the maximum distance between pixels at this stage. This restriction enhances the relevance of pixel interactions, making it more effective for image restoration tasks.}

\section{More Visual Results}
\label{sec:supp:more_visual_results}
To further support the effectiveness and generalizability of the proposed Hi-IR intuitively. We provide more visual comparison in terms of image SR (Fig.~\ref{fig:supp_visual_sr_b100_part1}, 
and
Fig.~\ref{fig:supp_visual_sr_manga109_part1}),
image denoising (Fig.~\ref{fig:supp_visual_dn_urban100}), JPEG compression artifact removal (Fig.~\ref{fig:supp_visual_jpeg_color_bsd500}
), image restoration in adverse weather conditions(Fig.~\ref{fig:supp_weather_fig}), and single-image deblurring (Fig.~\ref{fig:supp_visual_motion_db_part1} and Fig.~\ref{fig:supp_visual_motion_db_part2}) blow. As shown in those figures, the visual results of the proposed Hi-IR are improved compared with the other methods.

\section{Limitations}
\label{sec:supp:limitation}
Despite the state-of-the-art performance of Hi-IR, our explorations towards scaling up the model for IR in this paper are still incomplete. Scaling up the IR model is intricate, involving considerations like model design, data collection, and computing resources. We hope our work can catalyze positive impacts on future research, encouraging more comprehensive scaling-up explorations and propelling IR into the domain of large-scale models.

\input{figs/visual_supp}

%% file: tables/ablation_large_model_full.tex
\begin{table}[!h]
\caption{Model scaling-up exploration with SR.}
\label{tab:scaling_up_convergence_full}
\centering
\scalebox{0.65}{
    \begin{tabular}{c|c|cc|ccccc}
    \toprule[0.1em]
    \multirow{2}{*}{\textbf{Scale}}  &  \multirow{2}{*}{\makecell{\textbf{Model} \\ \textbf{Size}}} & \multirow{2}{*}{\makecell{\textbf{Warm} \\ \textbf{up}}} & \multirow{2}{*}{\makecell{\textbf{Conv} \\ \textbf{Type}}}  &  \multicolumn{5}{c}{\textbf{PSNR}} 
    \\\cline{5-9}
    &   & & & \textbf{Set5}& \textbf{Set14} & \textbf{BSD100} & \textbf{Urban100} & \textbf{Manga109}
    \\ \hline
    \myrowcolourpink $2\times$	
     &15.69	&No	&\texttt{conv1}	&38.52	&34.47	&32.56	&34.17	&39.77	\\
    \myrowcolourpink $2\times$	
    &57.60	&No	&\texttt{conv1}	&38.33	&34.17	&32.46	&33.60	&39.37	\\ 
    $2\times$	
    &57.60	&Yes	&\texttt{conv1}	&38.41	&34.33	&32.50	&33.80	&39.51	\\
    $2\times$	
    &54.23	&Yes	&\texttt{linear}	&\sotab{38.56}	&\sotaa{34.59}	&\sotaa{32.58}	&\sotab{34.32}	&\sotab{39.87}	\\
    $2\times$	
    &55.73	&Yes	&\texttt{conv3}	&\sotaa{38.65}	&\sotab{34.48}	&\sotaa{32.58}	&\sotaa{34.33}	&\sotaa{40.12}	\\ \midrule
    \myrowcolourpink $3\times$	
    &15.87	&No	&\texttt{conv1}	&35.06	&30.91	&29.48	&30.02	&34.41	\\
    \myrowcolourpink $3\times$	
    &57.78	&No	&\texttt{conv1}	&34.70	&30.62	&29.33	&29.11	&33.96	\\ 
    $3\times$	
    &57.78	&Yes	&\texttt{conv1}	&34.91	&30.77	&29.39	&29.53	&34.12	\\
    $3\times$	
    &54.41	&Yes	&\texttt{linear}	&\sotab{35.13}	&\sotaa{31.04}	&\sotaa{29.52}	&\sotab{30.20}	&\sotab{34.54}	\\
    $3\times$	
    &55.91	&Yes	&\texttt{conv3}	&\sotaa{35.14}	&\sotab{31.03}	&\sotab{29.51}	&\sotaa{30.22}	&\sotaa{34.76}	\\ 
    \midrule
    \myrowcolourpink $4\times$	&15.84	&No	&\texttt{conv1}	&33.00	&29.11	&27.94	&27.67	&31.41	\\
    \myrowcolourpink $4\times$	&57.74	&No	&\texttt{conv1}	&33.08	&29.19	&27.97	&27.83	&31.56	\\ 
    $4\times$	&57.74	&Yes	&\texttt{conv1}	&32.67	&28.93	&27.83	&27.11	&30.97	\\
    $4\times$	&54.37	&Yes	&\texttt{linear}	&\sotaa{33.06}	&\sotaa{29.16}	&\sotaa{27.99}	&\sotaa{27.93}	&\sotaa{31.66}	\\
    $4\times$	&55.88	&Yes	&\texttt{conv3}	&\sotaa{33.06}	&\sotaa{29.16}	&\sotab{27.97}	&\sotab{27.87}	&\sotab{31.54}	\\
    \bottomrule[0.1em]
    \end{tabular}
    }
\end{table}

%% file: tables/defocus_deblur_table_single_pixel.tex
\begin{table}[!t]
    \begin{center}
    \caption{\textit{\textbf{Sinlge-image Defocus deblurring}} results. 
    \textbf{S:} single-image defocus deblurring. 
    }
    \label{tab:defocus_deblurring_single_image}
    \setlength{\tabcolsep}{2.0pt}
    \setlength{\extrarowheight}{0.7pt}
    \scalebox{0.65}{
    \begin{tabular}{l | c | c | c | c | c | c | c | c | c | c | c | c}
    \toprule[0.1em]
    \multirow{2}{*}{Method} & \multicolumn{4}{c|}{\textbf{Indoor Scenes}} & \multicolumn{4}{c|}{\textbf{Outdoor Scenes}} & \multicolumn{4}{c}{\textbf{Combined}} \\ \cline{2-13}
    	&PSNR$\uparrow$ & SSIM$\uparrow$ & MAE$\downarrow$ & LPIPS$\downarrow$			&PSNR$\uparrow$ & SSIM$\uparrow$ & MAE$\downarrow$ & LPIPS$\downarrow$ &PSNR$\uparrow$ & SSIM$\uparrow$ & MAE$\downarrow$ & LPIPS$\downarrow$		\\ \hline
    \myrowcolour EBDB$_S$~\citep{karaali2017edge_EBDB}	&25.77	&0.772	&0.040	&0.297	&21.25	&0.599	&0.058	&0.373	&23.45	&0.683	&0.049	&0.336	\\
    DMENet$_S$~\citep{lee2019deep_dmenet}	&25.50	&0.788	&0.038	&0.298	&21.43	&0.644	&0.063	&0.397	&23.41	&0.714	&0.051	&0.349	\\
    \myrowcolour JNB$_S$~\citep{shi2015just_jnb}	&26.73	&0.828	&0.031	&0.273	&21.10	&0.608	&0.064	&0.355	&23.84	&0.715	&0.048	&0.315	\\
    DPDNet$_S$~\citep{abuolaim2020defocus}	&26.54	&0.816	&0.031	&0.239	&22.25	&0.682	&0.056	&0.313	&24.34	&0.747	&0.044	&0.277	\\
    \myrowcolour KPAC$_S$~\citep{son2021single_kpac}	&27.97	&0.852	&0.026	&0.182	&22.62	&0.701	&0.053	&0.269	&25.22	&0.774	&0.040	&0.227	\\
    IFAN$_S$~\citep{Lee_2021_CVPRifan}	&28.11	&0.861	&0.026	&0.179	&22.76	&0.720	&0.052	&0.254	&25.37	&0.789	&0.039	&0.217	\\
    \myrowcolour Restormer$_S$~\citep{zamir2022restormer}	&\textcolor{red}{28.87}	&\textcolor{blue}{0.882}	&\textcolor{blue}{0.025}	&\textcolor{blue}{0.145}	&\textcolor{blue}{23.24}	&\textcolor{blue}{0.743}	&\textcolor{blue}{0.050}	&\textcolor{blue}{0.209}	&\textcolor{blue}{25.98}	&\textcolor{blue}{0.811}	&\textcolor{blue}{0.038}	&\textcolor{blue}{0.178}	\\
    Hi-IR$_S$-B (Ours)	&\textcolor{blue}{28.73}	&\textcolor{red}{0.885}	&\textcolor{red}{0.025}	&\textcolor{red}{0.140}	&\textcolor{red}{23.66}	&\textcolor{red}{0.766}	&\textcolor{red}{0.048}	&\textcolor{red}{0.196}	&\textcolor{red}{26.13}	&\textcolor{red}{0.824}	&\textcolor{red}{0.037}	&\textcolor{red}{0.169}		\\											
    \bottomrule[0.1em]
    \end{tabular}}
    \end{center}
\end{table}

%% file: tables/dn_table_one_for_all.tex
\begin{table*}[t]
\centering
\caption{\textit{\textbf{Color and grayscale image denoising}} results. A single model is trained to handle multiple noise levels.
}
\label{tab:denoising_one_for_all}
\setlength{\extrarowheight}{0.7pt}
\setlength{\tabcolsep}{1.2pt}
\scalebox{0.625}{
    \begin{tabular}{l | c | c c c | c c c | c c c | c c c || c c c | c c c }
    \toprule[0.1em]
    \multirow{3}{*}{\textbf{Method}} & \multicolumn{1}{c|}{\multirow{3}{*}{ \makecell{\textbf{Params} \\ \textbf{[M]}} }}
     & \multicolumn{12}{c||}{\textbf{Color}} & \multicolumn{6}{c}{\textbf{Grayscale}} \\ \cline{3-20}
    & & \multicolumn{3}{c|}{\textbf{CBSD68}} & \multicolumn{3}{c|}{\textbf{Kodak24}} & \multicolumn{3}{c|}{\textbf{McMaster}} & \multicolumn{3}{c||}{\textbf{Urban100}}  & \multicolumn{3}{c|}{\textbf{Set12}}  & \multicolumn{3}{c}{\textbf{Urban100}} \\
            &  & $\sigma$$=$$15$ & $\sigma$$=$$25$ & $\sigma$$=$$50$ & $\sigma$$=$$15$ & $\sigma$$=$$25$ & $\sigma$$=$$50$ & $\sigma$$=$$15$ & $\sigma$$=$$25$ & $\sigma$$=$$50$ & $\sigma$$=$$15$ & $\sigma$$=$$25$ & $\sigma$$=$$50$ & $\sigma$$=$$15$ & $\sigma$$=$$25$ & $\sigma$$=$$50$ & $\sigma$$=$$15$ & $\sigma$$=$$25$ & $\sigma$$=$$50$ \\ \midrule
    \myrowcolour DnCNN~\citep{kiku2016beyond}	& 0.56	&33.90	&31.24	&27.95	&34.60	&32.14	&28.95	&33.45	&31.52	&28.62	&32.98	&30.81	&27.59	&32.67	&30.35	&27.18				&32.28	&29.80	&26.35	\\
    FFDNet~\citep{zhang2018ffdnet}	& 0.49	&33.87	&31.21	&27.96	&34.63	&32.13	&28.98	&34.66	&32.35	&29.18	&33.83	&31.40	&28.05	&32.75	&30.43	&27.32				&32.40	&29.90	&26.50	\\
    \myrowcolour IRCNN~\citep{zhang2017learning}	& 0.19	&33.86	&31.16	&27.86	&34.69	&32.18	&28.93	&34.58	&32.18	&28.91	&33.78	&31.20	&27.70	&32.76	&30.37	&27.12				&32.46	&29.80	&26.22	\\
    DRUNet~\citep{zhang2021plug}	& 32.64	&34.30	&31.69	&28.51	&35.31	&32.89	&29.86	&35.40	&33.14	&30.08	&34.81	&32.60	&29.61	&33.25	&30.94	&27.90				&33.44	&31.11	&27.96	\\
    \myrowcolour Restormer~\citep{zamir2022restormer}	& 26.13	&\sotab{34.39}	&\sotab{31.78}	&\sotab{28.59}	&\sotab{35.44}	&\sotaa{33.02}	&\sotaa{30.00}	&\sotab{35.55}	&\sotab{33.31}	&\sotab{30.29}	&\sotab{35.06}	&\sotab{32.91}	&\sotab{30.02}	&\sotab{33.35}	&\sotab{31.04}	&\sotab{28.01}		&\sotab{33.67}	&\sotab{31.39}	&\sotab{28.33}	\\
    TreeIR (Ours)	& 22.33	&\sotaa{34.43}	&\sotaa{31.80}	&\sotaa{28.60}	&\sotaa{35.42}	&\sotab{33.00}	&\sotab{29.95}	&\sotaa{35.67}	&\sotaa{33.43}	&\sotaa{30.38}	&\sotaa{35.46}	&\sotaa{33.32}	&\sotaa{30.47}	&\sotaa{33.49}	&\sotaa{31.18}	&\sotaa{28.14}		&\sotaa{34.09}	&\sotaa{31.87}	&\sotaa{28.86}	\\
    \bottomrule[0.1em]
    \end{tabular}
}
\end{table*}

%% file: figs/visual_supp.tex
\begin{figure*}[!t]
    \centering
    \includegraphics[width=0.9\linewidth]{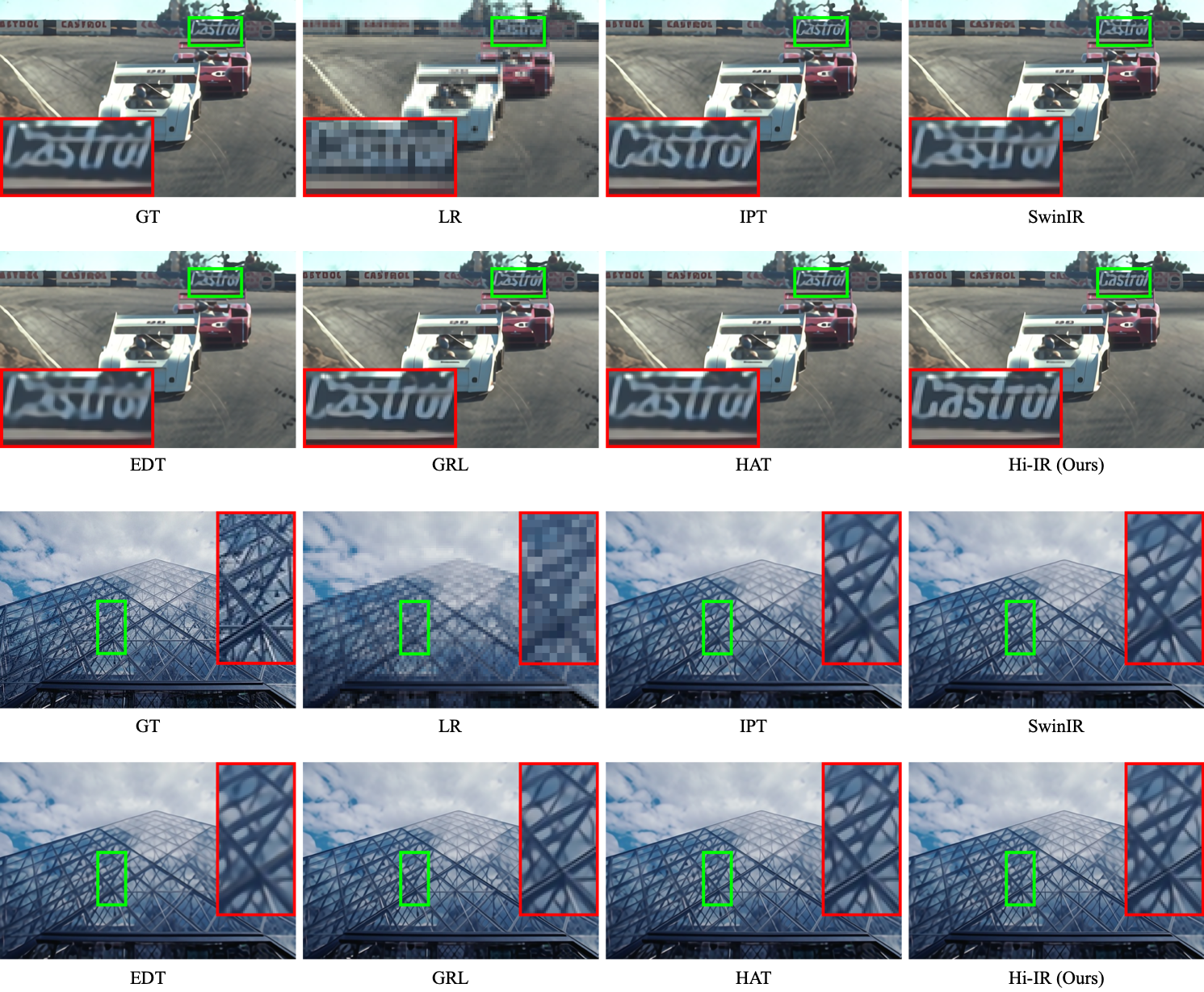}
    \vspace{-3mm}
    \caption{Visual results for classical image $\times 4$ SR on B100 dataset.}
    \label{fig:supp_visual_sr_b100_part1}
\end{figure*}



\begin{figure*}[!t]
    \centering
    \includegraphics[width=1.0\linewidth]{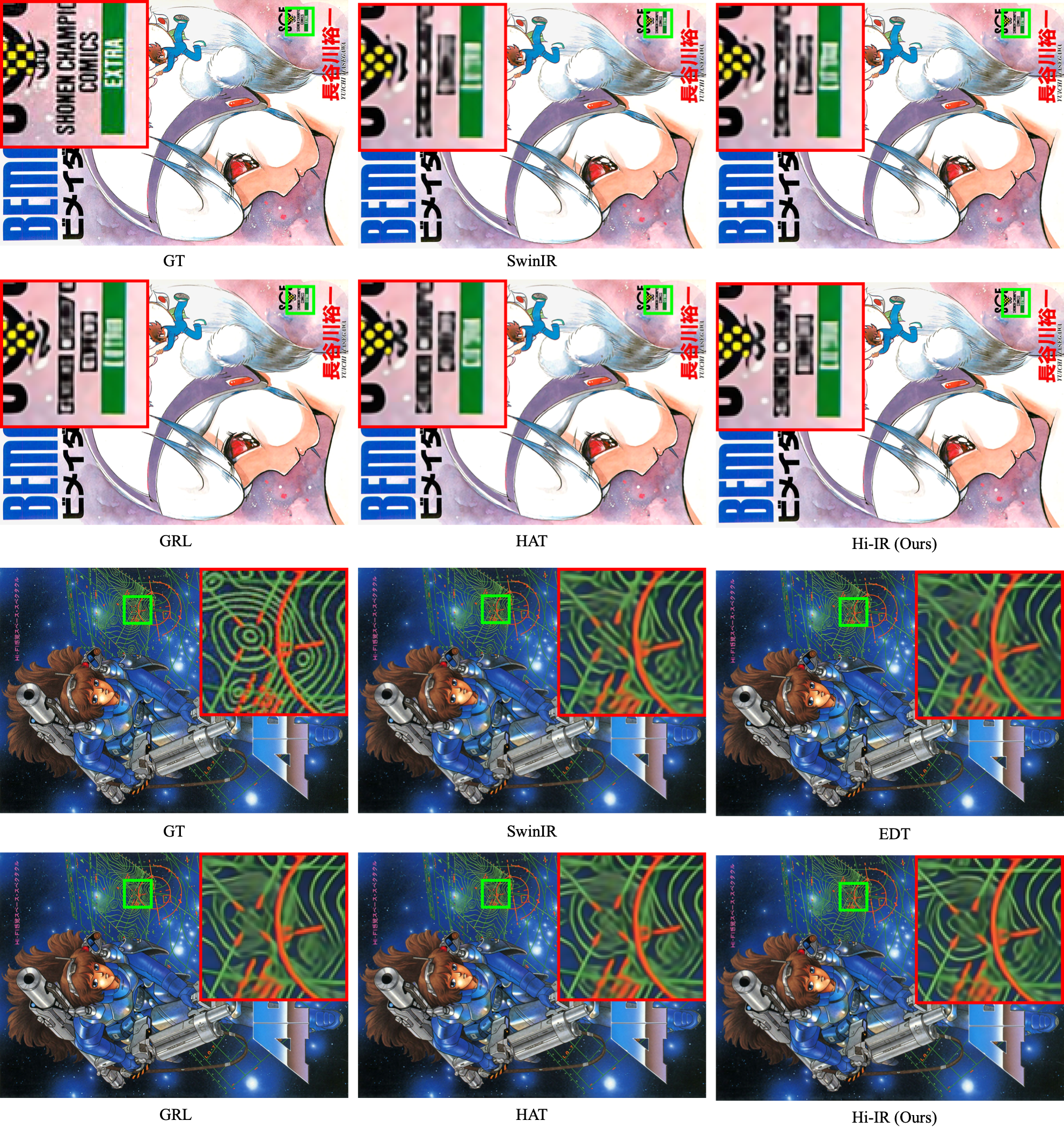}
    \vspace{-6mm}
    \caption{Visual results for classical image $\times 4$ SR on Manga109 dataset.}
    \label{fig:supp_visual_sr_manga109_part1}
\end{figure*}


\begin{figure*}[!t]
    \centering
    \includegraphics[width=1.0\linewidth]{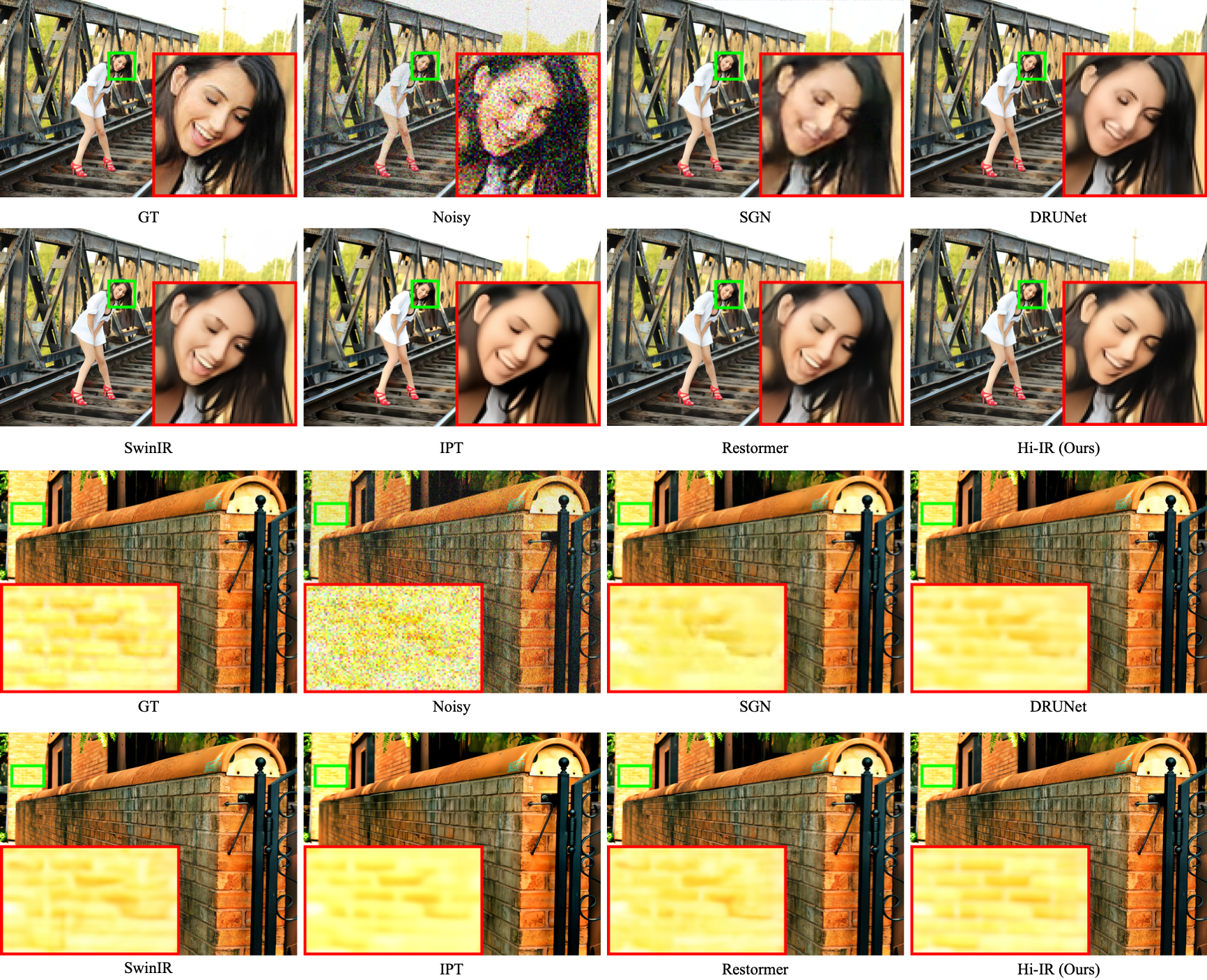}
    \vspace{-6mm}
    \caption{Visual results for classical color image denoising on Urban100 dataset. The noise level is $\sigma=50$.}
    \label{fig:supp_visual_dn_urban100}
\end{figure*}

\begin{figure*}[!t]
    \centering
    \includegraphics[width=1.0\linewidth]{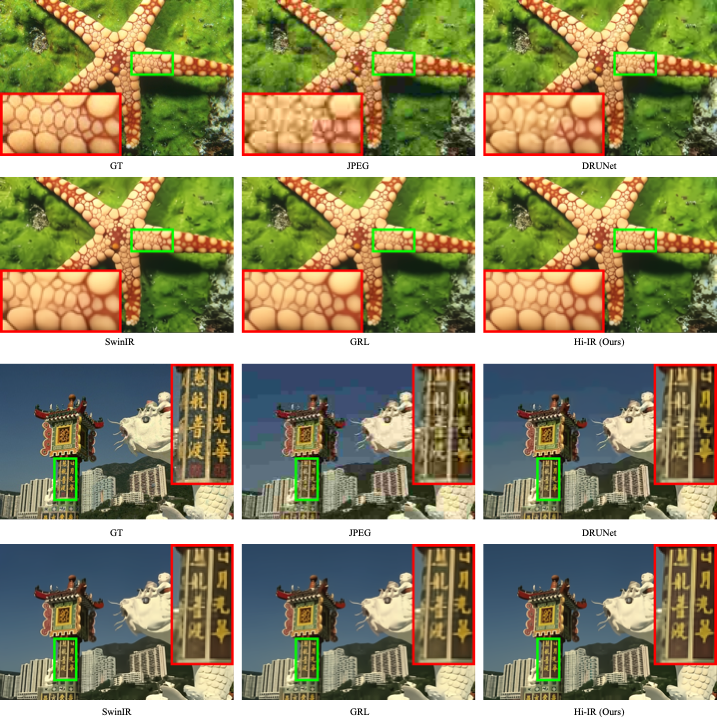}
    \vspace{-6mm}
    \caption{Visual results for color image JPEG compression artifact removal on BSD500 dataset. The quality factor of JPEG image compression is $10$.}
    \label{fig:supp_visual_jpeg_color_bsd500}
\end{figure*}


\begin{figure*}[!t]
    \centering
    \includegraphics[width=1.0\linewidth]{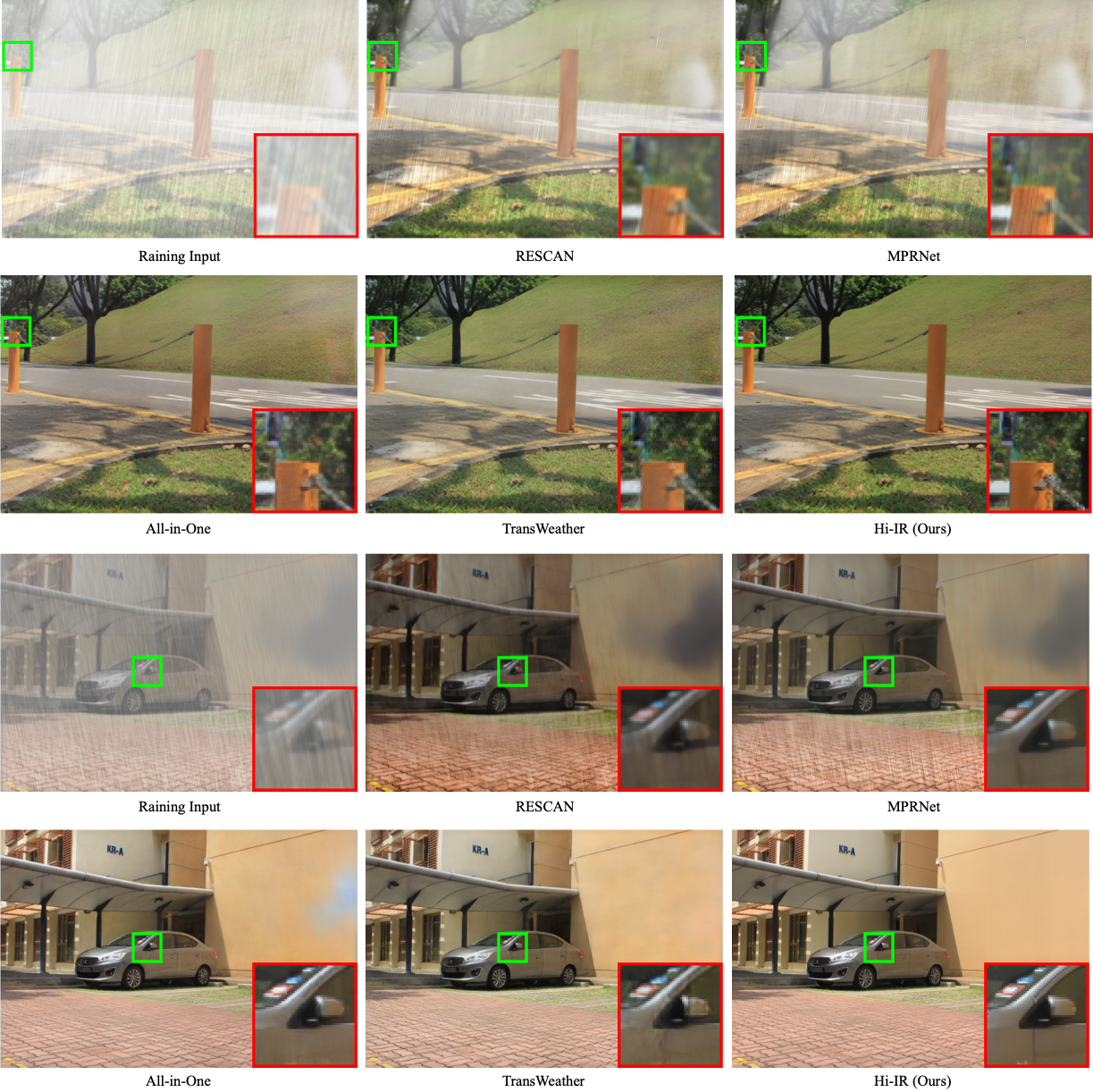}
    \vspace{-6mm}
    \caption{Visual results for restoring images in adverse weather conditions.}
    \label{fig:supp_weather_fig}
\end{figure*}

\begin{figure*}[!t]
    \centering
    \includegraphics[width=1.0\linewidth]{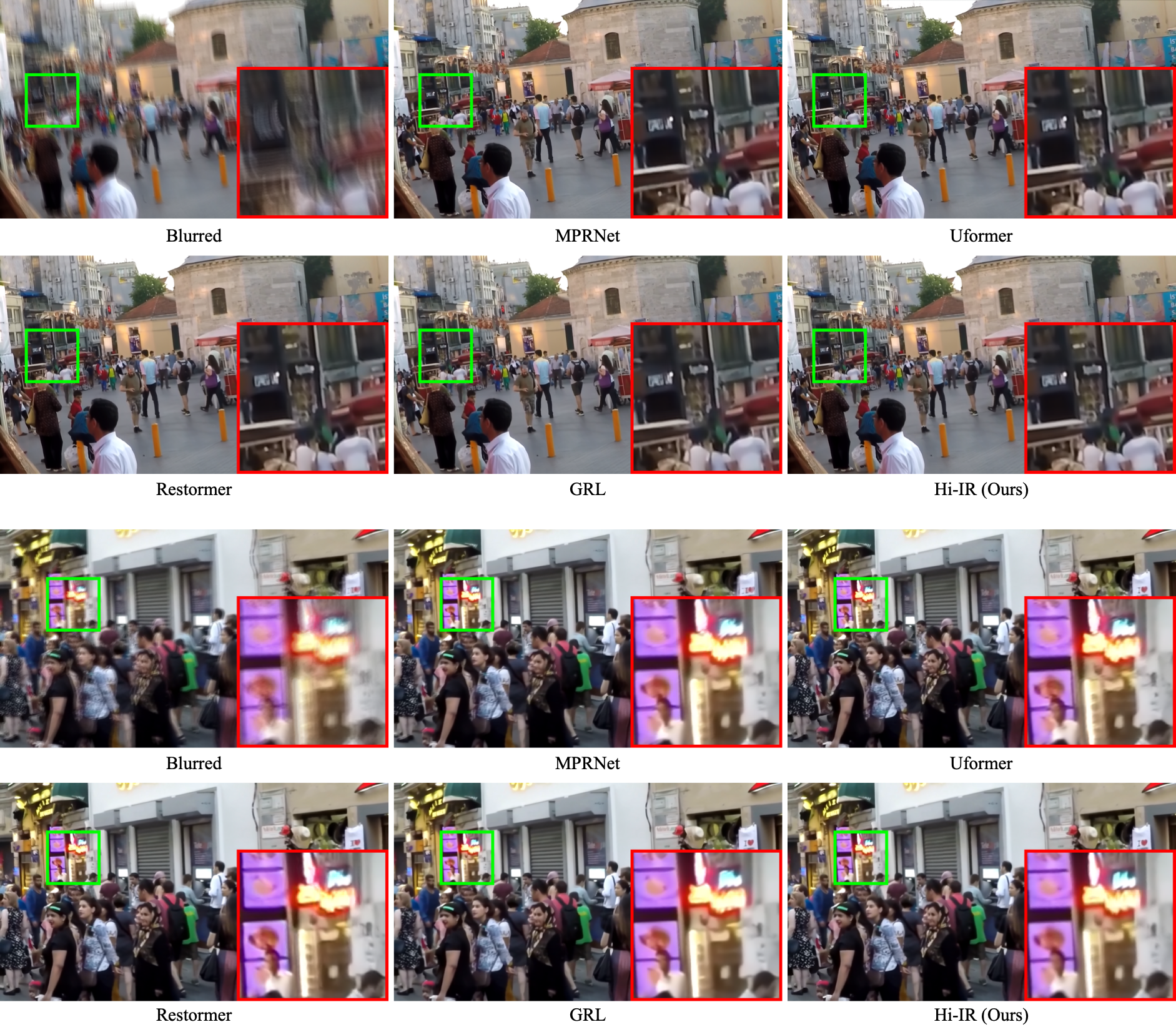}
    \vspace{-6mm}
    \caption{Visual results for single image motion deblurring. The proposed method Hi-IR could recover sharper details compared with the other methods.}
    \label{fig:supp_visual_motion_db_part1}
\end{figure*}

\begin{figure*}[!t]
    \centering
    \includegraphics[width=1.0\linewidth]{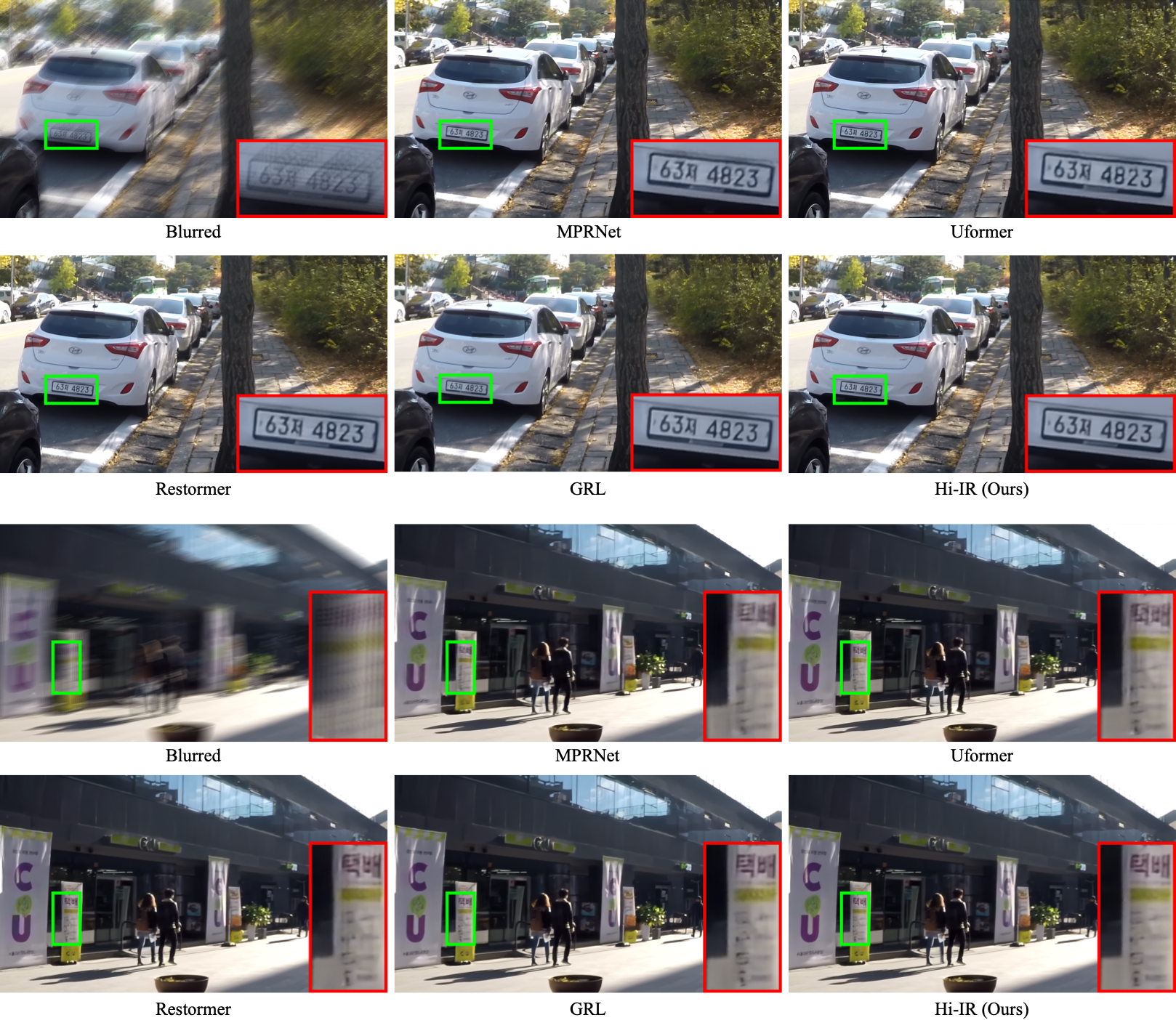}
    \vspace{-6mm}
    \caption{Visual results for single image motion deblurring. The proposed method Hi-IR could recover sharper details compared with the other methods.}
    \label{fig:supp_visual_motion_db_part2}
\end{figure*}

%% file: main.bbl
\begin{thebibliography}{97}
\providecommand{\natexlab}[1]{#1}
\providecommand{\url}[1]{\texttt{#1}}
\expandafter\ifx\csname urlstyle\endcsname\relax
  \providecommand{\doi}[1]{doi: #1}\else
  \providecommand{\doi}{doi: \begingroup \urlstyle{rm}\Url}\fi

\bibitem[Abuolaim \& Brown(2020)Abuolaim and Brown]{abuolaim2020defocus}
Abdullah Abuolaim and Michael~S Brown.
\newblock Defocus deblurring using dual-pixel data.
\newblock In \emph{ECCV}, pp.\  111--126. Springer, 2020.

\bibitem[Abuolaim et~al.(2021)Abuolaim, Delbracio, Kelly, Brown, and Milanfar]{abdullah2021rdpd}
Abdullah Abuolaim, Mauricio Delbracio, Damien Kelly, Michael~S. Brown, and Peyman Milanfar.
\newblock Learning to reduce defocus blur by realistically modeling dual-pixel data.
\newblock In \emph{ICCV}, 2021.

\bibitem[Agustsson \& Timofte(2017)Agustsson and Timofte]{agustsson2017ntire}
Eirikur Agustsson and Radu Timofte.
\newblock {NTIRE} 2017 challenge on single image super-resolution: Dataset and study.
\newblock In \emph{CVPRW}, pp.\  126--135, 2017.

\bibitem[Anwar \& Barnes(2020)Anwar and Barnes]{anwar2020densely}
Saeed Anwar and Nick Barnes.
\newblock Densely residual laplacian super-resolution.
\newblock \emph{IEEE TPAMI}, 44\penalty0 (3):\penalty0 1192--1204, 2020.

\bibitem[Arbelaez et~al.(2010)Arbelaez, Maire, Fowlkes, and Malik]{arbelaez2010contour}
Pablo Arbelaez, Michael Maire, Charless Fowlkes, and Jitendra Malik.
\newblock Contour detection and hierarchical image segmentation.
\newblock \emph{IEEE TPAMI}, 33\penalty0 (5):\penalty0 898--916, 2010.

\bibitem[Bevilacqua et~al.(2012)Bevilacqua, Roumy, Guillemot, and Alberi-Morel]{bevilacqua2012low}
Marco Bevilacqua, Aline Roumy, Christine Guillemot, and Marie~Line Alberi-Morel.
\newblock Low-complexity single-image super-resolution based on nonnegative neighbor embedding.
\newblock In \emph{BMVC}, 2012.

\bibitem[Brown et~al.(2020)Brown, Mann, Ryder, Subbiah, Kaplan, Dhariwal, Neelakantan, Shyam, Sastry, Askell, et~al.]{brown2020language}
Tom Brown, Benjamin Mann, Nick Ryder, Melanie Subbiah, Jared~D Kaplan, Prafulla Dhariwal, Arvind Neelakantan, Pranav Shyam, Girish Sastry, Amanda Askell, et~al.
\newblock Language models are few-shot learners.
\newblock \emph{NeurIPS}, 33:\penalty0 1877--1901, 2020.

\bibitem[Chen et~al.(2021)Chen, Wang, Guo, Xu, Deng, Liu, Ma, Xu, Xu, and Gao]{chen2021pre}
Hanting Chen, Yunhe Wang, Tianyu Guo, Chang Xu, Yiping Deng, Zhenhua Liu, Siwei Ma, Chunjing Xu, Chao Xu, and Wen Gao.
\newblock Pre-trained image processing transformer.
\newblock In \emph{CVPR}, pp.\  12299--12310, 2021.

\bibitem[Chen et~al.(2022{\natexlab{a}})Chen, Chu, Zhang, and Sun]{chen2022simple}
Liangyu Chen, Xiaojie Chu, Xiangyu Zhang, and Jian Sun.
\newblock Simple baselines for image restoration.
\newblock In \emph{ECCV}, pp.\  17--33. Springer, 2022{\natexlab{a}}.

\bibitem[Chen et~al.(2023)Chen, Wang, Zhou, Qiao, and Dong]{chen2023activating}
Xiangyu Chen, Xintao Wang, Jiantao Zhou, Yu~Qiao, and Chao Dong.
\newblock Activating more pixels in image super-resolution transformer.
\newblock In \emph{CVPR}, pp.\  22367--22377, 2023.

\bibitem[Chen et~al.(2022{\natexlab{b}})Chen, Zhang, Gu, Kong, Yuan, et~al.]{chen2022cross}
Zheng Chen, Yulun Zhang, Jinjin Gu, Linghe Kong, Xin Yuan, et~al.
\newblock Cross aggregation transformer for image restoration.
\newblock \emph{NeurIPS}, 35:\penalty0 25478--25490, 2022{\natexlab{b}}.

\bibitem[Cho et~al.(2021)Cho, Ji, Hong, Jung, and Ko]{cho2021rethinking_mimo}
Sung-Jin Cho, Seo-Won Ji, Jun-Pyo Hong, Seung-Won Jung, and Sung-Jea Ko.
\newblock Rethinking coarse-to-fine approach in single image deblurring.
\newblock In \emph{ICCV}, 2021.

\bibitem[Chu et~al.(2022)Chu, Tian, Zhang, Wang, and Shen]{chu2022conditional}
Xiangxiang Chu, Zhi Tian, Bo~Zhang, Xinlong Wang, and Chunhua Shen.
\newblock Conditional positional encodings for vision transformers.
\newblock In \emph{ICLR}, 2022.

\bibitem[Dai et~al.(2019)Dai, Cai, Zhang, Xia, and Zhang]{dai2019SAN}
Tao Dai, Jianrui Cai, Yongbing Zhang, Shu-Tao Xia, and Lei Zhang.
\newblock Second-order attention network for single image super-resolution.
\newblock In \emph{CVPR}, pp.\  11065--11074, 2019.

\bibitem[Deng et~al.(2009)Deng, Dong, Socher, Li, Li, and Fei-Fei]{deng2009imagenet}
Jia Deng, Wei Dong, Richard Socher, Li-Jia Li, Kai Li, and Li~Fei-Fei.
\newblock Image{N}et: A large-scale hierarchical image database.
\newblock In \emph{CVPR}, pp.\  248--255. IEEE, 2009.

\bibitem[Dong et~al.(2014)Dong, Loy, He, and Tang]{dong2014learning}
Chao Dong, Chen~Change Loy, Kaiming He, and Xiaoou Tang.
\newblock Learning a deep convolutional network for image super-resolution.
\newblock In \emph{ECCV}, pp.\  184--199. Springer, 2014.

\bibitem[Dosovitskiy et~al.(2020)Dosovitskiy, Beyer, Kolesnikov, Weissenborn, Zhai, Unterthiner, Dehghani, Minderer, Heigold, Gelly, et~al.]{dosovitskiy2020image}
Alexey Dosovitskiy, Lucas Beyer, Alexander Kolesnikov, Dirk Weissenborn, Xiaohua Zhai, Thomas Unterthiner, Mostafa Dehghani, Matthias Minderer, Georg Heigold, Sylvain Gelly, et~al.
\newblock An image is worth 16x16 words: Transformers for image recognition at scale.
\newblock \emph{arXiv preprint arXiv:2010.11929}, 2020.

\bibitem[Ehrlich et~al.(2020)Ehrlich, Davis, Lim, and Shrivastava]{ehrlich2020quantization}
Max Ehrlich, Larry Davis, Ser-Nam Lim, and Abhinav Shrivastava.
\newblock Quantization guided {JPEG} artifact correction.
\newblock In \emph{ECCV}, pp.\  293--309. Springer, 2020.

\bibitem[Foi et~al.(2007)Foi, Katkovnik, and Egiazarian]{foi2007Classic5}
Alessandro Foi, Vladimir Katkovnik, and Karen Egiazarian.
\newblock Pointwise shape-adaptive dct for high-quality denoising and deblocking of grayscale and color images.
\newblock \emph{IEEE TIP}, 16\penalty0 (5):\penalty0 1395--1411, 2007.

\bibitem[Franzen(1999)]{franzen1999kodak}
Rich Franzen.
\newblock Kodak lossless true color image suite.
\newblock \emph{source: http://r0k. us/graphics/kodak}, 4\penalty0 (2), 1999.

\bibitem[Gharbi et~al.(2016)Gharbi, Chaurasia, Paris, and Durand]{gharbi2016deep}
Micha{\"e}l Gharbi, Gaurav Chaurasia, Sylvain Paris, and Fr{\'e}do Durand.
\newblock Deep joint demosaicking and denoising.
\newblock \emph{ACM TOG}, 35\penalty0 (6):\penalty0 1--12, 2016.

\bibitem[Goyal(2017)]{goyal2017accurate}
P~Goyal.
\newblock Accurate, large minibatch sg d: training imagenet in 1 hour.
\newblock \emph{arXiv preprint arXiv:1706.02677}, 2017.

\bibitem[Gu \& Dao(2023)Gu and Dao]{gu2023mamba}
Albert Gu and Tri Dao.
\newblock Mamba: Linear-time sequence modeling with selective state spaces.
\newblock \emph{arXiv preprint arXiv:2312.00752}, 2023.

\bibitem[Guo et~al.(2024)Guo, Li, Dai, Ouyang, Ren, and Xia]{guo2024mambair}
Hang Guo, Jinmin Li, Tao Dai, Zhihao Ouyang, Xudong Ren, and Shu-Tao Xia.
\newblock {MambaIR}: A simple baseline for image restoration with state-space model.
\newblock \emph{arXiv preprint arXiv:2402.15648}, 2024.

\bibitem[Guo et~al.(2020)Guo, Jin, Facciolo, Zeng, and Morel]{guo2020residual}
Yu~Guo, Qiyu Jin, Gabriele Facciolo, Tieyong Zeng, and Jean-Michel Morel.
\newblock Residual learning for effective joint demosaicing-denoising.
\newblock \emph{arXiv preprint arXiv:2009.06205}, 2020.

\bibitem[Huang et~al.(2015)Huang, Singh, and Ahuja]{huang2015single}
Jia-Bin Huang, Abhishek Singh, and Narendra Ahuja.
\newblock Single image super-resolution from transformed self-exemplars.
\newblock In \emph{CVPR}, pp.\  5197--5206, 2015.

\bibitem[Huang et~al.(2021)Huang, Ben, Luo, Cheng, Yu, and Fu]{huang2021shuffle}
Zilong Huang, Youcheng Ben, Guozhong Luo, Pei Cheng, Gang Yu, and Bin Fu.
\newblock Shuffle transformer: Rethinking spatial shuffle for vision transformer.
\newblock \emph{arXiv preprint arXiv:2106.03650}, 2021.

\bibitem[Jiang et~al.(2021)Jiang, Zhang, and Timofte]{jiang2021FBCNN}
Jiaxi Jiang, Kai Zhang, and Radu Timofte.
\newblock Towards flexible blind {JPEG} artifacts removal.
\newblock In \emph{ICCV}, pp.\  4997--5006, 2021.

\bibitem[Johnson et~al.(2016)Johnson, Alahi, and Fei-Fei]{johnson2016perceptual}
Justin Johnson, Alexandre Alahi, and Li~Fei-Fei.
\newblock Perceptual losses for real-time style transfer and super-resolution.
\newblock In \emph{ECCV}, pp.\  694--711. Springer, 2016.

\bibitem[Kalra \& Barkeshli(2024)Kalra and Barkeshli]{kalra2024warmup}
Dayal~Singh Kalra and Maissam Barkeshli.
\newblock Why warmup the learning rate? underlying mechanisms and improvements.
\newblock \emph{arXiv preprint arXiv:2406.09405}, 2024.

\bibitem[Kang et~al.(2023)Kang, Zhu, Zhang, Park, Shechtman, Paris, and Park]{kang2023scaling}
Minguk Kang, Jun-Yan Zhu, Richard Zhang, Jaesik Park, Eli Shechtman, Sylvain Paris, and Taesung Park.
\newblock Scaling up {GAN}s for text-to-image synthesis.
\newblock In \emph{CVPR}, pp.\  10124--10134, 2023.

\bibitem[Karaali \& Jung(2017)Karaali and Jung]{karaali2017edge_EBDB}
Ali Karaali and Claudio~Rosito Jung.
\newblock Edge-based defocus blur estimation with adaptive scale selection.
\newblock \emph{TIP}, 2017.

\bibitem[Kiku et~al.(2016)Kiku, Monno, Tanaka, and Okutomi]{kiku2016beyond}
Daisuke Kiku, Yusuke Monno, Masayuki Tanaka, and Masatoshi Okutomi.
\newblock Beyond color difference: Residual interpolation for color image demosaicking.
\newblock \emph{IEEE TIP}, 25\penalty0 (3):\penalty0 1288--1300, 2016.

\bibitem[Kim et~al.(2016)Kim, Lee, and Lee]{kim2016accurate}
Jiwon Kim, Jung~Kwon Lee, and Kyoung~Mu Lee.
\newblock Accurate image super-resolution using very deep convolutional networks.
\newblock In \emph{CVPR}, pp.\  1646--1654, 2016.

\bibitem[Kim et~al.(2022)Kim, Lee, and Cho]{kim2022mssnet}
Kiyeon Kim, Seungyong Lee, and Sunghyun Cho.
\newblock {MSSNet}: Multi-scale-stage network for single image deblurring.
\newblock In \emph{ECCVW}, pp.\  524--539. Springer, 2022.

\bibitem[Kingma \& Ba(2014)Kingma and Ba]{kingma2014adam}
Diederik~P Kingma and Jimmy Ba.
\newblock Adam: A method for stochastic optimization.
\newblock \emph{arXiv preprint arXiv:1412.6980}, 2014.

\bibitem[Kupyn et~al.(2019)Kupyn, Martyniuk, Wu, and Wang]{deblurganv2}
Orest Kupyn, Tetiana Martyniuk, Junru Wu, and Zhangyang Wang.
\newblock {DeblurGAN-v2}: Deblurring (orders-of-magnitude) faster and better.
\newblock In \emph{ICCV}, 2019.

\bibitem[Lee et~al.(2019)Lee, Lee, Cho, and Lee]{lee2019deep_dmenet}
Junyong Lee, Sungkil Lee, Sunghyun Cho, and Seungyong Lee.
\newblock Deep defocus map estimation using domain adaptation.
\newblock In \emph{CVPR}, 2019.

\bibitem[Lee et~al.(2021)Lee, Son, Rim, Cho, and Lee]{Lee_2021_CVPRifan}
Junyong Lee, Hyeongseok Son, Jaesung Rim, Sunghyun Cho, and Seungyong Lee.
\newblock Iterative filter adaptive network for single image defocus deblurring.
\newblock In \emph{CVPR}, 2021.

\bibitem[Li et~al.(2019{\natexlab{a}})Li, Cheong, and Tan]{li2019heavy}
Ruoteng Li, Loong-Fah Cheong, and Robby~T Tan.
\newblock Heavy rain image restoration: Integrating physics model and conditional adversarial learning.
\newblock In \emph{CVPR}, pp.\  1633--1642, 2019{\natexlab{a}}.

\bibitem[Li et~al.(2020)Li, Tan, and Cheong]{li2020all}
Ruoteng Li, Robby~T Tan, and Loong-Fah Cheong.
\newblock All in one bad weather removal using architectural search.
\newblock In \emph{CVPR}, pp.\  3175--3185, 2020.

\bibitem[Li et~al.(2021)Li, Lu, Lu, Zhang, and Jia]{li2021efficient}
Wenbo Li, Xin Lu, Jiangbo Lu, Xiangyu Zhang, and Jiaya Jia.
\newblock On efficient transformer and image pre-training for low-level vision.
\newblock \emph{arXiv preprint arXiv:2112.10175}, 2021.

\bibitem[Li et~al.(2023{\natexlab{a}})Li, Fan, Xiang, Demandolx, Ranjan, Timofte, and Van~Gool]{li2023efficient}
Yawei Li, Yuchen Fan, Xiaoyu Xiang, Denis Demandolx, Rakesh Ranjan, Radu Timofte, and Luc Van~Gool.
\newblock Efficient and explicit modelling of image hierarchies for image restoration.
\newblock In \emph{CVPR}, pp.\  18278--18289, 2023{\natexlab{a}}.

\bibitem[Li et~al.(2023{\natexlab{b}})Li, Zhang, Liang, Cao, Liu, Gong, Zhang, Tang, Liu, Demandolx, et~al.]{li2023lsdir}
Yawei Li, Kai Zhang, Jingyun Liang, Jiezhang Cao, Ce~Liu, Rui Gong, Yulun Zhang, Hao Tang, Yun Liu, Denis Demandolx, et~al.
\newblock {LSDIR}: A large scale dataset for image restoration.
\newblock In \emph{CVPRW}, pp.\  1775--1787, 2023{\natexlab{b}}.

\bibitem[Li et~al.(2019{\natexlab{b}})Li, Yang, Liu, Yang, Jeon, and Wu]{li2019feedback}
Zhen Li, Jinglei Yang, Zheng Liu, Xiaomin Yang, Gwanggil Jeon, and Wei Wu.
\newblock Feedback network for image super-resolution.
\newblock In \emph{CVPR}, pp.\  3867--3876, 2019{\natexlab{b}}.

\bibitem[Li et~al.(2022)Li, Liu, Chen, Cai, Gu, Qiao, and Dong]{li2022blueprint}
Zheyuan Li, Yingqi Liu, Xiangyu Chen, Haoming Cai, Jinjin Gu, Yu~Qiao, and Chao Dong.
\newblock Blueprint separable residual network for efficient image super-resolution.
\newblock In \emph{CVPR}, pp.\  833--843, 2022.

\bibitem[Liang et~al.(2021)Liang, Cao, Sun, Zhang, Van~Gool, and Timofte]{liang2021swinir}
Jingyun Liang, Jiezhang Cao, Guolei Sun, Kai Zhang, Luc Van~Gool, and Radu Timofte.
\newblock {SwinIR}: Image restoration using swin transformer.
\newblock In \emph{ICCVW}, pp.\  1833--1844, 2021.

\bibitem[Lim et~al.(2017)Lim, Son, Kim, Nah, and Lee]{lim2017enhanced}
Bee Lim, Sanghyun Son, Heewon Kim, Seungjun Nah, and Kyoung~Mu Lee.
\newblock Enhanced deep residual networks for single image super-resolution.
\newblock In \emph{CVPRW}, pp.\  1132--1140, 2017.

\bibitem[Liu et~al.(2018)Liu, Jaw, Huang, and Hwang]{liu2018desnownet}
Yun-Fu Liu, Da-Wei Jaw, Shih-Chia Huang, and Jenq-Neng Hwang.
\newblock {DesnowNet}: Context-aware deep network for snow removal.
\newblock \emph{IEEE TIP}, 27\penalty0 (6):\penalty0 3064--3073, 2018.

\bibitem[Liu et~al.(2021)Liu, Lin, Cao, Hu, Wei, Zhang, Lin, and Guo]{liu2021swin}
Ze~Liu, Yutong Lin, Yue Cao, Han Hu, Yixuan Wei, Zheng Zhang, Stephen Lin, and Baining Guo.
\newblock Swin transformer: Hierarchical vision transformer using shifted windows.
\newblock In \emph{ICCV}, pp.\  10012--10022, 2021.

\bibitem[Liu et~al.(2022)Liu, Hu, Lin, Yao, Xie, Wei, Ning, Cao, Zhang, Dong, et~al.]{liu2022swin}
Ze~Liu, Han Hu, Yutong Lin, Zhuliang Yao, Zhenda Xie, Yixuan Wei, Jia Ning, Yue Cao, Zheng Zhang, Li~Dong, et~al.
\newblock Swin transformer v2: Scaling up capacity and resolution.
\newblock In \emph{CVPR}, pp.\  12009--12019, 2022.

\bibitem[Loshchilov \& Hutter(2018)Loshchilov and Hutter]{loshchilov2018decoupled}
Ilya Loshchilov and Frank Hutter.
\newblock Decoupled weight decay regularization.
\newblock In \emph{ICLR}, 2018.

\bibitem[Ma et~al.(2016)Ma, Duanmu, Wu, Wang, Yong, Li, and Zhang]{ma2016waterloo}
Kede Ma, Zhengfang Duanmu, Qingbo Wu, Zhou Wang, Hongwei Yong, Hongliang Li, and Lei Zhang.
\newblock Waterloo exploration database: New challenges for image quality assessment models.
\newblock \emph{IEEE TIP}, 26\penalty0 (2):\penalty0 1004--1016, 2016.

\bibitem[Mao et~al.(2023)Mao, Liu, Liu, Li, Shen, and Wang]{mao2023intriguing}
Xintian Mao, Yiming Liu, Fengze Liu, Qingli Li, Wei Shen, and Yan Wang.
\newblock Intriguing findings of frequency selection for image deblurring.
\newblock In \emph{AAAI}, pp.\  1905--1913, 2023.

\bibitem[Martin et~al.(2001)Martin, Fowlkes, Tal, and Malik]{martin2001database}
David Martin, Charless Fowlkes, Doron Tal, and Jitendra Malik.
\newblock A database of human segmented natural images and its application to evaluating segmentation algorithms and measuring ecological statistics.
\newblock In \emph{ICCV}, volume~2, pp.\  416--423. IEEE, 2001.

\bibitem[Matsui et~al.(2017)Matsui, Ito, Aramaki, Fujimoto, Ogawa, Yamasaki, and Aizawa]{matsui2017sketch}
Yusuke Matsui, Kota Ito, Yuji Aramaki, Azuma Fujimoto, Toru Ogawa, Toshihiko Yamasaki, and Kiyoharu Aizawa.
\newblock Sketch-based manga retrieval using manga109 dataset.
\newblock \emph{Multimedia Tools and Applications}, 76\penalty0 (20):\penalty0 21811--21838, 2017.

\bibitem[Mei et~al.(2021)Mei, Fan, and Zhou]{mei2021NLSA}
Yiqun Mei, Yuchen Fan, and Yuqian Zhou.
\newblock Image super-resolution with non-local sparse attention.
\newblock In \emph{CVPR}, pp.\  3517--3526, 2021.

\bibitem[Nah et~al.(2017)Nah, Hyun~Kim, and Mu~Lee]{nah2017deep}
Seungjun Nah, Tae Hyun~Kim, and Kyoung Mu~Lee.
\newblock Deep multi-scale convolutional neural network for dynamic scene deblurring.
\newblock In \emph{CVPR}, pp.\  3883--3891, 2017.

\bibitem[Niu et~al.(2020)Niu, Wen, Ren, Zhang, Yang, Wang, Zhang, Cao, and Shen]{niu2020HAN}
Ben Niu, Weilei Wen, Wenqi Ren, Xiangde Zhang, Lianping Yang, Shuzhen Wang, Kaihao Zhang, Xiaochun Cao, and Haifeng Shen.
\newblock Single image super-resolution via a holistic attention network.
\newblock In \emph{ECCV}, pp.\  191--207, 2020.

\bibitem[Purohit et~al.(2021)Purohit, Suin, Rajagopalan, and Boddeti]{purohit2021spatially_spair}
Kuldeep Purohit, Maitreya Suin, AN~Rajagopalan, and Vishnu~Naresh Boddeti.
\newblock Spatially-adaptive image restoration using distortion-guided networks.
\newblock In \emph{ICCV}, 2021.

\bibitem[Qian et~al.(2018)Qian, Tan, Yang, Su, and Liu]{qian2018attentive}
Rui Qian, Robby~T Tan, Wenhan Yang, Jiajun Su, and Jiaying Liu.
\newblock Attentive generative adversarial network for raindrop removal from a single image.
\newblock In \emph{CVPR}, pp.\  2482--2491, 2018.

\bibitem[Ren et~al.(2024)Ren, Li, Liang, Ranjan, Liu, Cucchiara, Van~Gool, Yang, and Sebe]{ren2024sharing}
Bin Ren, Yawei Li, Jingyun Liang, Rakesh Ranjan, Mengyuan Liu, Rita Cucchiara, Luc Van~Gool, Ming-Hsuan Yang, and Nicu Sebe.
\newblock Sharing key semantics in transformer makes efficient image restoration.
\newblock In \emph{NeurIPS}, 2024.

\bibitem[Richardson(1972)]{richardson1972bayesian}
William~Hadley Richardson.
\newblock Bayesian-based iterative method of image restoration.
\newblock \emph{JoSA}, 62\penalty0 (1):\penalty0 55--59, 1972.

\bibitem[Rim et~al.(2020)Rim, Lee, Won, and Cho]{rim2020real}
Jaesung Rim, Haeyun Lee, Jucheol Won, and Sunghyun Cho.
\newblock Real-world blur dataset for learning and benchmarking deblurring algorithms.
\newblock In \emph{ECCV}, pp.\  184--201. Springer, 2020.

\bibitem[Saharia et~al.(2022)Saharia, Chan, Saxena, Li, Whang, Denton, Ghasemipour, Gontijo~Lopes, Karagol~Ayan, Salimans, et~al.]{saharia2022photorealistic}
Chitwan Saharia, William Chan, Saurabh Saxena, Lala Li, Jay Whang, Emily~L Denton, Kamyar Ghasemipour, Raphael Gontijo~Lopes, Burcu Karagol~Ayan, Tim Salimans, et~al.
\newblock Photorealistic text-to-image diffusion models with deep language understanding.
\newblock \emph{NeurIPS}, 35:\penalty0 36479--36494, 2022.

\bibitem[Sheikh(2005)]{sheikh2005live}
HR~Sheikh.
\newblock Live image quality assessment database release 2.
\newblock \emph{http://live. ece. utexas. edu/research/quality}, 2005.

\bibitem[Shen et~al.(2019)Shen, Wang, Lu, Shen, Ling, Xu, and Shao]{shen2019human}
Ziyi Shen, Wenguan Wang, Xiankai Lu, Jianbing Shen, Haibin Ling, Tingfa Xu, and Ling Shao.
\newblock Human-aware motion deblurring.
\newblock In \emph{ICCV}, pp.\  5572--5581, 2019.

\bibitem[Shi et~al.(2015)Shi, Xu, and Jia]{shi2015just_jnb}
Jianping Shi, Li~Xu, and Jiaya Jia.
\newblock Just noticeable defocus blur detection and estimation.
\newblock In \emph{CVPR}, 2015.

\bibitem[Simonyan \& Zisserman(2015)Simonyan and Zisserman]{simonyan2015very}
Karen Simonyan and Andrew Zisserman.
\newblock Very deep convolutional networks for large-scale image recognition.
\newblock In \emph{ICLR}, 2015.

\bibitem[Son et~al.(2021)Son, Lee, Cho, and Lee]{son2021single_kpac}
Hyeongseok Son, Junyong Lee, Sunghyun Cho, and Seungyong Lee.
\newblock Single image defocus deblurring using kernel-sharing parallel atrous convolutions.
\newblock In \emph{ICCV}, 2021.

\bibitem[Tao et~al.(2018)Tao, Gao, Shen, Wang, and Jia]{tao2018scale}
Xin Tao, Hongyun Gao, Xiaoyong Shen, Jue Wang, and Jiaya Jia.
\newblock Scale-recurrent network for deep image deblurring.
\newblock In \emph{CVPR}, 2018.

\bibitem[Touvron et~al.(2023)Touvron, Lavril, Izacard, Martinet, Lachaux, Lacroix, Rozi{\`e}re, Goyal, Hambro, Azhar, et~al.]{touvron2023llama}
Hugo Touvron, Thibaut Lavril, Gautier Izacard, Xavier Martinet, Marie-Anne Lachaux, Timoth{\'e}e Lacroix, Baptiste Rozi{\`e}re, Naman Goyal, Eric Hambro, Faisal Azhar, et~al.
\newblock {LLaMA}: Open and efficient foundation language models.
\newblock \emph{arXiv preprint arXiv:2302.13971}, 2023.

\bibitem[Tsai et~al.(2022{\natexlab{a}})Tsai, Peng, Lin, Tsai, and Lin]{tsai2022stripformer}
Fu-Jen Tsai, Yan-Tsung Peng, Yen-Yu Lin, Chung-Chi Tsai, and Chia-Wen Lin.
\newblock Stripformer: Strip transformer for fast image deblurring.
\newblock In \emph{ECCV}, pp.\  146--162. Springer, 2022{\natexlab{a}}.

\bibitem[Tsai et~al.(2022{\natexlab{b}})Tsai, Peng, Tsai, Lin, and Lin]{tsai2022banet}
Fu-Jen Tsai, Yan-Tsung Peng, Chung-Chi Tsai, Yen-Yu Lin, and Chia-Wen Lin.
\newblock {BANet}: A blur-aware attention network for dynamic scene deblurring.
\newblock \emph{IEEE TIP}, 31:\penalty0 6789--6799, 2022{\natexlab{b}}.

\bibitem[Tu et~al.(2022)Tu, Talebi, Zhang, Yang, Milanfar, Bovik, and Li]{tu2022maxim}
Zhengzhong Tu, Hossein Talebi, Han Zhang, Feng Yang, Peyman Milanfar, Alan Bovik, and Yinxiao Li.
\newblock {MAXIM}: Multi-axis {MLP} for image processing.
\newblock In \emph{CVPR}, pp.\  5769--5780, 2022.

\bibitem[Valanarasu et~al.(2022)Valanarasu, Yasarla, and Patel]{valanarasu2022transweather}
Jeya Maria~Jose Valanarasu, Rajeev Yasarla, and Vishal~M Patel.
\newblock {TransWeather}: Transformer-based restoration of images degraded by adverse weather conditions.
\newblock In \emph{CVPR}, pp.\  2353--2363, 2022.

\bibitem[Vaswani et~al.(2017)Vaswani, Shazeer, Parmar, Uszkoreit, Jones, Gomez, Kaiser, and Polosukhin]{vaswani2017attention}
Ashish Vaswani, Noam Shazeer, Niki Parmar, Jakob Uszkoreit, Llion Jones, Aidan~N Gomez, {\L}ukasz Kaiser, and Illia Polosukhin.
\newblock Attention is all you need.
\newblock \emph{NeurIPS}, 30, 2017.

\bibitem[Wang et~al.(2020)Wang, Li, Khabsa, Fang, and Ma]{wang2020linformer}
Sinong Wang, Belinda~Z Li, Madian Khabsa, Han Fang, and Hao Ma.
\newblock Linformer: Self-attention with linear complexity.
\newblock \emph{arXiv preprint arXiv:2006.04768}, 2020.

\bibitem[Wang et~al.(2018)Wang, Yu, Wu, Gu, Liu, Dong, Qiao, and Change~Loy]{wang2018esrgan}
Xintao Wang, Ke~Yu, Shixiang Wu, Jinjin Gu, Yihao Liu, Chao Dong, Yu~Qiao, and Chen Change~Loy.
\newblock {ESRGAN}: Enhanced super-resolution generative adversarial networks.
\newblock In \emph{ECCVW}, pp.\  0--0, 2018.

\bibitem[Wang et~al.(2022)Wang, Cun, Bao, Zhou, Liu, and Li]{wang2022uformer}
Zhendong Wang, Xiaodong Cun, Jianmin Bao, Wengang Zhou, Jianzhuang Liu, and Houqiang Li.
\newblock Uformer: A general {U}-shaped transformer for image restoration.
\newblock In \emph{CVPR}, pp.\  17683--17693, 2022.

\bibitem[Wu et~al.(2016)Wu, Timofte, and Van~Gool]{wu2016demosaicing}
Jiqing Wu, Radu Timofte, and Luc Van~Gool.
\newblock Demosaicing based on directional difference regression and efficient regression priors.
\newblock \emph{IEEE TIP}, 25\penalty0 (8):\penalty0 3862--3874, 2016.

\bibitem[Xiao et~al.(2023)Xiao, Fu, Zhou, Liu, and Zha]{xiao2023random}
Jie Xiao, Xueyang Fu, Man Zhou, Hongjian Liu, and Zheng-Jun Zha.
\newblock Random shuffle transformer for image restoration.
\newblock In \emph{ICML}, pp.\  38039--38058, 2023.

\bibitem[Xu et~al.(2021)Xu, Zhang, Zhang, and Tao]{xu2021vitae}
Yufei Xu, Qiming Zhang, Jing Zhang, and Dacheng Tao.
\newblock Vitae: Vision transformer advanced by exploring intrinsic inductive bias.
\newblock \emph{NeurIPS}, 34:\penalty0 28522--28535, 2021.

\bibitem[Yu et~al.(2024)Yu, Gu, Li, Hu, Kong, Wang, He, Qiao, and Dong]{yu2024scaling}
Fanghua Yu, Jinjin Gu, Zheyuan Li, Jinfan Hu, Xiangtao Kong, Xintao Wang, Jingwen He, Yu~Qiao, and Chao Dong.
\newblock Scaling up to excellence: Practicing model scaling for photo-realistic image restoration in the wild.
\newblock In \emph{CVPR}, 2024.

\bibitem[Zamir et~al.(2021)Zamir, Arora, Khan, Hayat, Khan, Yang, and Shao]{zamir2021multi}
Syed~Waqas Zamir, Aditya Arora, Salman Khan, Munawar Hayat, Fahad~Shahbaz Khan, Ming-Hsuan Yang, and Ling Shao.
\newblock Multi-stage progressive image restoration.
\newblock In \emph{CVPR}, pp.\  14821--14831, 2021.

\bibitem[Zamir et~al.(2022)Zamir, Arora, Khan, Hayat, Khan, and Yang]{zamir2022restormer}
Syed~Waqas Zamir, Aditya Arora, Salman Khan, Munawar Hayat, Fahad~Shahbaz Khan, and Ming-Hsuan Yang.
\newblock Restormer: Efficient transformer for high-resolution image restoration.
\newblock In \emph{CVPR}, pp.\  5728--5739, 2022.

\bibitem[Zeyde et~al.(2010)Zeyde, Elad, and Protter]{zeyde2010single}
Roman Zeyde, Michael Elad, and Matan Protter.
\newblock On single image scale-up using sparse-representations.
\newblock In \emph{Proceedings of International Conference on Curves and Surfaces}, pp.\  711--730. Springer, 2010.

\bibitem[Zhang et~al.(2022)Zhang, Zhang, Gu, Zhang, Kong, and Yuan]{zhang2022accurate}
Jiale Zhang, Yulun Zhang, Jinjin Gu, Yongbing Zhang, Linghe Kong, and Xin Yuan.
\newblock Accurate image restoration with attention retractable transformer.
\newblock In \emph{ICLR}, 2022.

\bibitem[Zhang et~al.(2023)Zhang, Zhang, Gu, Dong, Kong, and Yang]{zhang2023xformer}
Jiale Zhang, Yulun Zhang, Jinjin Gu, Jiahua Dong, Linghe Kong, and Xiaokang Yang.
\newblock Xformer: Hybrid x-shaped transformer for image denoising.
\newblock \emph{arXiv preprint arXiv:2303.06440}, 2023.

\bibitem[Zhang et~al.(2017{\natexlab{a}})Zhang, Zuo, Chen, Meng, and Zhang]{zhang2017beyond}
Kai Zhang, Wangmeng Zuo, Yunjin Chen, Deyu Meng, and Lei Zhang.
\newblock Beyond a {G}aussian denoiser: residual learning of deep {CNN} for image denoising.
\newblock \emph{IEEE TIP}, 26\penalty0 (7):\penalty0 3142--3155, 2017{\natexlab{a}}.

\bibitem[Zhang et~al.(2017{\natexlab{b}})Zhang, Zuo, Gu, and Zhang]{zhang2017learning}
Kai Zhang, Wangmeng Zuo, Shuhang Gu, and Lei Zhang.
\newblock Learning deep cnn denoiser prior for image restoration.
\newblock In \emph{CVPR}, pp.\  3929--3938, 2017{\natexlab{b}}.

\bibitem[Zhang et~al.(2018{\natexlab{a}})Zhang, Zuo, and Zhang]{zhang2018ffdnet}
Kai Zhang, Wangmeng Zuo, and Lei Zhang.
\newblock Ffdnet: Toward a fast and flexible solution for cnn-based image denoising.
\newblock \emph{IEEE TIP}, 27\penalty0 (9):\penalty0 4608--4622, 2018{\natexlab{a}}.

\bibitem[Zhang et~al.(2021)Zhang, Li, Zuo, Zhang, Van~Gool, and Timofte]{zhang2021plug}
Kai Zhang, Yawei Li, Wangmeng Zuo, Lei Zhang, Luc Van~Gool, and Radu Timofte.
\newblock Plug-and-play image restoration with deep denoiser prior.
\newblock \emph{IEEE TPAMI}, 2021.

\bibitem[Zhang et~al.(2011)Zhang, Wu, Buades, and Li]{zhang2011color}
Lei Zhang, Xiaolin Wu, Antoni Buades, and Xin Li.
\newblock Color demosaicking by local directional interpolation and nonlocal adaptive thresholding.
\newblock \emph{Journal of Electronic imaging}, 20\penalty0 (2):\penalty0 023016, 2011.

\bibitem[Zhang et~al.(2018{\natexlab{b}})Zhang, Li, Li, Wang, Zhong, and Fu]{zhang2018rcan}
Yulun Zhang, Kunpeng Li, Kai Li, Lichen Wang, Bineng Zhong, and Yun Fu.
\newblock Image super-resolution using very deep residual channel attention networks.
\newblock In \emph{ECCV}, pp.\  286--301, 2018{\natexlab{b}}.

\bibitem[Zhang et~al.(2018{\natexlab{c}})Zhang, Tian, Kong, Zhong, and Fu]{zhang2018residual}
Yulun Zhang, Yapeng Tian, Yu~Kong, Bineng Zhong, and Yun Fu.
\newblock Residual dense network for image super-resolution.
\newblock In \emph{CVPR}, 2018{\natexlab{c}}.

\bibitem[Zhang et~al.(2019)Zhang, Li, Li, Zhong, and Fu]{zhang2019residual}
Yulun Zhang, Kunpeng Li, Kai Li, Bineng Zhong, and Yun Fu.
\newblock Residual non-local attention networks for image restoration.
\newblock \emph{arXiv preprint arXiv:1903.10082}, 2019.

\end{thebibliography}
